\documentclass[twoside, 11pt]{article}
\usepackage{jair, theapa, rawfonts}
\usepackage[utf8]{inputenc}
\usepackage{todonotes}
\usepackage{xcolor}
\usepackage{amsfonts}

\usepackage{algorithm}
\usepackage{algorithmic}
\usepackage{fancyvrb}

\usepackage{amsmath,amssymb}
\usepackage{enumitem}
\usepackage{balance}
\usepackage{listings}
\usepackage{booktabs}
\usepackage{mathtools}
\usepackage{pxfonts}
\usepackage{graphicx}
\usepackage{subcaption}
\usepackage{multirow}

\newcommand{\citet}[1]{\citeauthor{#1}~\shortcite{#1}}

\newtheorem{definition}{Definition}
\usepackage{url}

\def\Eff{\mathit{Eff}}

\ShortHeadings{Contrastive Explanations of Plans through Model Restrictions}{Krarup, Krivic, Magazzeni, Long, Cashmore, \& Smith}
\firstpageno{1}

\begin{document}

\title{Contrastive Explanations of Plans Through Model Restrictions}
\author{\name Benjamin Krarup \email benjamin.krarup@kcl.ac.uk\\
\name Senka Krivic \email senka.krivic@kcl.ac.uk \\
\name Daniele Magazzeni \email daniele.magazzeni@kcl.ac.uk \\
\name Derek Long \email derek.long@kcl.ac.uk \\
    \addr King's College London, Bush House, WC2B 4BG, London, UK
    \AND
\name Michael Cashmore \email michael.cashmore@strath.ac.uk \\
    \addr University of Strathclyde, Livingstone Tower, G1 1XH, Glasgow, UK
    \AND
\name David E. Smith \email david.smith@psresearch.xyz\\
    \addr PS Research, 25960 Quail Ln, Los Altos Hills, CA 94022, USA\\
}

\maketitle

\begin{abstract}
In automated planning, the need for explanations arises when there is a mismatch between a proposed plan and the user’s expectation. We frame Explainable AI Planning in the context of the plan negotiation problem, in which a succession of hypothetical planning problems are generated and solved. The object of the negotiation is for the user to understand and ultimately arrive at a satisfactory plan.
We present the results of a user study that demonstrates that when users ask questions about plans, those questions are contrastive, i.e. ``why A rather than B?''. We use the data from this study to construct a taxonomy of user questions that often arise during plan negotiation.
We formally define our approach to plan negotiation through model restriction as an iterative process. This approach generates hypothetical problems and contrastive plans by restricting the model through constraints implied by user questions.
We formally define model-based compilations in PDDL2.1 of each constraint derived from a user question in the taxonomy, and empirically evaluate the compilations in terms of computational complexity.
The compilations were implemented as part of an explanation framework that employs iterative model restriction. We demonstrate its benefits in a second user study.
\end{abstract}




\section{Introduction}
Automated planning is being used in increasingly complex applications, and explanation plays an important role in building trust, both in planners and in the plans they produce. A plan is a form of communication, either as a set of instructions to be enacted by autonomous or human agents, or as a proposal of intention communicated to a user. In either case, the plan conveys the means by which a goal is to be achieved, but not the reasons for the choices it embodies. When the audience for a plan includes humans then it is natural to suppose that the audience might wish to question the reasoning, intention and underlying assumptions that lead to those choices.

The need for explanations arises when there is a mismatch between a proposed plan and the audience's expectation. This might be because the audience had not managed to form an expected plan, or because a plan was successfully constructed, but it did not match the proposed plan. Explanations attempt to bridge the gap between these mismatched positions and might be {\em local}, focusing on the specific proposed plan and its properties, or {\em global}, focusing on the assumptions on which the plan rests, or the process by which it was constructed. 

In this paper we focus on local explanations, investigating the form of queries made by a user in interaction with a planner or plan-based system. We suppose that the audience might want to question why the plan is structured as it is, what intentions the plan seeks to address, and what alternative plans might be considered. Through active exploration of these specific cases, the user might also gain global insight into the way in which the planner makes decisions~\cite{lipton_1990,lip16,rib16}.

We treat explanation as a form of dialogue, an iterative process in which the user asks contrastive questions~\cite{mil18} (that is, questions of the form `why $A$ rather than $B$?') where the constrasting position is specified as a constraint that restricts the forms of acceptable solutions to the original problem, and responses are given in the form of alternative plans, satisfying the newly added constraints. 
We observe that many purposeful queries made by a user in interaction with a planner or plan-based system are contrastive. Fox et al.~\citeyear{fox17} highlight the \textit{why} query as an important one for XAI, and discuss possible responses. To answer these kinds of questions, one must reason about the hypothetical alternative $B$, which means constructing an alternative plan for which $B$ is satisfied, rather than $A$.

We address the problem of planning subject to additional constraints by compiling the constraints into the planning model. This approach offers a useful benefit, that the same planner can be used to solve the constrained problem and its use is unaffected by the iterative explanation process in which it is exploited. The fact that the compilation is independent of the planner serves to emphasise that the explanations cannot directly address questions the user might have about the planning process, but focus on reconciling the planning models held by the planner and by the user.

This iterative model restriction process does not require that the planning models used by the planner and the user be the same. Indeed, the focus on model reconciliation presupposes that there is some difference between the models. Nevertheless, the formulation of questions as constraints does require that the user and the planner share vocabulary, including the names and parameter types of actions and predicates, and the names of objects appearing in the problem. We also do not assume that the user has necessarily formulated an explicit alternative plan. In some cases, the user might not have such a plan in mind and, in that case, the iterative process might simply reflect the user exploring the family of plans around the initial plan in order to gain some insight into the alternatives that exist.

In this paper we:
\begin{itemize}
\item Present a user study investigating the queries that arise when humans are confronted with plans, from which we develop a taxonomy of common questions.
\item Formally define the iterative model restriction process, through which explanations can be provided as part of a dialogue.
\item Present compilations for the common questions into PDDL2.1 constraints that can be used in a model-based approach for explanations within the iterative model restriction process. We empirically evaluate the computational impact of these compilations.
\item Describe an implementation of this process, and the framework in which plans are presented to the user for comparison.
\item Present an evaluation of the framework using a second user study.
\end{itemize}

The paper is structured as follows: in Section~\ref{sec:taxonomy} we introduce the idea of the Contrastive Taxonomy and present the list of formal user questions that will be considered throughout the paper. In Section~\ref{sec:back} we briefly cover the background in Explainable AI Planning. Then, in Section~\ref{sec:plans} we formally describe the iterative model restriction process along with a running example. In Section~\ref{sec:comp}, we present the compilations that can be used within the plan negotiation problem to encode the list of formal user questions. We describe the implementation of our Explainable Planning framework in Section~\ref{sec:iter}. In Section~\ref{sec:eval} we describe the user study carried out with the framework and present the results. Section~\ref{sec:eval} also includes empirical evaluations of the computational costs of the compilations. Section~\ref{sec:rel} contains a discussion of related work in explainable planning and model reconciliation. The paper concludes in Section~\ref{sec:conc} with a discussion of future work.
\section{Contrastive Taxonomy}\label{sec:taxonomy}
Several researchers have observed (e.g: Mueller et al.~\citeyear{mue19}) that it is useful to draw a distinction between {\em local} and {\em global} questions and the corresponding explanations.
Local questions are asked when users want explanations for specific decisions made in a system. Whereas global questions are asked when users want a better understanding of how the system makes decisions in general.
In both cases, the context might be restricted to explanations relative to a specific model, so that a local question asks about a specific decision made in solving a problem framed within that model, while a global question asks about the model as a whole or the way that the model is used by the system. In the context of a plan, a global question might be asked because the inquirer does not fully understand the model used by the planner, and therefore does not understand how the plan represents a solution. On the other hand, a local question can be asked even in the case that the user fully understands the model, but does not wish to reason through the details themselves, or does not understand why the plan is a good one.

As an example, when plan-based control was used to automate drilling~\cite{der18}, the process involved a series of stages during which the nature of required explanations evolved. During initial development users primarily asked global questions to validate their understanding of the model, ensuring its correctness, and building trust in the system. As this trust was built and the model used by the planner became well-understood, users were more likely to ask local questions seeking to understand the intention behind specific actions in a plan, or to better understand alternatives to that choice.

These local explanations are asked in a variety of contexts. Domain experts wish to challenge a decision made by the planner when they possess insight into the domain that they believe can improve upon a sub-optimal plan. Often the users are simply interested in exploring the space of plans and ask questions to suggest alternative decisions, and better understand their impact. In the former case, a sceptical expert might seek to demonstrate weakness in the way that the system made a decision, while in the latter case the role of the system is promoted to an advisor or aide, with the user relying on the system to support exploration of the space of alternative solutions.

The interrogative word used when asking local questions is \textit{why}, whereas for global questions it is usually \textit{how} or \textit{what}. Research from the social sciences \cite{mil18} argues that \textit{why} questions are typically \emph{contrastive}; that is, they are of the form ``Why A rather than some hypothetical \textit{foil} B?". Based on these observations, we hypothesise that when the model is well-known users ask more local, contrastive \textit{why} questions than global \textit{how} or \textit{what} questions. 

Contrastive questions capture the context of the question, they provide an insight into what the questioner needs in an explanation~\cite{Lewis1986-LEWCE}. Garfinkel~\citeyear{gar82} illustrates this with a story about a famous bank robber, Willie Sutton, who, when asked asked why he robbed banks, replied ``That's where the money is.". Sutton answered the question ``Why do you rob banks rather than other things?", instead of the question ``Why do you rob banks rather than not robbing them?". The foil was not explicitly stated in the question and so was left ambiguous. Garfinkel argues that explanations are relative to these contrastive contexts, and that they can be made unambiguous by explicitly stating the contrast case. 

A contrastive question asked about a plan can be answered with a \textit{contrastive explanation} which will highlight the differences between the original plan and a contrastive plan that accounts for the user suggested foil.
Providing contrastive explanations is not only effective in improving understanding, but is simpler than providing a full causal analysis~\cite{mil19}.
They are also naturally good for comparisons, as we can directly compare the original plan with a plan containing the user foil. 


To support our hypothesis empirically we investigated which questions users ask when faced with a plan produced by a planner. We conducted a study with 15 participants, which is a typical number for this type of user study~\cite{nie00,fau03}, to gain an insight into the types of questions that users pose about a planning system in three planning scenarios. Our null hypothesis and alternate hypothesis, $H_0$ and $H_a$ are as follows:

\begin{quote}$H_0$: Users ask an equal distribution of \textit{why}, \textit{how}, and \textit{what} questions about planning scenarios, when the model is well known.\end{quote}

\begin{quote}$H_a$: Users ask more \textit{why} questions than \textit{how} or \textit{what} questions about planning scenarios, when the model is well known.\end{quote}

Each question asked was categorised by the interrogative word used, either \textit{what}, \textit{how}, or \textit{why}. The full description of the experiment design, results, and analysis can be found in Appendix~\ref{appendixA}. The results are summarised in Table~\ref{tablehyp}. Performing a chi-square test, $\chi^2(2, 168) = 273.25$, P-value $< 0.00001$, these results are therefore significant at $p < 0.001$. We can therefore reject our null hypothesis, $H_0$, and accept our alternate hypothesis, $H_a$.

\begin{table}[t]
\centering
\begin{tabular}{lrrrr} 
 \toprule
 Question  &  &  & \\
 Type &     Video 1  &    Video 2  &   Video 3    \\
 \midrule
 What?         &   2  &  1    & 3\\
 How?          &   0    &  3    & 2\\
 Why?          &   65 &  50 & 42\\
 \bottomrule
\end{tabular}
\caption{Frequency of questions by video categorised by Miller's taxonomy.}
\label{tablehyp}
\end{table}

\subsection{Taxonomy of Questions}
Following our accepted hypothesis, we focus on providing explanations for \textit{why} questions. We categorised each \textit{why} question from the three different domains in the user study above into a taxonomy of questions which we call the \textit{Contrastive Taxonomy}.
The Contrastive Taxonomy is shown in Table~\ref{table0}. This shows the frequency of questions asked by users about the plans produced for three different domains, and represents a set of questions that are important for a plan-based system to answer.

The Contrastive Taxonomy shown in Figure~\ref{table0} illustrates the breakdown of the questions categorised by the explanatory objective of the question. The questions in categories FQ1 to FQ7 are of the form ``Why \textit{A} rather than \textit{B}?" and are clearly contrastive. They are also local questions because they each query decisions made in the plan in terms of actions that were or were not chosen to be performed and when. Prompted by the results gained from this study, we have chosen to focus on explanation for local \textit{why} questions. In this categorization, 89.9\% of the 168 questions are contrastive and local in nature. As a result, the remainder of this paper focuses on explanation for contrastive local questions.

A small number of questions that were not contrastive or local, \textit{how} (5) and \textit{what} (6) questions, were classed in the final category FQ8. There were also a small number (6) of \textit{why} questions that were classified as out of the scope of this paper. A question was classed out of the scope of the paper if it was not related to the planning system or the plan produced. For example, some participants questioned the animation system used to visualise the plan execution. These questions were still local and contrastive in nature, just not questions relevant to planning systems and therefore not ones we are concerned with answering.


In Section~\ref{sec:comp} we present a novel approach to compiling constraints derived from these questions into planning models to demonstrate the users query. Using the Contrastive Taxonomy, we can assert the percentage of user questions that we can address with this approach, as well as gain insights into the different types of questions users ask in real world examples. Our approach directly addresses formal question (FQ) types FQ1-7 which cover all of the contrastive questions asked by the users about plans in the above study. We can provide compilations of 89.8\% of the 168 questions that users asked.

\begin{table}[!t]
\centering
\begin{tabular}{ l p{13cm} r} 
 \toprule
 & Question Type& \#\\
 \midrule
 
FQ1 & Why is action A not used in the plan, rather than being used?        &   17\\
FQ2 & Why is action A used in the plan, rather than not being used?        &   75\\
FQ3 & Why is action A used in state S, rather than action B?              &   35\\
FQ4 & Why is action A used outside of time window W, rather than only being allowed within W?        &   6\\
FQ5 & Why is action A not performed before (after) action B, rather than A being performed after (before) B?         &   10\\
FQ6 & Why is action A not used in time window W, rather than being used within W? &   2\\
FQ7 & Why is action A used at time T, rather than at least some time T' after/before T?       &   6\\
FQ8 & Non-contrastive or out of scope& 17\\
 \bottomrule
\end{tabular}
\caption{Frequency of questions categorised into the Contrastive Taxonomy. We provide explanations for questions FQ1 - 7.} 
\label{table0}
\end{table}

\section{Background}\label{sec:back}

The primary thrust of this work is in the explanation of automatically generated plans, which can be seen as a special case of explanation of the output of AI programs in general. Even though the area of Explainable AI Planning (XAIP) is relatively young, there has been considerable work in the field in recent years. Chakraborti et al.~\citeyear{cha20} outline the different approaches to XAIP that have emerged in the last couple of years, and contrast them with earlier efforts in the field. They group the approaches for XAIP into two main categories: algorithm-based explanations and model-based explanations.
Algorithm-based explanations are typically {\em global} in nature, as they attempt to explain the underlying planning algorithm so that a user can better understand the workings of the planning system. For example, Magnaguagno et al.~\citeyear{mag17} provide an interactive visualisation of the search tree for a given problem. 
Model-based explanations are algorithm-agnostic methods for generating explanations for the solutions to a planning problem. These can be considered to be global or {\em local} explanations depending on whether the user is interested in the model itself, or in explaining particular decisions resulting from the model for a particular problem.

In this work, we focus on model-based local explanation. Within this framework, user questions about a plan can still result from two different sources: 1) differences between the user's domain model and the domain model used by the system, and 2) limitations in the user's (or planner's) reasoning abilities. When the planner and user have different models, the explanation problem becomes one of \textit{model reconciliation} -- identifying the differences between the two models so that the models can be updated to achieve reconciliation and an understanding of the source of the differences in plans (see, for example, work by Chakraborti et al.~\citeyear{cha17} and Sreedharan~\citeyear{sre18}, further discussed in Section~\ref{sec:rel}).

\begin{figure}[thb]
    \centering
    \includegraphics[width=0.4\columnwidth]{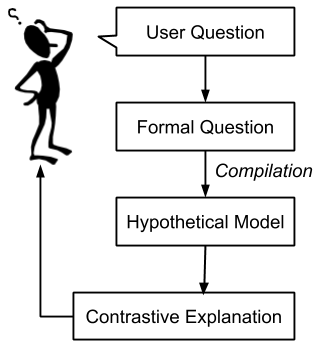}
    \caption{The four-stage iterative process for generating a contrastive explanation from a user question. The hypothetical model is created by compiling the formal question into the planning model (in PDDL 2.1).
    }
    \label{fig:process}
\end{figure}

In this paper, we present explanation as an iterative and collaborative process. We focus on contrastive questions that are motivated by some implied gap between the models of the world held by the user and by the system. The framework for within which the collaborative process takes place is the
four-stage mixed-initiative process illustrated in Figure~\ref{fig:process}. In this figure, (i) the user asks a contrastive question in natural language; (ii) a constraint is derived from the user question (forming the \textit{formal question});
(iii) a hypothetical planning model (HModel) is generated which encapsulates this constraint;
(iv) a solution for the HModel is called the HPlan, and it 
contains the contrast case expected by the user, and that can be compared to the original plan to show the consequence of the user suggestion. 

The user can compare plans and iterate the process by asking further questions, and refining the hypothetical model, or HModel. This allows the user to combine different compilations to create a more constrained HModel, producing more meaningful explanations, until the explanation is satisfactory. The process ends when the user is either satisfied with the explanation provided or with the plan generated for the HModel at some stage in this process. 

As a user engages with this process, through an interface that supports the construction of appropriate contrastive questions (see Section~\ref{sec:iter}), a collection of HModels can be constructed. 

The need for explanations typically arises when the solution generated by a planner does not match the users expectations.
The user might expect a particular solution, or they might expect that the solution exhibit qualities that the proposed solution does not. These expectations usually arise from a model held by the user (which might not be fully specified) that differs in some respects from the model used by the planner. We say `usually', because it could also be that the user holds an under-specified model and, in seeking an explanation of a plan, fills in details of their model leading to acceptance of the proposed plan. On the other hand, if the differences between the user model and the system model are more significant, the user can add constraints to the system model in order to generate HModels that more closely approximate their own model. We discuss this process more formally in Section~\ref{sec:plans}, but here we observe that it can be seen as a restricted form of model reconciliation that we refer to as {\em model restriction}. 

Model reconciliation arises when two different models attempt to describe the same phenomenon and yield different responses. In order to align the responses, one or both of the models must change. In general, for planning models, these changes could include a revision of the actions, the structure of the actions (pre- or post-conditions), changes in the collection of objects identified in the state, changes in the properties of those objects, the goal, the constraints on the plan and also the preferences and metric used to evaluate the plan. In the work we present in this paper, we limit the range of these changes. We start with an assumption that the user and system share the same collection of actions, with essentially the same pre- and post-conditions (we will discuss the slight qualification in Section~\ref{sec:plans}), the same collection of objects and the same goal. We also assume that the initial state given to the planner is essentially shared (again, we return to the qualification at a later point). Therefore, we focus on differences that arise from the constraints, preferences and plan metric in each of the models. Smith~\citeyear{smi12} argued that for mission planning, questions about plans often arise because of differences in preferences between the users and the planner. 

Smith~\citeyear{smi12} also observed that planning and explanation is an iterative process in which the user comes to understand and helps to improve a plan. We have taken inspiration from this idea in creating our framework for explanation as an iterative process described in Section~\ref{sec:iter}. Our framework allows a user to specify a sequence of tightening constraints to be applied to the original model. In this way, the user can {\em restrict} the original system model in an attempt to find a reconciliation between the tightened model and their own. We do not assume that the user maintains a static model throughout the process, so we acknowledge the possibility that the reconciliation might lead to an alignment between a restricted HModel and a modified, or more closely specified, user model. Furthermore, the outcome of the process is simply an explanation generated from a series of plans that satisfies the user in some sense, but that does not imply that the restricted models necessarily include a model that is aligned with the model the user holds. The user might conclude the process persuaded that their own plan is better than any plan the planner produces. Equally, even if a particular HModel leads to a plan that the user accepts, it is not necessarily the case that that HModel is the same as the user's model. We are only concerned with reconciling the models to the extent that the HModel responds satisfactorily to the specific planning problem under consideration. There is no generalisation of the restrictions added to the original model, so it cannot be assumed that the HModel would yield a satisfactory plan for a different initial state, or different goal.

\section{Plans: Queries and Explanations}\label{sec:plans}


In this section we provide the formal definitions that support our approach to explanation. We define the planning model and give a reference example, and then focus on the process of model restriction as a special case of model reconciliation, as described in Section~\ref{sec:back}.

\subsection{Formal Definition of a Planning Problem}
\label{sec:formalplan}

Our definition of a planning model follows the definition of PDDL2.1 given by \cite{fox03}, extended by a set of time windows and explicit record of the plan metric. 
The formal description of such a planning model is as follows.

\begin{definition}
\label{defn:planning-model}
A \textbf{planning model} is a pair $\Pi = \langle D, Prob \rangle$. The domain $D = \langle Ps, Vs, As, arity \rangle$ is a tuple where $Ps$ is a finite set of predicate symbols, $Vs$ is a finite set of function symbols, $As$ is a set of action schemas, called operators, and $arity$ is a function mapping all of these symbols to their respective arity. The problem $Prob = \langle Os, I, G, M, W \rangle$ is a tuple where $Os$ is the set of objects in the planning instance, $I$ is the initial state, $G$ is the goal condition, $M$ is a plan-metric function from plans to real values (plan costs) and $W$ is a set of time windows.
\end{definition}

A set of atomic \textit{propositions} $P$ is formed by applying the predicate symbols $Ps$ to the objects $Os$ (respecting arities). One proposition $p$ is formed by applying an ordered set of objects $o \subseteq O$ to one predicate $ps$, respecting its arity. For example, applying the predicate \textit{(robot\_at ?v - robot ?wp - waypoint)} with arity $2$ to the ordered set of objects $\{Jerry,sh3\}$ forms the proposition \textit{(robot\_at Jerry sh3)}. This process is called ``grounding'' and is denoted with:
$$ground(ps,\chi) = p$$
where $\chi\subseteq O$ is an ordered set of objects. 
Similarly the set of \textit{primitive numeric expressions} (PNEs) $V$ are formed by applying the function symbols $Vs$ to $Os$.

A state $s$ consists of a time $t\in\mathbb{R}$, a logical part $s_{l}\subseteq P$, and a numeric part $s_{v}$ that describes the values for the PNE's at that state. The initial state $I$ is the state at time $t=0$. 

The goal $G = g_1, ..., g_n$ is a set of constraints over $P$ and $V$ that must hold at the end of an action sequence for a plan to be valid. 
%
More specifically, for an action sequence $\pi = \langle a_1, a_2,\ldots,a_n \rangle$ each with a respective time denoted by $Dispatch(a_i)$, we use the definition of plan validity from \cite{fox03} (Definition 15 ``Validity of a Simple Plan''). A simple plan is the sequence of actions $\pi$ which defines a happening sequence, $t_{i=0\ldots k}$ and a sequence of states, $s_{i=0\ldots k+1}$ such that $s_0 = I$ and for each $i = 0\ldots k$, $s_{i+1}$ is the result of executing the happening at time $t_i$. The simple plan $\pi$ is valid if $s_{k+1}\models G$. 


The plan-metric function is, by default, the makespan of the plan to which it is applied. More generally, the metric assesses plan quality by taking into account both the extent to which a plan respects user preferences and also the costs associated with choices of action or combinations of actions within a plan. It is often the case that plans fail to meet expectations because of a mismatch in the way that plans are evaluated.

Each time window $w\in W$ is a tuple $w = \langle w_{lb}, w_{ub}, w_v \rangle$ where $w_v$ is a proposition which becomes true or a numeric effect which acts upon some $n \in V$. $w_{lb}\in\mathbb{R}$ is the time at which the proposition becomes true, or the numeric effect is applied. $w_{ub}\in\mathbb{R}$ is the time at which the proposition becomes false. The constraint $w_{lb} < w_{ub}$ must hold. Note that the numeric effect is not applied or reverted at $w_{ub}$, so $w_{ub}$ is superfluous for numeric effects.


Similar to propositions and PNEs, the set of ground actions $A$ is generated from the substitution of objects for operator parameters with respect to it's arity.
Each ground action is defined as follows:
\begin{definition}
A \textbf{ground action} $a \in A$ has a duration $Dur(a)$ which constrains the length of time that must pass between the start and end of $a$; a start (end) condition $Pre_{\vdash}(a)$ ($Pre_{\dashv}(a)$) which must hold at the state that $a$ starts (ends); an invariant condition $Pre_\leftrightarrow(a)$ which must hold throughout the entire execution of $a$; add effects $\Eff(a)_{\vdash}^+, \Eff(a)_{\dashv}^+ \subseteq P$ that are made true at the start and end of the action respectively; delete effects $\Eff(a)_{\vdash}^-, \Eff(a)_{\dashv}^- \subseteq P$ that are made false at the start and end of the action respectively; 
and numeric effects $\Eff(a)^n_{\vdash}$, $\Eff(a)^n_{\leftrightarrow}$, $\Eff(a)^n_{\dashv}$ that act upon some $n \in V$.
\end{definition}

\subsection{Running Example}\label{sec:running_example}

\begin{figure}[t!]
    \centering
    \includegraphics[width=0.9\linewidth]{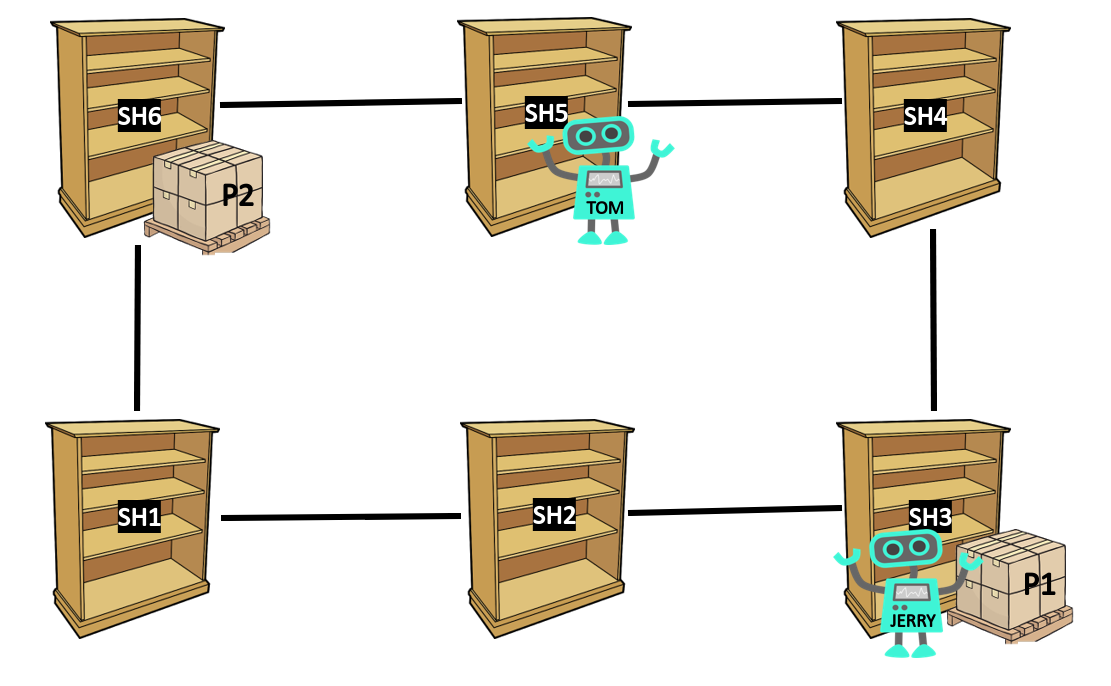}
    \caption{Diagram of the warehouse delivery system domain.}
    \label{fig:runningexample}
\end{figure}

As a reference example, we use a simplified version of a model of a warehouse delivery system. There are multiple robots that work to move pallets from their delivery location to the correct storage shelf. Before the pallets can be stored, the shelf must be set up.

Figure~\ref{fig:domain} defines the domain for this model.
There are four durative actions, $goto\_waypoint$, $set\_shelf$, $load\_pallet$, and $unload\_pallet$. The $goto\_waypoint$ action is used for the robots to navigate the factory. 
The $set\_shelf$ action ensures that the shelf is ready to store a package (the robot cannot perform this action while holding a pallet). The $load\_pallet$ action loads the pallet from a shelf on to the robot.  Finally, the $unload\_pallet$ action unloads the pallet onto a previously set shelf. 

For illustration purposes, we use a very simple problem with
two robots, two pallets, and six waypoints.
An example problem is shown in Figure~\ref{fig:problem}, and
an example plan for this planning problem is shown in Figure~\ref{fig:plan}. Figure~\ref{fig:plan} consists of a sequence of actions each with two attached values denoting the time they are executed and for how long. A diagram illustrating this domain is shown in Figure~\ref{fig:runningexample}.
For simplicity, we assume the cost of this plan is its duration (20.003) which in this case is optimal%
~\footnote{Optimal under PDDL 2.1 epsilon semantics with epsilon equal to .001. The plan is obtained using the planner POPF~\cite{col10}. However, our framework theoretically works with any PDDL2.1 planner.}.


\begin{figure}
\centering
\begin{minipage}{.48\textwidth}
\footnotesize
\centering
\begin{BVerbatim}
(:types
  waypoint robot - locatable
  pallet)
(:predicates
  (robot_at ?v - robot ?wp - waypoint)
  (connected ?from ?to - waypoint)
  (visited ?wp - waypoint)
  (not_occupied ?wp - waypoint)
  (set_shelf ?shelf - waypoint)
  (pallet_at ?p - pallet ?l - locatable)
  (not_holding_pallet ?v - robot))
(:functions
  (travel_time ?wp1 ?wp2 - waypoint))
(:durative-action goto_waypoint
  :parameters (?v - robot ?from ?to - waypoint)
  :duration(= ?duration (travel_time ?from ?to))
  :condition (and
    (at start (robot_at ?v ?from))
    (at start (not_occupied ?to))
    (over all (connected ?from ?to)))
  :effect (and
    (at start (not (not_occupied ?to)))
    (at end (not_occupied ?from))
    (at start (not (robot_at ?v ?from)))
    (at end (robot_at ?v ?to)))
)
(:durative-action set_shelf
  :parameters (?v - robot 
                ?shelf - waypoint)
  ...)
(:durative-action load_pallet
  :parameters (?v - robot ?p - pallet 
                ?shelf - waypoint)
  ...)
(:durative-action unload_pallet ...)
\end{BVerbatim}
\caption{A fragment of a robotics domain used as a running example. Some of the operator bodies have been omitted for space. The full description of the goto\_waypoint action is shown.}
\label{fig:domain}
\end{minipage}
\hspace{0.02\textwidth}
\begin{minipage}{.48\textwidth}
 \footnotesize
\begin{BVerbatim}
 (define (problem task)
 (:domain warehouse_domain)
 (:objects
     sh1 sh2 sh3 sh4 sh5 sh6 - waypoint
     p1 p2 - pallet
     Jerry Tom - robot)
 (:init
    (robot_at Jerry sh3) (robot_at Tom sh5)
    (not_holding_pallet Jerry) 
    (not_holding_pallet Tom)
    (not_occupied sh1) (not_occupied sh2)
    (not_occupied sh4) (not_occupied sh6)
    (pallet_at p1 sh3) (pallet_at p2 sh6)
    (connected sh1 sh2) (connected sh2 sh1)
    (connected sh2 sh3) (connected sh3 sh2)
    (connected sh3 sh4) (connected sh4 sh3)
    (connected sh4 sh5) (connected sh5 sh4)
    (connected sh5 sh6) (connected sh6 sh5)
    (connected sh6 sh1) (connected sh1 sh6)
    (= (travel_time sh1 sh2) 4)
    (= (travel_time sh2 sh1) 4)
    (= (travel_time sh2 sh3) 8)
    (= (travel_time sh3 sh2) 8)
    (= (travel_time sh3 sh4) 5)
    (= (travel_time sh4 sh3) 5)
    (= (travel_time sh4 sh5) 1)
    (= (travel_time sh5 sh4) 1)
    (= (travel_time sh5 sh6) 3)
    (= (travel_time sh6 sh5) 3)
    (= (travel_time sh6 sh1) 4)
    (= (travel_time sh1 sh6) 4)
)
(:goal (and
    (pallet_at p1 sh6)
    (pallet_at p2 sh1))))
\end{BVerbatim}
\caption{The problem instance used in the running example.}
\label{fig:problem}
\end{minipage}
\end{figure}

% ...

\begin{figure}[h!]
\centering
\begin{BVerbatim}
0.000: (goto_waypoint Tom sh5 sh6) [3.000]
0.000: (load_pallet Jerry p1 sh3) [2.000]
2.000: (goto_waypoint Jerry sh3 sh4) [5.000]
3.001: (set_shelf Tom sh6) [1.000]
4.001: (goto_waypoint Tom sh6 sh1) [4.000]
7.001: (goto_waypoint Jerry sh4 sh5) [1.000]
8.001: (set_shelf Tom sh1) [1.000]
8.002: (goto_waypoint Jerry sh5 sh6) [3.000]
9.001: (goto_waypoint Tom sh1 sh2) [4.000]
11.002: (unload_pallet Jerry p1 sh6) [1.500]
12.503: (load_pallet Jerry p2 sh6) [2.000]
14.503: (goto_waypoint Jerry sh6 sh1) [4.000]
18.503: (unload_pallet Jerry p2 sh1) [1.500]
\end{BVerbatim}
\caption{ The Plan with a cost of 20.003 generated from the example domain and problem.}
\label{fig:plan}
\end{figure}


Tying the reference example back to the definitions in Section~\ref{sec:formalplan}, the first action present in Figure~\ref{fig:plan} is the operator \textit{goto\_waypoint} in Figure~\ref{fig:domain} grounded with the objects \textit{\{Tom,sh5,sh6\}}. Each operator parameter is substituted with the corresponding object to give a ground action, this is represented in Figure~\ref{fig:groundaction} which shows the duration, conditions, and effects.

\begin{figure}[h!]
\centering
\begin{BVerbatim}
(:ground-action goto_waypoint Tom sh5 sh6
      :duration (= 3.000)
      :condition (and
            (at start (robot_at Tom sh5))
            (at start (not_occupied sh6))
            (over all (connected sh5 sh6)))
      :effect (and
            (at start (not (not_occupied sh6)))
            (at end (not_occupied sh5))
            (at start (not (robot_at Tom sh5)))
            (at end (robot_at Tom sh6)))
)
\end{BVerbatim}
\caption{The ground action \textit{(goto\_waypoint Tom sh5 sh6)}.}
\label{fig:groundaction}
\end{figure}

For ease of notation we allow access to multiple types of effects or preconditions through the ground action functions at once. For example for some ground action $a$, $\Eff^+$ denotes all add effects of $a$, $Pre_{\vdash \dashv}(a)$ denotes all start and end preconditions of $a$ but not invariant conditions, $\Eff(a)$ denotes all effects of $a$ including numeric effects.

\subsection{Plan Negotiation Problem}
\label{subsec:problem}


Fundamentally, the need for plan explanation is driven by the fact that a human and a planning agent may have different models of the planning problem and different computational capabilities. In Definition~\ref{defn:planning-model}  a planning model $\Pi$ was defined in terms of a domain $D = \langle Ps, Vs, As, arity \rangle$ and problem $Prob = \langle Os, I, G, M, W \rangle$. 
As mentioned in Section ~\ref{sec:back}, for purposes of this paper we assume that the human's planning model $\Pi^H$, and planning agent's model $\Pi^P$ share the same vocabulary, 
namely the same predicate symbols $Ps$, function symbols $Vs$, and actions $As$ from the domain $D$, 
and objects $Os$ from the problem.  
However, the action durations, conditions, and effects may be different, and the initial states $I$, goals $G$, and plan metric $M$ may be different. 
We do not assume that the human knows the planning agent's model $\Pi^P$, or vice versa. This assumption differs from previous work on model reconciliation~\cite{cha17} in that we do not assume that the planner knows (or learns) the planning model of the human.

Even when a human and a planning agent have the same planning models $\Pi^H=\Pi^P$, there are typically multiple plans satisfying this planning model. Although a planner is intended to optimise the plan with respect to the plan metric, it is common to produce only one of the valid plans, rather than an optimal plan for a model. A planner might even fail to produce a plan at all, for some problems. In part, this is an inevitable consequence of the undecidability of planning problems with numeric variables and functions~\cite{hel02}, but it is also a consequence of the practical limits on the computational resources available to a planner (time and memory). These observations are equally valid for automated and human planners. In order to discuss the process of developing plan explanations, it is helpful to define the planning abilities of both the planner and the user. 
We model the planning capability of an agent as a partial function from planning models to plans:

\begin{definition}\label{def:capability}
The {\bf planning capability} of an agent $A$ (human or machine), is a partial function, $\mathbb{C}^A$, from planning models to plans. Given the agents planning model, $\Pi^A$, if $\mathbb{C}^A(\Pi^A)$ is defined, then it is a candidate plan $\pi^A$ for the agent.
\end{definition}

The planning capability $\mathbb{C}^A$, can be affected by a multitude of factors. The part of the function domain on which $\mathbb{C}^A$ is defined determines the planning competency of the agent~-- domain-problem pairs for which the agent cannot find a plan lie outside this competency. Note that the planning competency of an agent can be restricted by a bound on the computational resources the agent is allowed to devote to the problem, as well as by the capabilities of the agent in constructing and adequately searching the search space that the problem defines.
When $A$ is an automated AI planner $P$, the computational ability is determined by the search strategy implemented in the planner, its heuristic (if there is one), and the resources allocated to the task. For sound planners, when $\mathbb{C}^P(\Pi^P)$ is defined it is a valid plan for $\Pi^P$.

When $A$ is a human planner $H$, the planning capability is determined by the understanding that the human has of the planning model and the patience and problem-solving effort they are willing to devote to solving the problem. It cannot be assumed that, if $\mathbb{C}^H(\Pi^H)$ is defined, that the human's model $\Pi^H$ accurately reflects the world, or that the reasoning $\mathbb{C}^H$ is sound. This means that the plan may not be valid. One aspect of the process of planning and explanation is that the user can revise their model $\Pi^H$ as the process unfolds. However, it is also possible that the user can change their planning capability $\mathbb{C}^H$, by coming to a greater understanding of the model, by engaging in more reasoning, or by simply concluding that the solution provided by an automated system is satisfactory. It is also possible that the planner responses lead to the user changing their view of what might be a good plan to solve a problem, while still not adopting the solution offered by the planner. Thus, the user's planning model and capability might be extended or modified by consideration of the planner output or question responses. This revision might include correcting flawed plans produced by the original planning model and capability of the user. 

In this paper, we do not explicitly attempt to model any learning process on the part of the human, although we allow that this may happen. 
Furthermore, we do not consider any learning by the planning agent. Instead, we adopt the approach that the human user asks contrastive questions that impose additional restrictions $\phi$ on the agent's planing problem $\Pi^P$ to generate a succession of hypothetical planning problems. The object of these questions and the resulting hypothetical plans is for the user to understand and ultimately arrive at a satisfactory plan. Model learning and reconciliation by the human and planning agent can be seen as complementary techniques that could make this process more effective and more efficient. 

Given the planning models $\Pi^H$ and $\Pi^P$, and planning capabilities $\mathbb{C}^H$ and $\mathbb{C}^P$  of a human and planning agent, the two agents disagree when $\mathbb{C}^H(\Pi^H) \not = \mathbb{C}^P(\Pi^P)$, which can arise in the case that either of these terms is undefined, or if both terms are defined and yield different plans. We assume that, in this case, the user is capable of inspecting the planner output and determining a question that will expose some part of the explanation for this difference. By questioning why certain decisions were made in the plan and receiving contrastive explanations the user can gain an initial understanding. As their understanding of the plan develops they can ask more educated questions to gain a deeper understanding or try to arrive at an alternative plan that they consider more satisfactory. Ultimately, this process concludes when the user is satisfied with some plan. In an ideal case, this will be when the user and the planner have converged on the same plan, but this need not happen. 
For example, suppose $\mathbb{C}^P(\Pi^P) = \pi$ and $\mathbb{C}^H(\Pi^H) = \pi'$ and $\pi \not = \pi'$. The user might inspect $\pi$ and, after seeking explanation for the differences between it and $\pi'$, conclude that there is some deficiency in the planner's model $\Pi^P$ or planner's reasoning $\mathbb{C}^P$ and therefore decide that $\pi'$ is the plan they want. Thus, the sequence, in this case, might conclude with the user rejecting the plan offered by the planner and not changing their own model or computational ability at all.

We formalise the iterative process of questioning and explanation as one of successive {\em model restriction}, in which the user asks contrastive questions in an attempt to understand the planning agent's plan and potentially steer the planning agent towards a satisfactory solution. 
We suppose that, when $\mathbb{C}^H(\Pi^H) \not = \mathbb{C}^P(\Pi^P)$, the user can construct some {\em foil}, $\phi$, in the form of a constraint that $\mathbb{C}^P(\Pi^P)$ does not satisfy, so that seeking an explanation for the plan, $\mathbb{C}^P(\Pi^P)$, can be seen as seeking a plan for $\Pi^P$ that also satisfies $\phi$. This requirement acts as a restriction on $\Pi^P$ and is captured as follows.

\begin{definition}\label{def:constraintaddition}
A {\bf constraint property} is a predicate, $\phi$, over plans.

A {\bf constraint operator}, $\times$ is defined so that, for a planning model $\Pi$ and any constraint property $\phi$, $\Pi \times \phi$ is a model (an {\em HModel}), $\Pi'$, called a {\bf model restriction} of $\Pi$, satisfying the condition that any plan for $\Pi'$ is a plan for $\Pi$ that also satisfies $\phi$. A plan for an HModel is refered to as an {\em HPlan}.
\end{definition}

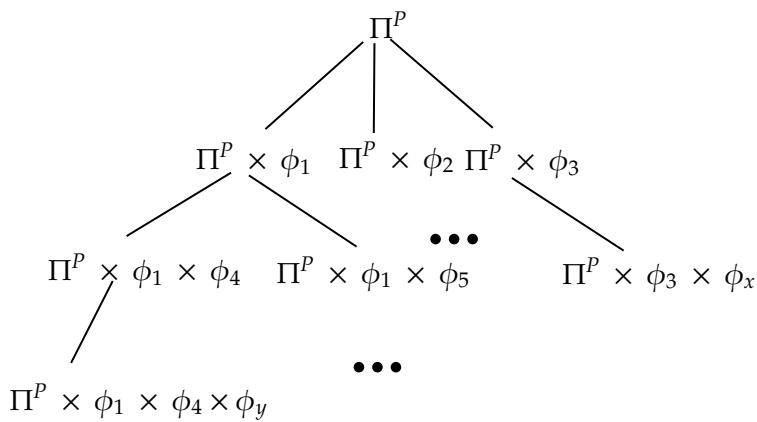
\begin{figure}
    \centering
\tikzset{every picture/.style={line width=0.75pt}} 
\begin{tikzpicture}[x=0.5pt,y=0.5pt,yscale=-1,xscale=1]

\draw    (331,52) -- (408.5,119) ;
\draw    (310.5,54) -- (236.5,120) ;
\draw    (319,55) -- (318.5,123) ;
\draw    (210.5,153) -- (131.5,201) ;
\draw    (224,155) -- (305.5,209) ;
\draw  [fill={rgb, 255:red, 0; green, 0; blue, 0 }  ,fill opacity=1 ] (362,203.25) .. controls (362,201.46) and (363.46,200) .. (365.25,200) .. controls (367.04,200) and (368.5,201.46) .. (368.5,203.25) .. controls (368.5,205.04) and (367.04,206.5) .. (365.25,206.5) .. controls (363.46,206.5) and (362,205.04) .. (362,203.25) -- cycle ;
\draw  [fill={rgb, 255:red, 0; green, 0; blue, 0 }  ,fill opacity=1 ] (376,203.25) .. controls (376,201.46) and (377.46,200) .. (379.25,200) .. controls (381.04,200) and (382.5,201.46) .. (382.5,203.25) .. controls (382.5,205.04) and (381.04,206.5) .. (379.25,206.5) .. controls (377.46,206.5) and (376,205.04) .. (376,203.25) -- cycle ;
\draw  [fill={rgb, 255:red, 0; green, 0; blue, 0 }  ,fill opacity=1 ] (390,203.25) .. controls (390,201.46) and (391.46,200) .. (393.25,200) .. controls (395.04,200) and (396.5,201.46) .. (396.5,203.25) .. controls (396.5,205.04) and (395.04,206.5) .. (393.25,206.5) .. controls (391.46,206.5) and (390,205.04) .. (390,203.25) -- cycle ;
\draw    (423,157) -- (507.5,212) ;
\draw    (121,235) -- (89.5,297) ;
\draw  [fill={rgb, 255:red, 0; green, 0; blue, 0 }  ,fill opacity=1 ] (304,300.25) .. controls (304,298.46) and (305.46,297) .. (307.25,297) .. controls (309.04,297) and (310.5,298.46) .. (310.5,300.25) .. controls (310.5,302.04) and (309.04,303.5) .. (307.25,303.5) .. controls (305.46,303.5) and (304,302.04) .. (304,300.25) -- cycle ;
\draw  [fill={rgb, 255:red, 0; green, 0; blue, 0 }  ,fill opacity=1 ] (318,300.25) .. controls (318,298.46) and (319.46,297) .. (321.25,297) .. controls (323.04,297) and (324.5,298.46) .. (324.5,300.25) .. controls (324.5,302.04) and (323.04,303.5) .. (321.25,303.5) .. controls (319.46,303.5) and (318,302.04) .. (318,300.25) -- cycle ;
\draw  [fill={rgb, 255:red, 0; green, 0; blue, 0 }  ,fill opacity=1 ] (332,300.25) .. controls (332,298.46) and (333.46,297) .. (335.25,297) .. controls (337.04,297) and (338.5,298.46) .. (338.5,300.25) .. controls (338.5,302.04) and (337.04,303.5) .. (335.25,303.5) .. controls (333.46,303.5) and (332,302.04) .. (332,300.25) -- cycle ;

\draw (313,29) node [anchor=north west][inner sep=0.75pt]   [align=left] {$\displaystyle \Pi^P $};
\draw (182,128) node [anchor=north west][inner sep=0.75pt]   [align=left] {$\displaystyle \Pi^P \ \times \ \phi _{1}$};
\draw (290,127) node [anchor=north west][inner sep=0.75pt]   [align=left] {$\displaystyle \Pi^P \ \times \ \phi _{2}$};
\draw (385,129) node [anchor=north west][inner sep=0.75pt]   [align=left] {$\displaystyle \Pi^P \ \times \ \phi _{3}$};
\draw (70,212) node [anchor=north west][inner sep=0.75pt]   [align=left] {$\displaystyle \Pi^P \ \times \ \phi _{1} \ \times \ \phi _{4}$};
\draw (243,214) node [anchor=north west][inner sep=0.75pt]   [align=left] {$\displaystyle \Pi^P \ \times \ \phi _{1} \ \times \ \phi _{5}$};
\draw (459,215) node [anchor=north west][inner sep=0.75pt]   [align=left] {$\displaystyle \Pi^P \ \times \ \phi _{3} \ \times \ \phi _{x}$};
\draw (41,310) node [anchor=north west][inner sep=0.75pt]   [align=left] {$\displaystyle \Pi^P \ \times \ \phi _{1} \ \times \ \phi _{4} \times \phi _{y}$};

\end{tikzpicture}
    \caption{A fragment of a tree of model restrictions for a planner $P$.}
    \label{fig:tree}
\end{figure}

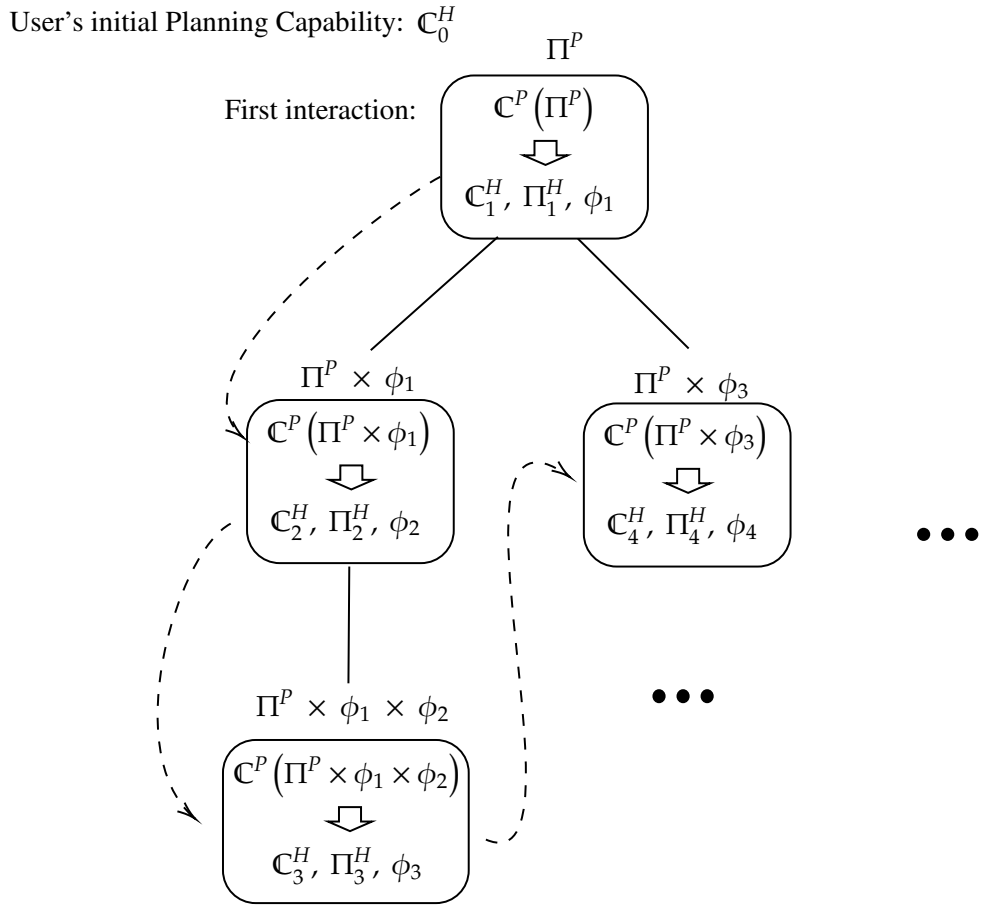
\begin{figure}
    \centering

\tikzset{every picture/.style={line width=0.75pt}} 

\begin{tikzpicture}[x=0.65pt,y=0.65pt,yscale=-1,xscale=1]

\draw    (284.5,147) -- (210.5,213) ;
\draw    (198,339) -- (197.5,407) ;
\draw  [fill={rgb, 255:red, 0; green, 0; blue, 0 }  ,fill opacity=1 ] (375,414.5) .. controls (375,416.43) and (376.46,418) .. (378.25,418) .. controls (380.04,418) and (381.5,416.43) .. (381.5,414.5) .. controls (381.5,412.57) and (380.04,411) .. (378.25,411) .. controls (376.46,411) and (375,412.57) .. (375,414.5) -- cycle ;
\draw  [fill={rgb, 255:red, 0; green, 0; blue, 0 }  ,fill opacity=1 ] (389,414.5) .. controls (389,416.43) and (390.46,418) .. (392.25,418) .. controls (394.04,418) and (395.5,416.43) .. (395.5,414.5) .. controls (395.5,412.57) and (394.04,411) .. (392.25,411) .. controls (390.46,411) and (389,412.57) .. (389,414.5) -- cycle ;
\draw  [fill={rgb, 255:red, 0; green, 0; blue, 0 }  ,fill opacity=1 ] (403,414.5) .. controls (403,416.43) and (404.46,418) .. (406.25,418) .. controls (408.04,418) and (409.5,416.43) .. (409.5,414.5) .. controls (409.5,412.57) and (408.04,411) .. (406.25,411) .. controls (404.46,411) and (403,412.57) .. (403,414.5) -- cycle ;

\draw    (331,148) -- (395.5,212) ;
\draw  [fill={rgb, 255:red, 0; green, 0; blue, 0 }  ,fill opacity=1 ] (529,320.25) .. controls (529,318.46) and (530.46,317) .. (532.25,317) .. controls (534.04,317) and (535.5,318.46) .. (535.5,320.25) .. controls (535.5,322.04) and (534.04,323.5) .. (532.25,323.5) .. controls (530.46,323.5) and (529,322.04) .. (529,320.25) -- cycle ;
\draw  [fill={rgb, 255:red, 0; green, 0; blue, 0 }  ,fill opacity=1 ] (543,320.25) .. controls (543,318.46) and (544.46,317) .. (546.25,317) .. controls (548.04,317) and (549.5,318.46) .. (549.5,320.25) .. controls (549.5,322.04) and (548.04,323.5) .. (546.25,323.5) .. controls (544.46,323.5) and (543,322.04) .. (543,320.25) -- cycle ;
\draw  [fill={rgb, 255:red, 0; green, 0; blue, 0 }  ,fill opacity=1 ] (557,320.25) .. controls (557,318.46) and (558.46,317) .. (560.25,317) .. controls (562.04,317) and (563.5,318.46) .. (563.5,320.25) .. controls (563.5,322.04) and (562.04,323.5) .. (560.25,323.5) .. controls (558.46,323.5) and (557,322.04) .. (557,320.25) -- cycle ;

\draw   (251.5,72) .. controls (251.5,61.51) and (260.01,53) .. (270.5,53) -- (352.79,53) .. controls (363.29,53) and (371.79,61.51) .. (371.79,72) -- (371.79,129) .. controls (371.79,139.49) and (363.29,148) .. (352.79,148) -- (270.5,148) .. controls (260.01,148) and (251.5,139.49) .. (251.5,129) -- cycle ;
\draw   (299.36,99.5) -- (305.78,99.5) -- (305.78,91.1) -- (318.62,91.1) -- (318.62,99.5) -- (325.03,99.5) -- (312.2,105.1) -- cycle ;

\draw   (124.5,459) .. controls (124.5,448.51) and (133.01,440) .. (143.5,440) -- (248.5,440) .. controls (258.99,440) and (267.5,448.51) .. (267.5,459) -- (267.5,516) .. controls (267.5,526.49) and (258.99,535) .. (248.5,535) -- (143.5,535) .. controls (133.01,535) and (124.5,526.49) .. (124.5,516) -- cycle ;
\draw  [dash pattern={on 4.5pt off 4.5pt}]  (251,112) .. controls (220.31,127.84) and (93.08,224.05) .. (135.17,262.84) ;
\draw [shift={(136.5,264)}, rotate = 219.56] [color={rgb, 255:red, 0; green, 0; blue, 0 }  ][line width=0.75]    (10.93,-3.29) .. controls (6.95,-1.4) and (3.31,-0.3) .. (0,0) .. controls (3.31,0.3) and (6.95,1.4) .. (10.93,3.29)   ;
\draw  [dash pattern={on 4.5pt off 4.5pt}]  (129,314) .. controls (98.31,329.84) and (63.21,445.65) .. (107.14,484.84) ;
\draw [shift={(108.5,486)}, rotate = 219.56] [color={rgb, 255:red, 0; green, 0; blue, 0 }  ][line width=0.75]    (10.93,-3.29) .. controls (6.95,-1.4) and (3.31,-0.3) .. (0,0) .. controls (3.31,0.3) and (6.95,1.4) .. (10.93,3.29)   ;
\draw  [dash pattern={on 4.5pt off 4.5pt}]  (277.5,498) .. controls (342.18,525.86) and (242.51,227.01) .. (325.24,286.08) ;
\draw [shift={(326.5,287)}, rotate = 216.54] [color={rgb, 255:red, 0; green, 0; blue, 0 }  ][line width=0.75]    (10.93,-3.29) .. controls (6.95,-1.4) and (3.31,-0.3) .. (0,0) .. controls (3.31,0.3) and (6.95,1.4) .. (10.93,3.29)   ;
\draw   (138.5,261) .. controls (138.5,250.51) and (147.01,242) .. (157.5,242) -- (239.79,242) .. controls (250.29,242) and (258.79,250.51) .. (258.79,261) -- (258.79,318) .. controls (258.79,328.49) and (250.29,337) .. (239.79,337) -- (157.5,337) .. controls (147.01,337) and (138.5,328.49) .. (138.5,318) -- cycle ;
\draw   (186.36,288.5) -- (192.78,288.5) -- (192.78,280.1) -- (205.62,280.1) -- (205.62,288.5) -- (212.03,288.5) -- (199.2,294.1) -- cycle ;

\draw   (334.5,263) .. controls (334.5,252.51) and (343.01,244) .. (353.5,244) -- (435.79,244) .. controls (446.29,244) and (454.79,252.51) .. (454.79,263) -- (454.79,320) .. controls (454.79,330.49) and (446.29,339) .. (435.79,339) -- (353.5,339) .. controls (343.01,339) and (334.5,330.49) .. (334.5,320) -- cycle ;
\draw   (382.36,290.5) -- (388.78,290.5) -- (388.78,282.1) -- (401.62,282.1) -- (401.62,290.5) -- (408.03,290.5) -- (395.2,296.1) -- cycle ;

\draw   (185.58,487.5) -- (192,487.5) -- (192,479.1) -- (204.84,479.1) -- (204.84,487.5) -- (211.25,487.5) -- (198.42,493.1) -- cycle ;

\draw (311,26) node [anchor=north west][inner sep=0.75pt]   [align=left] {$\displaystyle \Pi ^{P}$};
\draw (168,217) node [anchor=north west][inner sep=0.75pt]   [align=left] {$\displaystyle \Pi ^{P} \ \times \ \phi _{1}$};
\draw (362,221) node [anchor=north west][inner sep=0.75pt]   [align=left] {$\displaystyle \Pi ^{P} \ \times \ \phi _{3}$};
\draw (142,409) node [anchor=north west][inner sep=0.75pt]   [align=left] {$\displaystyle \Pi ^{P} \ \times \ \phi _{1} \ \times \ \phi _{2}$};
\draw (236,11) node [anchor=north west][inner sep=0.75pt]   [align=left] (initial_computational_ability) {$\displaystyle \mathbb{C}^{H}_{0}$};
\draw (22,13) node [anchor=north west][inner sep=0.75pt] [left = 3pt of initial_computational_ability]  [align=left] {{\fontfamily{ptm}\selectfont User's initial Planning Capability}:};
\draw (129,446) node [anchor=north west][inner sep=0.75pt]   [align=left] {$\displaystyle \mathbb{C}^{P}\left( \Pi ^{P} \times \phi _{1} \times \phi _{2}\right)$};
\draw (151.62,501) node [anchor=north west][inner sep=0.75pt]   [align=left] {$\displaystyle \mathbb{C}^{H}_{3} ,\ \Pi ^{H}_{3} ,\ \phi _{3}$};
\draw (344.65,249) node [anchor=north west][inner sep=0.75pt]   [align=left] {$\displaystyle \mathbb{C}^{P}\left( \Pi ^{P} \times \phi _{3}\right)$};
\draw (346.62,303) node [anchor=north west][inner sep=0.75pt]   [align=left] {$\displaystyle \mathbb{C}^{H}_{4} ,\ \Pi ^{H}_{4} ,\ \phi _{4}$};
\draw (148.65,247) node [anchor=north west][inner sep=0.75pt]   [align=left] {$\displaystyle \mathbb{C}^{P}\left( \Pi ^{P} \times \phi _{1}\right)$};
\draw (150.62,301) node [anchor=north west][inner sep=0.75pt]   [align=left] {$\displaystyle \mathbb{C}^{H}_{2} ,\ \Pi ^{H}_{2} ,\ \phi _{2}$};
\draw (263.62,112) node [anchor=north west][inner sep=0.75pt]   [align=left] {$\displaystyle \mathbb{C}^{H}_{1} ,\ \Pi ^{H}_{1} ,\ \phi _{1}$};
\draw (280.65,58) node [anchor=north west][inner sep=0.75pt]   [align=left] (initial_planner_model) {$\displaystyle \mathbb{C}^{P}\left( \Pi ^{P}\right)$};
\draw (140,93) node [anchor=north west][inner sep=0.75pt] [left = 27pt of initial_planner_model] [align=left] {{\fontfamily{ptm}\selectfont First interaction:}};

\end{tikzpicture}

    \caption{An example of a sequence of interactions between a user and a planner. At each interaction the user updates their planning model and capability, identifies a new constraint, which may or may not incorporate previous constraints. The order in which nodes are explored, as indicated by the dotted line, is entirely under the control of the user. }
    \label{fig:tree2}
\end{figure}

The process in which the user interacts with a planner is an iterative one -- the user successively views plans and seeks explanations by generating foils that impose additional restrictions on the planning problem. The collection of model restrictions forms a tree, rooted at the original model and extended by the incremental addition of new constraint properties, as shown in Figure~\ref{fig:tree}. The user can visit the nodes of this tree in any order. As the user inspects the result of applying $\mathbb{C}^P$ to a node in this tree, their own planning model and capability, $\Pi^H$ and $\mathbb{C}^H$,  may change, reflecting accumulating understanding of the plans that can be constructed for the model.  As a result, the order in which the user visits the nodes matters and can lead to different outcomes. One possible path, showing the evolving capability and model for the user, is shown in Figure~\ref{fig:tree2}. This figure should not be interpreted as implying that the user must explore the tree in a systematic way. It is also worth emphasising that any constraint, $\phi$, may be added to any model, so that the user is not forced to develop a tree of models in any particular way to arrive at the consequence of adding any specific constraint to a model.

We call what we have described above the \textit{Plan Negotiation Problem}. In this problem the user and the planning system must negotiate, through the planning process, to produce an acceptable plan. This problem can arise for many reasons. In the case where a user has an expectation of what the plan should look like that differs from the proposed plan, the user may not accept the proposed plan without understanding why it was produced, or exploring other plan options. A user might be unsure of the quality of the plan but not have the reasoning abilities to properly evaluate the plan quality. The user can explore how alternate actions and decisions that could have been made in the plan affects the plan quality. This will either refute or support their concerns, that either the plan they were presented was of good quality or that there is a plan with better quality.
If a better plan cannot be found under the added constraint, this might allay their concerns, while, if a better plan is found, it will confirm the user's suspicions. In either case, the user might go on to explore additional constraints, in search of a better plan, or of better understanding of the plan space.
A user might have hidden preferences that are not modelled, and through the addition of constraints can make sure that the plan behaves in such a way that their preferences are fulfilled. Or the user might simply intend to increase their understanding of the model by questioning why certain decisions were made in the plan before being willing to accept it.

In each of these cases, reasoning about what did not happen in the plan can give a deeper understanding of the decisions made in the plan and simultaneously explore potentially more suitable plan candidates. The user is offered the opportunity to consider what did not happen in a plan (in particular, why a plan does not satisfy some constraint), by asking the contrastive question ``why is the plan for this model as it is and not one that also satisfies the constraint $\phi$?". 
As indicated earlier, we require that the user and the system share the same vocabulary. This qualification ensures that any user restriction $\phi$ can actually be understood by the planner -- i.e. that the model $\Pi^P \time \phi$ makes sense. As a result,  the user can restrict a model in a way that prevents the planner from using an action in states where some condition is not satisfied, effectively adding a precondition to that action. Similarly, the model can be restricted to prevent the planner exploiting an effect of an action, by constraining the actions that can be applied after the particular action. Although this process will not allow the user to add arbitrary preconditions or eliminate arbitrary effects (since the states that are generated remain faithful to the model the planner is actually using), this observation makes the point that the model restrictions can include close approximations to model revisions that act directly on the actions themselves.

An example that illustrates a fragment of the negotiation process is as follows. Using the model $\Pi$ shown in Figures~\ref{fig:domain} and~\ref{fig:problem} and the plan $\pi$ shown in Figure~\ref{fig:plan}, the user might think that the action \textit{(goto\_waypoint Jerry sh4 sh5)} should not be present in the plan ($\phi$), so $\Pi \times \phi = \Pi'$ where the plan $\pi'$ for $\Pi'$ does not contain \textit{(goto\_waypoint Jerry sh4 sh5)}. The user either needs an explanation that will support acceptance of the original plan $\pi$ (by modifying the user's model or planning capability to make this plan acceptable), or the constraint, $\phi$, will guide the search of the planner to a plan $\pi'$ where $M(\pi') > M(\pi)$.

The new plan $\pi'$ might not entirely reconcile the user's concerns. It might trigger new questions or still not satisfy the user's expectations. The user can explore the space of plans by iteratively extending and specifying the foil $\phi$, until they are satisfied with the result. 

It should be noted that, depending on the planning models and capabilities of the two participants, there might not exist any constraint achieving a common solution. For example, in the degenerate case in which $\mathbb{C}^P$ produces no plan at all, for any value of $\phi$, then there can be no negotiated common plan. Typically, the greater the differences between the planning models and capabilities of the two agents, the more likely it will be that there is no common satisfactory solution. 


We formally capture the iterative process of model restriction and planning as:
\begin{definition}\label{def:negprob}
\textbf{Iterative Model Restriction} For a planner $P$, and a user $H$: Let $\mathbb{C}^P$ and $\Pi^P$  be the planner's underlying capability and planning model and $\mathbb{C}^H_0$ and $\Pi^H_0$ be the initial capability and planning model of $H$. 
Let $\phi_i$ be the set of user imposed constraints, which is initially empty, i.e. $\phi_0 = \emptyset$.
Each stage, $i$ (initially zero), of this process starts with the planner producing a plan $\pi^P_i = \mathbb{C}^P(\Pi^P_i)$ for the model $\Pi^P_i = \Pi^P \times \phi_i$. 

The user responds to this plan $\pi^P_i$ by potentially updating their capability and model to $\mathbb{C}^H_{i+1}$ and $Pi^H_{i+1}$ and then either terminating the interaction, or asking a question that imposes a new constraint $\phi_{i+1}$ on the problem.  This results in the planner solving a new constrained problem $\Pi^P_{i+1} = \Pi^P \times \phi_{i+1}$ at the next step . 
\end{definition}

Although it is possible that the planner will fail to produce a plan at some stage, $i$, we do not address the problem of explaining the unsolvability of plans in this paper~\cite{bel10,sre19}. Nevertheless, the failure will be observed by the user and it can trigger a decision to either select a previous plan $\pi^P_j$ for some $j<i$, or explore a new constraint $\phi_{i+1}$ for the next iteration.

We have assumed here that the planners underlying capability and planning model $\mathbb{C}^P$ and $\Pi^P$ do not evolve during the process. While this is not strictly necessary, possible evolution or improvement of the planner capabilities and model based on the sequence of user questions and the resulting $\phi_i$ is an issue we do not consider here.  In contrast, the user's capability and planning model $\mathbb{C}^H$ and $\Pi^H$ are assumed to evolve, but in unknown ways.  Again, we do not attempt to model the user's learning process.

\subsection{Ending Negotiation}

The process we have described is one in which a user explores a tree of model restrictions, rooted at the original model. At each node in the tree the planner will produce some output (although possibly no plan) and the user will revise their personal planning model and capability. This revision might be trivial, in that the user might simply retain the model and capability they held at the previous iteration. The model revisions need not converge in any sense, but at some point the negotiation will end. We now briefly consider the status of the negotiation at the conclusion of an interaction.  

One way that the negotiation can end is that the plan produced for the final model yields a plan that is acceptable to the user, so that the user adopts this plan for the original model. This is a case where the system and the user converge on a plan that is mutually agreed to be a solution to the original model, meeting conditions that might or might not have been part of the original model and that the user might or might not have envisaged at the outset of the negotiation. 

Another way the negotiation can end is with the user having explored the plans for several models and, finally, having been persuaded in this process that the first plan produced by the planner for the original model is actually the desired plan, the user modifies their planning model and capability so that this is a plan for the original model of the planner and revised model of the user. Again, this is a mutually agreed plan, but in this case it is not the last plan produced, but the first; the negotiation process in this case acts to help the user to arrive at a point where they are persuaded that it is the plan that they want. In contrast to the first case, where the user might not ever modify their planning model or capability, in this second case the user must modify their planning model and capability to accept the plan for the original planner model. This process is the idealised form of plan explanation we anticipate: the user explores the plans for restricted models in order to understand why the original plan is the correct plan for the problem and they adapt their own planning model and capability to reflect this conclusion.

The negotiation can also lie somewhere between these two variants, with the user concluding the negotiation after adopting a plan produced for some intermediate model in the negotiation, modifying their planning model and capability to include this mutually agreed plan. 

A final outcome is one in which the user explores the space and then rejects all of the plans the planner offers. In this case, the user might modify their planning model and capability as a consequence of what they observe and they might or might not conclude the process with a satisfactory plan for the original model. In this case, there is no mutually agreed plan and the negotiation might not even have helped the user arrive at any useful conclusions about the problem.

Despite the fact that all of these outcomes are possible, it is impossible to determine, from the perspective of the system, which of them has been achieved at the end of a negotiation. The system has no access to the planning model or capability of the user and does not construct queries to probe it. The hypothesis we explore, in the user study we describe in Section~\ref{sec:userstudy}, is that the user will usually find value in the negotiation and conclude in one of the three cases in which a mutually agreed plan is identified. As can be seen, it remains impossible to be sure which plan is the mutually agreed plan at the conclusion of the negotiation. 


\section{Model-Based Compilations}
\label{sec:comp}

Armed with a formal description of the interactive process of model refinement that underpins the construction of our explanations, we now consider how the system can generate plans for the series of models generated in the process. In particular, given a planning model $\Pi$ and a constraint $\phi$, we aim to construct a plan for $\Pi \times \phi$. The approach we adopt is to {\em compile} the constraint $\phi$ into the model $\Pi$, so that $\Pi \times \phi$ can be presented to a generic planner as another model to be solved. This approach avoids embedding the iterative process inside a planner, instead using a planner as a service inside the process of iterative model refinement.

Although the point was not explicitly addressed in Definitions~\ref{def:constraintaddition} and~\ref{def:negprob}, it is not necessarily possible to combine an arbitrary constraint, $\phi$ with a model $\Pi$ to yield a model $\Pi \times \phi$ that is expressible in the language we use to describe our planning models (Definition~\ref{defn:planning-model}). The compilation strategy exploits the case in which $\Pi \times \phi$ {\em can} be expressed in our modelling language and, in this section, we demonstrate how this is achieved for a collection of different forms for $\phi$. In the case where the user wishes to capture some constraint that cannot be captured in this way, it is often possible to incrementally converge on a model restriction that approximates the constraint, by the addition of constraints that can be expressed and that steadily remove parts of the plan space that violate the intended constraint. This process is discussed further in Section~\ref{comp:composition}, and is analogous to the addition of cuts to a linear program in order to find a solution to an integer program. The constraints in this section, for which we present compilations, were chosen in response to the user study presented in Section~\ref{sec:taxonomy} and are examples of real questions for which users sought explanations.
 
The addition of a constraint to a model never increases the collection of feasible solutions, and so might make the search for a solution harder. There are two reasons that this intuition might not match observations. First, let us consider the construction of feasible solutions by an incremental series of choices to variables (such as actions added to the head of a developing plan, as in forward search planning). The addition of constraints will prune the collection of feasible solutions in this space, but it can also prune early partial solutions that were previously feasible, but for which there were no extensions into complete feasible solutions. That is, the constraints can act to prune partial solutions that previously appeared promising, leading to a reduction in search in that part of the space.  Secondly, where solutions are constructed by search, the addition of features to the model can lead to choices being explored in a different order, possibly for entirely implementation-dependent reasons (such as reordering of action choices inside an internal data-structure, based on order of grounding). These changes can lead to unpredictable effects on the performance of a planner, possibly leading to a lucky reduction of search or an unlucky increase in search. These effects will be observed in all search-based solvers and different families of constraints might interact with the solution strategy of specific planners in different ways. For example, adding timed-effects to the initial conditions of a problem for {\sc popf}~\cite{popf} can create additional choice branches at every step in the construction of a plan. In Section~\ref{sec:eval} we explore the effects of the compilations on performance for a range of representative examples.

\subsection{Explanation Problem}

\begin{definition}
An \textbf{explanation problem} is a tuple $E = \langle \Pi, \pi, Q\rangle$, in which $\Pi$ is a planning model (Definition~\ref{defn:planning-model}), $\pi$ is the plan generated by the planner, and $Q$ is the specific question posed by the user.
\end{definition}

We are interested when the user question $Q$ is a 
\textit{contrastive question} of the form \textit{``Why A rather than B?''}, where A occurred in the plan and B is the hypothetical alternative expected by the user. 
This question can be captured as a constraint that enforces the \textit{foil}. A foil is normally partial -- i.e. a set of additional constraints on the form of the solution rather than being a complete alternative. This fits with the framing of this entire process as being one of iterative model restriction.

As in our user study, we assume that the user knows the model $\Pi$ and the plan $\pi$, so responses such as stating the goal of the problem will not increase their understanding. Based on the outcome of the user study, we provide a formal description for compilations of the questions in the Contrastive Taxonomy (Table~\ref{table0}), reiterated here:
\begin{itemize}
\item  \textbf{FQ1} - Why is action $a$ not used in the plan, rather than being used? (Section~\ref{comp:add})
\item \textbf{FQ2} - Why is action $a$ used in the plan, rather than not being used? (Section~\ref{comp:remove})
\item  \textbf{FQ3} - Why is action $a$ used in state $s$, rather than action $b$? (Section~\ref{comp:replace})
\item  \textbf{FQ4} - Why is action $a$ not performed before (after) action $b$, rather than $a$ being performed after (before) $b$? (Section~\ref{comp:reschedule})
\item  \textbf{FQ5} - Why is action $a$ used outside of time window $w$, rather than only being allowed within $w$? (Section~\ref{comp:tw1})
\item \textbf{FQ6} - Why is action $a$ not used in time window $w$, rather than being used within $w$? (Section~\ref{comp:tw2})
\item  \textbf{FQ7} - Why is action $a$ used at time $t$, rather than at least some time $t'$ after/before $t$? (Section~\ref{comp:delay})
\end{itemize}
This section formalises the compilations of the questions in the Contrastive Taxonomy to produce an HModel $\Pi' = \Pi\times\phi$, where $\phi$ is a constraint derived from \textit{Q} and $\Pi$ is a PDDL2.1 model~\cite{fox03}.
The HModel $\Pi'$ is:
$$
\Pi' = \langle \langle Ps', Vs, As', arity' \rangle, \langle Os, I', G', W'\rangle \rangle
$$



After the HModel is formed, it is solved to give the HPlan. Any new operators that are used in the compilation to enforce some constraint are trivially renamed to the original operators they represent. For each iteration of compilation the HPlan is validated against the original model $\Pi$.


\subsection{Add an Action to the Plan}\label{comp:add}

Given a plan $\pi$, a formal question $Q$ is asked of the form:
\begin{quote}
\textit{Why is the operator $o$ with parameters $\chi$ not used, rather than being used?}
\end{quote}
For example, given the example plan in Figure~\ref{fig:plan} the user might ask:
\begin{quote}
``Why is \textit{(load\_pallet Tom p2 sh6)} not used, rather than being used?''
\end{quote}
They might ask this because a goal of the problem is to load and move the pallet \textit{p2} to shelf \textit{sh1}. As the robot \textit{Tom} moves to shelf \textit{sh6} where the pallet \textit{p2} is located early in the plan, and the pallet \textit{p2} is located at \textit{sh6} and the shelves \textit{sh6} and \textit{sh1} are connected, it might make sense to the user for the robot \textit{Tom} to deliver this pallet.

To generate the HPlan, a compilation is formed such that the action $a=ground(o,\chi)$ must be applied for the plan to be valid. The compilation introduces a new predicate $has\_done\_a$, which represents which actions have been applied. Using this, the goal is extended to include that the user suggested action has been applied.
The HModel $\Pi'$ is:
$$
\Pi' = \langle \langle Ps', Vs, As', arity' \rangle, \langle Os, I, G', W\rangle \rangle
$$
where
\begin{itemize}
    \item $Ps' = Ps \cup \{has\_done\_a\}$
    \item $As' = \{o_a\} \cup As \setminus \{o\}$
    \item $arity'(x) = arity(x),\, \forall x\in arity$
    \item $arity'(has\_done\_a) = arity'(o_a) = arity(o)$
    \item $G' = G \cup \{ground(has\_done\_a,\chi)\}$
\end{itemize}
where the new operator $o_a$ extends $o$ with the add effect $has\_done\_a$ with corresponding parameters, i.e.
$$
\Eff^+_{\dashv}(o_a)=\Eff^+_{\dashv}(o) \cup \{has\_done\_a\}
$$

For example given the user question above where $a = ground(load\_pallet,\{Tom, p2, sh6\})$, the operator \textit{load\_pallet} from the running example is extended to \textit{load\_pallet\_prime} with the additional add effect \textit{has\_done\_load\_pallet}. The new operator is shown in the PDDL2.1 syntax in Figure~\ref{fig:loadpalletprime}.

\begin{figure}[h!]
\begin{lstlisting}[xleftmargin=.15\textwidth]
(:durative-action load_pallet_prime
  :parameters (?v - robot ?p - pallet ?shelf - waypoint)
  :duration(= ?duration 2)
  :condition (and 
    (over all (robot_at ?v ?shelf))
    (at start (pallet_at ?p ?shelf))
    (at start (not_holding_pallet ?v)))
  :effect (and
    (at start (not (pallet_at ?p ?shelf)))
    (at start (not (not_holding_pallet ?v)))
    (at end (pallet_at ?p ?v))
    @\textbf{(at end (has\_done\_load\_pallet ?v ?p ?shelf)))}@
)
\end{lstlisting}
\caption{The operator \textit{goto\_waypoint\_prime} which extends the original operator \textit{load\_pallet} with the new add effect \textit{has\_done\_load\_pallet}.}
\label{fig:loadpalletprime}
\end{figure}

The goal is then extended to include the proposition: \textit{(has\_done\_load\_pallet Tom p2 sh6)}. The  HPlan produced from solving the HModel described is shown in Figure~\ref{fig:addplan}.


\begin{figure}[h!]
\begin{lstlisting}[xleftmargin=.15\textwidth]
0.000: (goto_waypoint Tom sh5 sh6)  [3.000]
0.000: (load_pallet Jerry p1 sh3)  [2.000]
2.000: (goto_waypoint Jerry sh3 sh4)  [5.000]
3.001: (set_shelf Tom sh6)  [1.000]
4.001: (goto_waypoint Tom sh6 sh1)  [4.000]
7.001: (goto_waypoint Jerry sh4 sh5)  [1.000]
8.001: (set_shelf Tom sh1)  [1.000]
9.001: (goto_waypoint Tom sh1 sh6)  [4.000]
13.001: (load_pallet Tom p2 sh6)  [2.000]
15.001: (goto_waypoint Tom sh6 sh1)  [4.000]
19.001: (unload_pallet Tom p2 sh1)  [1.500]
19.002: (goto_waypoint Jerry sh5 sh6)  [3.000]
22.002: (unload_pallet Jerry p1 sh6)  [1.500]
\end{lstlisting}
\caption{The HPlan containing the user suggested action \textit{load\_pallet Tom p2 sh6} with a duration of 23.502}
\label{fig:addplan}
\end{figure}

\subsubsection{Justified Actions and Expected Plans}\label{comp:keycausal}


Usually, a user asks a contrastive question about a plan when they expected a different outcome or some sub-goal to be achieved in a certain way. In the example shown in~\ref{comp:add}, the user expected the robot \textit{Tom} to load the pallet \textit{p2} onto the shelf \textit{sh6}, which their question reflects. 
It is clear why the user asked this question as it fully describes the goal they wish to achieve and how to achieve it. 
The constraint derived from this question causes an immediate impact in the plan. The package is delivered using a different robot than previously. However, the objective of some questions are not as clear.
For example, if a user questioned ``Why is \textit{(set\_shelf Tom sh4)} not used, rather than being used?'', it is not clear what they intend to achieve with this action. 
The HPlan produced from the HModel containing the constraint for this question is shown in Figure\ref{fig:unjustified_plan}. The plan starts with some preliminary movement actions that allow the robot \textit{Tom} to set up the required shelf \textit{sh4}. \textit{Tom} then traverses to the shelf \textit{sh6}, the plan then continues the same as the original plan in Figure~\ref{fig:plan}. The action \textit{(set\_shelf Tom sh4)} does not affect the plan and it would still be valid if the action were removed. The reason for this could be due to the plan that utilises the action being more expensive, or it could be due to it not being possible for the action \textit{(set\_shelf Tom sh4)} to achieve anything useful. However, it could also be because the planner could not find a plan where the action is used in such a way that it contributes to the goal. 
For this reason, a user may not be satisfied with an HPlan where the action is not used in a way that is necessary for achieving a goal, we discuss what this means in more detail in Section~\ref{comp:just}. Although the compilations formalised in this section do not guarantee that any actions a user suggests are necessary for achieving a goal, the rest of this subsection provides a step towards this with the description and formalisation of a compilation.

\begin{figure}[h!]
\centering
\begin{BVerbatim}
0.000: (load_pallet jerry p1 sh3)  [2.000]
0.000: (goto_waypoint tom sh5 sh4)  [1.000]
1.001: (set_shelf tom sh4)  [1.000]
2.001: (goto_waypoint tom sh4 sh5)  [1.000]
3.002: (goto_waypoint jerry sh3 sh4)  [5.000]
3.002: (goto_waypoint tom sh5 sh6)  [3.000]
6.002: (set_shelf tom sh6)  [1.000]
7.002: (goto_waypoint tom sh6 sh1)  [4.000]
8.003: (goto_waypoint jerry sh4 sh5)  [1.000]
11.002: (set_shelf tom sh1)  [1.000]
11.003: (goto_waypoint jerry sh5 sh6)  [3.000]
12.002: (goto_waypoint tom sh1 sh2)  [4.000]
14.003: (unload_pallet jerry p1 sh6)  [1.500]
15.504: (load_pallet jerry p2 sh6)  [2.000]
17.504: (goto_waypoint jerry sh6 sh1)  [4.000]
21.504: (unload_pallet jerry p2 sh1)  [1.500]
\end{BVerbatim}
\caption{ The HPlan with a cost of 23.004 generated to satisfy the constraint derived from the question ``Why is \textit{(set\_shelf Tom sh4)} not used, rather than being used?''.}
\label{fig:unjustified_plan}
\end{figure}

The compilation works by tracking the facts that have been produced through effects of actions that the user suggested action $\tau$ has causally supported. One of these facts then has to be a goal fact. Therefore, there is a causal chain from $\tau$ to a goal and the action $\tau$ is necessary for achieving the goal in any plan produced by a model with this constraint applied. 
For example this compilation ensures that, in the HPlan $\pi'$, there will be a causal chain, $\mu \subseteq \pi' = \langle \tau, a_1, a_2, \dots, a_n \rangle$ where for the state $s_{n + 1}$ after $a_n$ is finished executing and some $g \in G$ then $s_{n + 1} \models g$, and for all actions $a_i \in \mu$ if $a_i$ was removed then $\pi' \not \models G$, assuming $g$ is not already satisfied in the initial state.

To generate an HPlan that adheres to these properties and satisfies the user question ``Why is \textit{a = (set\_shelf Tom sh4)} not used, rather than being used?'', the model is compiled in the following way. A new operator $o_a$ is created which has the same preconditions and effects as $a$, but for each positive effect, has a new effect which adds a copy of the fact, we call this the prime-fact.
A new operator is then created for each precondition $p$ for each operator $o$ in the domain. The precondition to this new operator is the same as $o$ with a new precondition $prime_{p}$. The effects are the same as $o$ but for each positive effect the corresponding prime-fact is also made true. These new actions behave the same as the existing actions in the domain, but they propagate the causal chains originating from $a$ through the prime-facts. A final set of operators is added for each goal which can be applied if both a goal and it's corresponding prime-fact are true, and at least one of these actions must appear in the plan for it to be valid. 
This is a work around used because the majority of PDDL2.1 planners do not accommodate disjunctive goals, however, this can be simplified by changing the goal to $G \wedge (\vee_{i=0}(g_i \wedge prime_{gi}))$. 
If a goal has already been achieved by another action in the plan that is not part of the causal chain from $a$ then this action can no longer be applied. The causality of the actions is tracked through these prime-facts and for any valid plan there will exist a goal that can have it's origin traced through prime-facts back to the user suggested action $a$.

The HModel $\Pi'$ is:
$$
\Pi' = \langle \langle Ps', Vs, As', arity' \rangle, \langle Os, I', G', W\rangle \rangle
$$
where:
\begin{itemize}
    \item $Ps_{p} = \{prime_{p} , \forall p \in Ps\}$
    \item $G_{p} = \{goal\_prime_{p}, \forall p \in Ps$ where $ground(p, \chi) = g \in G$ for some $\chi\}$
    \item $Ps' = Ps \cup \{has\_done\_a, can\_do\_a, true, Ps_{p}, G_{p}\}$
    \item $As' = As \cup \{o_a\} \cup \{conjunct_{xp}, \forall x \in As :\forall p \in Pre{\vdash \dashv}(x)\} \cup
    \{check\_conjunct_{g}, \forall g \in G_p$\}
    \item $arity'(x) = arity(x),\, \forall x\in Ps$
    \item $arity'(goal\_prime_{p}) = arity(p),\, \forall p \in Ps$ where $ground(p, \chi) = g \in G$ for some $\chi\}$
    \item $arity'(prime_{p}) = arity(p),\, \forall p \in Ps$
    \item $arity'(has\_done\_a) = arity'(can\_do\_a) = arity'(o_a) = arity(o)$
    \item $arity'(true) = \emptyset$
    \item $I' = I \cup \{ground(can\_do\_a, \chi) \cup \{ground(goal\_prime_{p}, \chi'), \forall p \in G_{p}$ where $ground(p, \chi') = g, \forall g \in G\}$
    \item $G' = G \cup \{ground(has\_done\_a,\chi)\} \cup true$
\end{itemize}
and the actions are defined such that the preconditions and effects are:$$
\begin{array}{r@{}l}
Pre_{\vdash}(o_a) = Pre_{\vdash}(o)\ &\cup\ \{can\_do\_a\} \\
\Eff^+_{\dashv}(o_a) = \Eff^+_{\dashv}(o)\ &\cup\ \{has\_done\_a\} \\
& \cup\ \{prime_{y} \in Ps_{p}, \forall y \in \Eff^+_{\vdash \dashv }(o)\}\\
Pre_{\vdash \dashv}(conjunct_{xp}) = Pre_{\vdash \dashv}(x)\ &\cup\ prime_{p}, \forall x \in As :\forall p \in Pre{\vdash \dashv}(x)\\
\Eff^{+}_{\vdash \dashv}(conjunct_{xp}) = \Eff^{+}_{\vdash \dashv}(x)\ &\cup\ \{prime_{y} \in Ps_{p}, \forall y \in \Eff^+_{\vdash \dashv}(x)\},\forall x \in As :\forall p \in Pre{\vdash \dashv}(x)\\
Dur(check\_conjunct_{g}) \, & = \epsilon \, \text{where} \, \epsilon \, \text{is a very small number}, \forall goal\_prime_{g} \in G_{p}\\
Pre_{\vdash}(check\_conjunct_{g}) \, &= prime_{g} \cup goal\_prime_{g}, \forall g\in Ps' \text{where} \, prime_{g} \in Ps_{p} \wedge goal\_prime_{g} \in G_p  \\
\Eff^{+}_{\vdash \dashv}(check\_conjunct_{g}) \, &= true \\
\Eff^{-}_{\dashv}(o) = \Eff^{-}_{\dashv} \,&\cup \, \{goal\_prime_{g}, \forall g \in \Eff^{+}_{\vdash \dashv}(o)\, \text{where} \, goal\_prime_{g} \in G_p \}, \forall o \in As
\end{array}
$$

\begin{figure}[h!]
\centering
\begin{BVerbatim}
0.000: (goto_waypoint jerry sh3 sh2)  [8.000]
0.000: (goto_waypoint tom sh5 sh4)  [1.000]
1.001: (done-set_shelf tom sh4)  [1.000]
8.001: (goto_waypoint jerry sh2 sh1)  [4.000]
8.001: (goto_waypoint tom sh4 sh3)  [5.000]
12.002: (set_shelf jerry sh1)  [1.000]
13.001: (load_pallet tom p1 sh3)  [2.000]
13.002: (goto_waypoint jerry sh1 sh6)  [4.000]
15.001: (goto_waypoint tom sh3 sh4)  [5.000]
17.002: (set_shelf jerry sh6)  [1.000]
18.002: (load_pallet jerry p2 sh6)  [2.000]
20.001: (unload_pallet-2-conjunct tom p1 sh4)  [1.500]
20.002: (goto_waypoint jerry sh6 sh1)  [4.000]
21.502: (load_pallet-0-conjunct tom p1 sh4)  [2.000]
23.502: (goto_waypoint tom sh4 sh5)  [1.000]
24.002: (unload_pallet jerry p2 sh1)  [1.500]
24.503: (goto_waypoint tom sh5 sh6)  [3.000]
27.503: (unload_pallet-0-conjunct tom p1 sh6)  [1.500]
29.003: (check-conjunct-pallet_at p1 sh6 true)  [0.100]
\end{BVerbatim}
\caption{ The HPlan with a cost of 29.003 generated to satisfy the constraint derived from the question ``Why is \textit{(set\_shelf Tom sh4)} not used, rather than being used?'', such that the action is necessary in the plan for achieving a goal. The action names are trivially renamed back to their corresponding actions, and the action (check-conjunct-pallet\_at p1 sh6 true) is removed.}
\label{fig:justified_plan}
\end{figure}

The plan for this is shown in Figure~\ref{fig:justified_plan} where the action \textit{(set\_shelf tom sh4)} is necessary for performing the action \textit{(unload\_pallet tom p1 sh6)} which achieves the goal \textit{(pallet\_at p1 sh6)}. However, this compilation does not guarantee that the action $a$ will be perfectly justified in the plan $\pi$, that is that there is no set of actions $A$ where $a \in A$ and $A \subseteq \pi$, such that if you removed the set of actions $A$ then $\pi \models G$~\cite{fin92}. This means that there are no groups of actions that together are redundant in the plan. This is not the case for the HPlan in Figure~\ref{fig:justified_plan}, if the set of actions \{(done-set\_shelf tom sh4), (unload\_pallet-2-conjunct tom p1 sh4), (load\_pallet-0-conjunct tom p1 sh4)\} is removed, the plan is still valid. To attempt to determine whether there is a plan where $a$ is perfectly justified would likely require an extended search over these redundancy sets. This search would be the repeated process of disallowing an action in the redundancy set to be applied in the plan, re-planning, and generating the new redundancy set. The search would end when a plan is found where the action is used in a perfectly justified way, or all the redundancy sets have been searched over and no plan was found, meaning the action cannot be used in a perfectly justified way.

This approach also works if the goal contains primitive numeric expressions in the same way. Any effects that alter the values of PNEs, will duplicate the behaviour with a prime-effect. The goal is checked in the same way as with a simple proposition. For example, if an action $\tau$ decreases the value of a PNE $n$, and there is a goal such that $5 < n < 10$ is true at the end of the plan. Then $\tau$ affects $prime_n$ in the same way as it does $n$ and both $5 < n < 10$ and $5 < prime_n < 10$ must be true at the end of the plan for it to be valid.

This approach can be adapted for use in the compilations for all formal questions apart from FQ2 where it would have no use as an action is removed rather than added.


\subsection{Remove a Specific Grounded Action}\label{comp:remove}

Given a plan $\pi$, a formal question $Q$ is asked of the form:
\begin{quote}
\textit{Why is the operator $o$ with parameters $\chi$ used, rather than not being used?}
\end{quote}
For example, given the example plan in Figure~\ref{fig:plan} the user might ask:
\begin{quote}
``Why is \textit{(goto\_waypoint Tom sh1 sh2)} used, rather than not being used?''
\end{quote}

A user might ask this because \textit{Tom} has already set up all of the shelves that are required. The user might question why \textit{Tom} is doing this extra action.

The specifics of the compilation is similar to the compilation in Section~\ref{comp:add}. The HModel is extended to introduce a new predicate \textit{not\_done\_action} which represents actions that have not yet been performed. The operator \textit{o} is extended with the new predicate as an additional delete effect. The initial state and goal are then extended to include the user selected grounding of \textit{not\_done\_action}. Now, when the user selected action is performed it deletes the new goal and so invalidates the plan. This ensures the user suggested action is not performed.

For example, given the user question above, an HPlan is generated that does not include the action \textit{(goto\_waypoint Tom sh1 sh2)}, and is shown in Figure~\ref{fig:removeplan}. This shows a plan with a longer duration than the original plan shown in Figure~\ref{fig:plan}. In this HPlan \textit{Tom} has to deliver pallet \textit{p2} because he is occupying shelf \textit{sh1} and cannot vacate it by going to shelf \textit{sh2}. This means \textit{Jerry} cannot pass by him to deliver the pallet more efficiently.

\begin{figure}[h!]
\begin{lstlisting}[xleftmargin=.15\textwidth]
0.000: (goto_waypoint Tom sh5 sh6)  [3.000]
0.000: (load_pallet Jerry p1 sh3)  [2.000]
2.000: (goto_waypoint Jerry sh3 sh4)  [5.000]
3.001: (set_shelf Tom sh6)  [1.000]
4.001: (goto_waypoint Tom sh6 sh1)  [4.000]
7.001: (goto_waypoint Jerry sh4 sh5)  [1.000]
8.001: (set_shelf Tom sh1)  [1.000]
9.001: (goto_waypoint Tom sh1 sh6)  [4.000]
13.001: (load_pallet Tom p2 sh6)  [2.000]
15.001: (goto_waypoint Tom sh6 sh1)  [4.000]
19.001: (unload_pallet Tom p2 sh1)  [1.500]
19.002: (goto_waypoint Jerry sh5 sh6)  [3.000]
22.002: (unload_pallet Jerry p1 sh6)  [1.500]
\end{lstlisting}
\caption{The HPlan without the action \textit{(goto\_waypoint Tom sh1 sh2)} with a duration of 23.502}
\label{fig:removeplan}
\end{figure}

\subsection{Replacing an Action in a State}\label{comp:replace}

Given a plan $\pi$, a formal question $Q$ is asked of the form:
\begin{quote}
\textit{Why is the operator $o$ with parameters $\chi$ used in state $s$, rather than the operator $n$ with parameters $\chi'$? where $o \neq n$ or $\chi \neq \chi'$ }
\end{quote}
For example, given the example plan in Figure~\ref{fig:plan} the user might ask:
\begin{quote}
``Why is \textit{(set\_shelf Tom sh6)} used, rather than \textit{(load\_pallet Tom p2 sh6)}?''
\end{quote}

The user might ask this because a goal of the problem is to deliver the pallet \textit{p2} to the shelf \textit{sh1}. As \textit{Tom} is by the pallet, the user might question why \textit{Tom} does not load the pallet in order to deliver it instead of setting up the shelf \textit{sh6}.

To generate the HPlan, a compilation is formed such that the ground action $b = ground(n, \chi')$ appears in the plan in place of the action $a_i = ground(o, \chi)$. Given the example above $b = ground(load\_pallet,\{Tom, p2, sh6\})$, and $a_i = ground(set\_shelf,\{Tom, sh6\})$. Given a plan:
$$\pi = \langle a_1,a_2,\ldots,a_n \rangle$$
The ground action $a_i$ at state $s$ is replaced with $b$, which is executed, resulting in state $I'$, which becomes the new initial state in the HModel. A time window is created for each durative action that is still executing in state $s$. These model the end effects of the concurrent actions. A plan is then generated from this new state with these new time windows for the original goal, which gives us the plan:
$$\pi' = \langle a'_1,a'_2,\ldots,a'_n \rangle$$
The HPlan is then the initial actions of the original plan $\pi$ concatenated with $b$ and the new plan $\pi'$:
$$\langle a_1,a_2,\ldots,a_{i-1}, b, a'_1,a'_2,\ldots,a'_n \rangle$$

\noindent Specifically, the HModel $\Pi'$ is:
$$
\Pi' = \langle \langle Ps, Vs, As, arity \rangle, \langle Os, I', G, W \cup C \rangle \rangle
$$
where:
\begin{itemize}
    \item $I'$ is the final state obtained by executing\footnote{We use VAL to validate this execution. We use the add and delete effects of each action, at each happening (provided by VAL), up to the replacement action to compute $I'$.} $\langle a_1,a_2,\ldots,a_{i-1}, b \rangle$ from state $I$.
    \item $C$ is a set of time windows $w_x$, for each durative action $a_j$ that is still executing in the state $I'$. For each such action, $w_x$ specifies that the end effects of that action will become true at the time point at which the action is scheduled to complete. Specifically:
    $w_x=\langle (Dispatch(a_j) + Dur(a_j)) - (Dispatch(b) + Dur(b)), inf, u \rangle$
    where
    $u = \Eff(a_j)_{\dashv}^- \cup \Eff(a_j)_{\dashv}^+ \cup \Eff(a_j)_{\dashv}^n$.
\end{itemize}


\noindent In the case in which an action $a_j$ that is executing in state $I'$ has an overall condition that is violated, this is detected when the plan is validated against the original model. 
As an example, given the user question above, the new initial state $I'$ from the running example is shown in Figure~\ref{fig:replaceinit}.

\begin{figure}[h!]
\hspace{1cm}\vdots
\begin{lstlisting} %[xleftmargin=.15\textwidth]

(:init
    (not_occupied sh1) (not_occupied sh2) (not_occupied sh5)
    (connected sh1 sh2) (connected sh2 sh1) (connected sh2 sh3) 
    (connected sh3 sh2) (connected sh3 sh4) (connected sh4 sh3) 
    (connected sh4 sh5) (connected sh5 sh4) (connected sh5 sh6)
    (connected sh6 sh5) (connected sh6 sh1) (connected sh1 sh6)
    (pallet_at p1 jerry) (pallet_at p2 tom) (robot_at tom sh6) 
    (at 3 (robot_at Jerry sh4)) (at 3 (not_occupied sh3)) 
     (= (travel_time sh1 sh2) 4) (= (travel_time sh1 sh6) 4)
    (= (travel_time sh2 sh1) 4) (= (travel_time sh2 sh3) 8)
    (= (travel_time sh3 sh2) 8) (= (travel_time sh3 sh4) 5)
    (= (travel_time sh4 sh3) 5) (= (travel_time sh4 sh5) 1)
    (= (travel_time sh5 sh4) 1) (= (travel_time sh5 sh6) 3)
    (= (travel_time sh6 sh5) 3) (= (travel_time sh6 sh1) 4))
(:goal (and (pallet_at p1 sh6) (pallet_at p2 sh1))))
\end{lstlisting}
\caption{The initial state \textit{I'} which captures the state directly after executing the alternate action \textit{b = \textit{(load\_pallet Tom p2 sh6)}} suggested by the user.}
\label{fig:replaceinit}
\end{figure}


This captures the state $I'$, resulting from executing the actions $a_1, a_2, a_3$, and $b$:\\
\noindent\verb|0.000: (goto_waypoint Tom sh5 sh6) [3.000]|\\
\noindent\verb|0.000: (load_pallet Jerry p1 sh3) [2.000]|\\
\noindent\verb|2.000: (goto_waypoint Jerry sh3 sh4) [5.000]|\\
\noindent\verb|3.001: (load_pallet Tom p2 sh6) [2.000]|



In this state \textit{Tom} is at shelf \textit{sh6} and has loaded the pallet \textit{p2}. \textit{Jerry} has loaded the pallet \textit{p1} and is currently moving from shelf \textit{sh3} to \textit{sh4},
This new initial state is then used to plan for the original goals to get the plan $\pi'$, which, along with $b$ and $\pi$, gives the HPlan.
However, the problem is unsolvable from this state as a robot cannot set up a shelf whilst it is transporting a pallet, a shelf must be set up to unload a pallet, \textit{Tom} and \textit{Jerry} are both holding pallets, and there are no shelves set up. Therefore, neither \textit{Tom} nor \textit{Jerry} can unload a pallet at any of the shelves and so can not achieve the goal.
By applying the user's constraint, and showing there are no more applicable actions, it answers the above question: ``because by doing $b$ rather than $a$, there is no way to complete the goals of the problem''.

This compilation keeps the position of the replaced action in the plan, however, it may not be optimal. This is because we are only re-planning after the inserted action has been performed. The first half of the plan, because it was originally planned to support a different set of actions, may now be inefficient, as shown by~\citeauthor{bor18} \shortcite{bor18}.

If the user instead wishes to replace the action without necessarily retaining its position in the plan, then the add and remove compilations shown in Sections~\ref{comp:add} and~\ref{comp:remove} can be applied iteratively. This is an example of how the compilations can be combined into something greater than the sum of it's parts, that answers an entirely new question.

\subsection{Reordering Actions}\label{comp:reschedule}

Given a plan $\pi$, a formal question $Q$ is asked of the form:
\begin{quote}
\textit{Why is the operator $o$ with parameters $\chi$ used before (after) the operator $n$ with parameters $\chi'$, rather than after (before)? where $o \neq n$ or $\chi \neq \chi'$}
\end{quote}
For example, given the example plan in Figure~\ref{fig:plan} the user might ask:
\begin{quote}
``Why is \textit{(unload\_pallet Jerry p1 sh6)} used before \textit{(unload\_pallet Jerry p2 sh1)}, rather than after?''
\end{quote}

A user might wonder what would be the outcome if \textit{Jerry} delivered the pallets the other way around. There are the same amount of shelves to traverse between each of the delivery points so the user might wonder if there is a reason it was done in this order. They can therefore ask the question posed above and see what happens if \textit{Jerry} delivered pallet \textit{p2} before \textit{p1}.

The compilation to the HModel is performed in the following way. 
First, a directed-acyclic-graph (DAG) $\langle N,E \rangle$ is built to represent each ordering between actions suggested by the user. For example the ordering of $Q$ is $a \prec b$ where $a = ground(o, \chi)$ and $b = ground(n, \chi')$. 


This DAG is then encoded into the model $\Pi$ to create $\Pi'$. For each edge $(a, b)\in E$ two new predicates are added: $ordered_{ab}$ representing that an edge exists between $a$ and $b$ in the DAG, and $traversed_{ab}$ representing that the edge between actions $a$ and $b$ has been traversed.

For each node representing a ground action $a\in N$, the action is disallowed using the compilation from Section~\ref{comp:remove}. Also, for each such action a new operator $o_a$ is added to the domain, with the same functionality of the original operator $o$. The arity of the new operator, $arity(o_a)$ is the combined arity of the original operator plus the arity of all of $a$'s sink nodes. Specifically, the HModel $\Pi'$ is:
$$
\Pi' = \langle \langle Ps', Vs, As', arity' \rangle, \langle Os, I', G', W\rangle \rangle
$$
where:
\begin{itemize}
\item $Ps' = Ps \cup \{ordered_{ab}\} \cup \{traversed_{ab}\},\,\forall (a,b)\in E$
\item $As' = As \cup \{o_a\},\,\forall a\in N$
\item $arity'(x)   = arity(x), \forall x \in arity$
\item $arity'(o_a) = arity(o) + \sum_{(a,b) \in E} arity(b), \forall a \in N$
\item $arity'(ordered_{ab}) = arity(a) + arity(b), \forall (a, b) \in E$
\item $arity'(traversed_{ab}) = arity(b), \forall (a, b) \in E$
\item $I' = I \cup ground(ordered_{ab}, \chi + \chi'),\,\forall (a,b)\in E$, where $\chi$ and $\chi'$ are the parameters of $a$ and $b$, respectively.
\end{itemize}
In the above, we abuse the $arity$ notation to specify the arity of an action to mean the arity of the operator from which it was ground; e.g. $arity(a) = arity(o)$ where $a=ground(o,\chi)$.

Each new operator $o_a$ extends $o$ with the precondition that all incoming edges must have been traversed, i.e. the source node has been performed. The effects are extended to add that its outgoing edges have been traversed. That is:
$$
\begin{array}{r@{}l}
Pre_{\vdash}(o_a) = Pre_{\vdash}(o)\ &\cup\ \{ordered_{ab} \in Ps', \forall b\}\\
&\cup\ \{traversed_{ca} \in Ps', \forall c\}\\
\Eff^+_{\dashv}(o_a) = \Eff^+_{\dashv}(o)\ & \cup\ \{traversed_{ab} \in Ps' , \forall b\}
\end{array}
$$

This ensures that the ordering the user has selected is maintained within the HPlan.

As the operator $o_a$ has a combined arity of the original operator plus the arity of all of $a$'s sink nodes, there exists a large set of possible ground actions. However, for all $b\in N$, $ordered_{ab}$ is a precondition of $o_a$; and for each edge $(a,b)\in E$ the ground proposition $ground(ordered_{ab}, \chi + \chi')$ is added to the initial state to represent that the edge exists in the DAG. This significantly prunes the possible, valid, groundings of $o_a$.
  
Given the user question above, two new operators \textit{node\_unload\_pallet\_Jerry\_p2\_sh1} (shown in Figure~\ref{fig:reorderaction}) and \textit{node\_unload\_pallet\_Jerry\_p1\_sh6} will be added to the domain. These extend operator \textit{unload\_pallet} from Figure~\ref{fig:domain} as described above. The HPlan generated is shown in Figure~\ref{fig:reorderplan}. In this case the plan does not contain the action \textit{unload\_pallet Jerry p1 sh6} and instead uses \textit{Tom} to deliver the pallet \textit{p1}. If the user wants both the before and after actions to be performed in the plan they can successively apply the add compilation shown in Section~\ref{comp:add}.

\begin{figure}[h!]
\begin{small}
\begin{lstlisting}
(:durative-action  node_unload_pallet_Jerry_p2_sh1
 :parameters (?v - robot ?p - pallet ?shelf - waypoint
  ?v0 - robot ?p0 - pallet ?shelf0 - waypoint)
 :duration (= ?duration 1.5)
 :condition (and (at start (pallet_at ?p ?v))
  (over all (robot_at ?v ?shelf)) (over all (scanned_shelf ?shelf)))
  (at start (ordered-node-unload_pallet-Jerry-p2-sh1-unload_pallet
  -Jerry-p1-sh6 ?v ?p ?shelf ?v0 ?p0 ?shelf0))
 :effect (and (at end (not_holding_pallet ?v))
  (at end (pallet_at ?p ?shelf)) (at start (not (pallet_at ?p ?v)))
  (at end (traversed-node-unload_pallet-Jerry-p2-sh1-unload_pallet
  -Jerry-p1-sh6 ?v0 ?p0 ?shelf0)))
)
\end{lstlisting}
\end{small}
\caption{An operator added to the original domain to capture an ordering constraint between actions. The operator extends the original \textit{unload\_pallet} operator.}
\label{fig:reorderaction}
\end{figure}

\begin{figure}[h!]
\begin{lstlisting}
0.000: (goto_waypoint jerry sh3 sh2)  [8.000]
0.000: (goto_waypoint tom sh5 sh6)  [3.000]
3.001: (set_shelf tom sh6)  [1.000]
4.001: (goto_waypoint tom sh6 sh5)  [3.000]
7.002: (goto_waypoint tom sh5 sh4)  [1.000]
8.001: (goto_waypoint jerry sh2 sh1)  [4.000]
8.003: (goto_waypoint tom sh4 sh3)  [5.000]
12.002: (set_shelf jerry sh1)  [1.000]
13.002: (goto_waypoint jerry sh1 sh6)  [4.000]
13.003: (load_pallet tom p1 sh3)  [2.000]
15.003: (goto_waypoint tom sh3 sh4)  [5.000]
17.002: (load_pallet jerry p2 sh6)  [2.000]
19.002: (goto_waypoint jerry sh6 sh1)  [4.000]
20.004: (goto_waypoint tom sh4 sh5)  [1.000]
23.002: (node-unload_pallet-jerry-p2-sh1 jerry p2 sh1 jerry p1 sh6)  [1.500]
23.003: (goto_waypoint tom sh5 sh6)  [3.000]
26.003: (node-unload_pallet-tom-p1-sh6 tom p1 sh6)  [1.500]
\end{lstlisting}
\caption{The HPlan with the action \textit{(unload\_pallet Jerry p2 sh1)} before \textit{(unload\_pallet Jerry p1 sh6)} with a duration of 27.503}
\label{fig:reorderplan}
\end{figure}

\subsection{Forbid an Action Outside a Time Window}\label{comp:tw1}

Given a plan $\pi$, a formal question $Q$ is asked of the form:
\begin{quote}
\textit{Why is the operator $o$ with parameters $\chi$ used outside of time $lb < t < ub$, rather than only being allowed within this time window?}
\end{quote}
For example, given the example plan in Figure~\ref{fig:plan} the user might ask:
\begin{quote}
``Why is \textit{(unload\_pallet Jerry p2 sh1)} used outside the interval 11 to 13, rather than being restricted to that time window?''
\end{quote}

From the HPlan provided as a result of the question asked in Section~\ref{comp:reschedule}, the user might wonder why changing the original order of the actions \textit{a = unload\_pallet Jerry p2 sh6} and \textit{b = unload\_pallet Jerry p2 sh1}, caused \textit{b} to be performed at the time 23.002 rather than at 11.002, which was the time that action \textit{a} was originally performed. The user might then ask the question above about the original plan, to receive an explanation for why the action \textit{b} cannot be performed at the same time as when \textit{a} was performed.

To generate the HPlan, the planning model is compiled so that the ground action $a = ground(o, \chi)$ can only be used between times $lb$ and $ub$. To do this, the original operator $o$ is replaced with two operators $o_{a}$ and $o_{\neg a}$, which extend $o$ with extra constraints.

Operator $o_{\neg a}$ replaces the original operator $o$ for all other actions $ground(o,\chi')$, where $\chi'\neq \chi$. The action $ground(o_{\neg a}, \chi)$ cannot be used (this is enforced using the compilation for forbidding an action described in Section~\ref{comp:remove}).
Operator $o_{a}$ acts as the operator $o$ specifically for the action $a=ground(o, \chi)$, which has an added constraint that it can only be performed between $lb$ and $ub$. 
\noindent Specifically, the HModel $\Pi'$ is:
$$
\Pi' = \langle \langle Ps', Vs, As', arity' \rangle, \langle Os, I', G', W'\rangle \rangle
$$
where:
\begin{itemize}
    \item $Ps' = Ps \cup \{can\_do\_a, not\_done\_a\}$
    \item $As' = \{o_{a}, o_{\neg a}\} \cup As \setminus \{o\} $
    \item $arity'(x) = arity(x), \forall x \in arity$
    \item $arity'(can\_do\_a) = arity'(not\_done\_a) = arity'(o_a) = arity'(o_{\neg a}) = arity(o)$
    \item $I' = I \cup \{ground(not\_done\_a, \chi)\}$
    \item $G' = G \cup \{ground(not\_done\_a, \chi)\}$
    \item $W' = W \cup \{\langle lb, ub, ground(can\_do\_a, \chi) \rangle\}$
\end{itemize}
where the new operators $o_{\neg a}$ and $o_{a}$ extend $o$ with the delete effect $not\_done\_a$ and the precondition $can\_do\_a$, respectively. i.e:
$$
\begin{array}{c}
\Eff_\vdash^-(o_{\neg a}) = \Eff_\vdash^-(o) \cup \{not\_done\_a\}\\
Pre_\vdash(o_{a}) = Pre_\vdash(o) \cup \{can\_do\_a\}
\end{array}
$$
As the proposition $ground(can\_do\_a,\chi)$ must be true for $ground(o_{a}, \chi)$ to be performed, this ensures that the action $a$ can only be performed within the times $lb$ and $ub$. Other actions from the same operator can still be applied at any time using the new operator $o_{\neg a}$. As in Section~\ref{comp:remove} we make sure the ground action $ground(o_{\neg a}, \chi)$ can never appear in the plan.


For example, given the user question above, the operator \textit{unload\_pallet} from Figure~\ref{fig:domain} is extended to $o_{\neg a}$ and $o_{a}$ as shown below in Figure~\ref{fig:tw1action}.

\begin{figure}[h!]
\begin{lstlisting}
(:durative-action  unload_pallet_nota
 :parameters (?v - robot ?p - pallet ?shelf - waypoint)
 :duration (= ?duration 1.5)
 :condition (and (at start (pallet_at ?p ?v))
  (over all (robot_at ?v ?shelf)) (over all (scanned_shelf ?shelf)))
 :effect (and (at end (not_holding_pallet ?v))
  (at end (pallet_at ?p ?shelf))
  (at start (not (pallet_at ?p ?v)))
  (at start (not (not-done-unload_pallet ?v ?p ?shelf)))))
\end{lstlisting}

\begin{lstlisting}
(:durative-action  unload_pallet_a
 :parameters (?v - robot ?p - pallet ?shelf - waypoint)
 :duration (= ?duration 1.5)
 :condition (and (at start (pallet_at ?p ?v))
  (over all (robot_at ?v ?shelf))
  (over all (scanned_shelf ?shelf))
  (over all (applicable-unload_pallet ?v ?p ?shelf)))
 :effect (and (at end (not_holding_pallet ?v))
  (at end (pallet_at ?p ?shelf))
  (at start (not (pallet_at ?p ?v)))))
\end{lstlisting}
\caption{The PDDL2.1 representation of the operators $o_{\neg a}$ and $o_{a}$.}
\label{fig:tw1action}
\end{figure}

The initial state is extended to include the proposition \textit{(not\_done\_unload\_pallet Jerry p2 sh1)} and the time window $\langle 11, 13, (can\_do\_load\_pallet\,Jerry\,p2\,sh1) \rangle$, which enforces that the proposition is true only between the times 11 and 13. The resulting HPlan is shown in Figure~\ref{fig:tw1plan}, in this case the action \textit{(unload\_pallet Jerry p2 sh1)} is no longer present in the plan as \textit{Tom} delivers the pallet \textit{p2} instead.

\begin{figure}[h!]
\begin{verbatim}
0.000: (goto_waypoint tom sh5 sh6)  [3.000]
0.000: (load_pallet jerry p1 sh3)  [2.000]
2.000: (goto_waypoint jerry sh3 sh4)  [5.000]
3.001: (set_shelf tom sh6)  [1.000]
4.001: (goto_waypoint tom sh6 sh1)  [4.000]
7.001: (goto_waypoint jerry sh4 sh5)  [1.000]
8.001: (set_shelf tom sh1)  [1.000]
9.001: (goto_waypoint tom sh1 sh6)  [4.000]
13.001: (load_pallet tom p2 sh6)  [2.000]
15.001: (goto_waypoint tom sh6 sh1)  [4.000]
19.001: (unload_pallet_nota tom p2 sh1)  [1.500]
19.002: (goto_waypoint jerry sh5 sh6)  [3.000]
22.002: (unload_pallet_nota jerry p1 sh6)  [1.500]
\end{verbatim}
\caption{The HPlan produced from solving the HModel that allows the action \textit{(unload\_pallet Jerry p2 sh1)} to only be performed between the times 11 and 13. \textit{Tom} was, therefore, chosen to deliver the package instead.}
\label{fig:tw1plan}
\end{figure}

\subsection{Add an Action Within a Time Window}\label{comp:tw2}

Given a plan $\pi$, a formal question $Q$ is asked of the form:
\begin{quote}
\textit{Why is the operator $o$ with parameters $\chi$ not used at time $lb < t < ub$, rather than being used in this time window?}
\end{quote}
For example, given the example plan in Figure~\ref{fig:plan} the user might ask:
\begin{quote}
``Why is \textit{(unload\_pallet Jerry p2 sh1)} not used between times 11 and 13, rather than being used in this time window?''
\end{quote}
The HPlan given in Section~\ref{comp:tw1} shows the user that there is a better plan which does not have the action in this time window. 
However, the user may only be satisfied once they have seen a plan where the action is performed in their given time window. To allow this the action may have to appear in other parts of the plan as well.

This constraint differs from Section~\ref{comp:tw1} in two ways: first the action is now forced to be applied in the time window, and second the action can be applied at other times in the plan. This constraint is useful in cases such as a robot that has a fuel level. As fuel is depleted when travelling between waypoints, the robot must refuel, possibly more than once. The user might ask ``why does the robot not refuel between the times $x$ and $y$ (as well as the other times it refuels)?''.

To generate the HPlan, the planning model is compiled into a form that forces the ground action, $a = ground(o, \chi)$, to be used between times $lb$ and $ub$, but can also appear at any other time. This is done using a combination of the compilation in Section~\ref{comp:add} and a variation of the compilation in Section~\ref{comp:tw1}. The former ensures that new action $ground(o_{a},\chi)$ must appear in the plan, and the latter ensures that the action can only be applied within the time window. The variation of the latter compilation is that the operator $o_{\neg a}$ is not included, and instead the original operator is kept in the domain. This allows the original action $a=ground(o,\chi)$ to be applied at other times in the plan.
Given this, the HModel $\Pi'$ is:
$$
\Pi' = \langle \langle Ps', Vs, As', arity' \rangle, \langle Os, I, G', W'\rangle \rangle
$$
where:
\begin{itemize}
    \item $Ps' = Ps \cup \{can\_do\_a, has\_done\_a\}$
    \item $As' = As \cup \{o_{a}\}$
    \item $arity'(x) = arity(x), \forall x \in arity$
    \item $arity'(can\_do\_a) = arity'(has\_done\_a)\\ = arity'(o_a) = arity(o)$
    \item $G' = G \cup \{ground(has\_done\_a, \chi)\}$
    \item $W' = W \cup \{\langle lb, ub, ground(can\_do\_a, \chi) \rangle\}$
\end{itemize}

\textit{Jerry} cannot deliver the pallet \textit{p2} in the time period required by the user and so the plan is unsolvable.


\subsection{Delay/Advance an Action}\label{comp:delay}


Given a plan $\pi$, a formal question $Q$ is asked of the form:
\begin{quote}
\textit{Why is the operator $o$ with parameters $\chi$ used at time $t$, rather than at least some duration $t'$ earlier/later $t$?}
\end{quote}
For example, given the example plan in Figure~\ref{fig:plan} the user might ask:
\begin{quote}
``Why is \textit{set\_shelf Tom sh1} used at time \textit{8.001}, rather than at least \textit{8} minutes later?''
\end{quote}

A user would ask this type of question when they expected an action to appear earlier or later in a plan. This could happen for a variety of reasons. In domains with resources that are depleted by specific actions, and are replenished by others, such as fuel for vehicles, these questions may arise often. A user might want an explanation of why a vehicle was refueled earlier or later than was expected. In this case the refuel action can be delayed or advanced to answer this question. 

A user might ask the question posed above about our running example because they think that \textit{Tom} is rushing to set up the shelf. \textit{Tom} sets up the shelf \textit{sh1} in preparation for the delivery of the pallet \textit{p2} eight minutes into the plan. However, \textit{Jerry} is not ready to deliver the pallet until the very end of the plan. \textit{Tom} might be able to complete other goals before he is required to set up the shelf for the delivery. 
The reasoning behind the early preparation can be explained by delaying setting up the shelf until it is completely necessary and comparing the HPlan produced with the original solution.

To generate the HPlan, the planning model is compiled such that the ground action $a = ground(o, \chi)$ is forced to be used in time window $w$ which is at least $t'$ before/after $t$.
This compilation is an example of a combination of two other compilations: adding an action (in Section~\ref{comp:add}) and forbidding the action outside of a time window (in Section~\ref{comp:tw1}). The latter enforces that the action can only be applied within the user specified time window, while the former enforces that the action must be applied. The HModel $\Pi'$ is:
$$
\Pi' = \langle \langle Ps', Vs, As', arity' \rangle, \langle Os, I', G', W'\rangle \rangle
$$
where:
\begin{itemize}
    \item $Ps' = Ps \cup \{can\_do\_a, not\_done\_a, has\_done\_a\}$
    \item $As' = \{o_{a}, o_{\neg a}\} \cup As \setminus \{o\} $
    \item $arity'(x) = arity(x), \forall x \in arity$
    \item $arity'(can\_do\_a) = arity'(not\_done\_a) =\\ arity'(has\_done\_a) = arity'(o_{a}) =\\ arity'(o_{\neg a}) = arity(o)$
    \item $I' = I \cup \{ground(not\_done\_a, \chi)\}$
    \item $G' = G \cup\{\begin{array}{ll}ground(not\_done\_a, \chi),\\ground(has\_done\_a, \chi)\end{array}\}$
    \item $W' = W \cup \begin{cases}
before:\langle 0, tReal, ground(can\_do\_a, \chi) \rangle\\
after:\langle tReal, \textit{inf}, ground(can\_do\_a, \chi) \rangle
\end{cases}$
\end{itemize}
where \textit{tReal} is \textit{t} ${\displaystyle \pm }$ \textit{t'} and the new operators $o_{a}$ and $o_{\neg a}$ both extend $o$. The latter with the delete effect $not\_done\_a$, while $o_{a}$ extends $o$ with the precondition $can\_do\_a$ and the add effect $has\_done\_a$; i.e.:
$$
\begin{array}{c}
\Eff_{\dashv}^-(o_{\neg a}) = \Eff_{\dashv}^-(o) \cup \{not\_done\_a\}\\
Pre_{\leftrightarrow}(o_{a}) = Pre_{\leftrightarrow}(o) \cup \{can\_do\_a\}\\
\Eff_{\dashv}^+(o_{a}) = \Eff_{\dashv}^+(o) \cup \{has\_done\_a\}
\end{array}
$$

This ensures that the ground action $a = ground(o_{a}, \chi)$ must be present in the plan between the times $0$ and $tReal$, or $tReal$ and \textit{inf}, depending on the user question, and between those times only. 
In addition, the user selected action is forced to be performed using the same approach as in Section~\ref{comp:add}.

The HPlan produced for the users question is shown in Figure~\ref{fig:tw3plan}. The delayed action \textit{(set\_shelf tom sh1)} is now performed at time 17 which, as the action takes one minute, would allow \textit{Jerry} to unload the pallet . However, \textit{Tom} is blocking \textit{Jerry} from getting to shelf \textit{sh1}. Consequently, \textit{Jerry} has to wait for \textit{Tom} to evacuate the shelf which delays the completion of the delivery by 7.5 minutes. Additionally, it can be seen from the plan that \textit{Tom} does not contribute to the completion of any other goals in the time before setting up shelf \textit{sh1}.


\begin{figure}[h!]
\begin{lstlisting}
0.000: (goto_waypoint tom sh5 sh6)  [3.000]
0.000: (set_shelf_nota jerry sh3)  [1.000]
1.000: (load_pallet jerry p1 sh3)  [2.000]
3.000: (goto_waypoint jerry sh3 sh4)  [5.000]
3.001: (set_shelf_nota tom sh6)  [1.000]
4.001: (goto_waypoint tom sh6 sh1)  [4.000]
8.001: (goto_waypoint jerry sh4 sh5)  [1.000]
9.002: (goto_waypoint jerry sh5 sh6)  [3.000]
12.002: (unload_pallet jerry p1 sh6)  [1.500]
13.503: (load_pallet jerry p2 sh6)  [2.000]
17.000: (set_shelf_a tom sh1)  [1.000]
18.000: (goto_waypoint tom sh1 sh2)  [4.000]
22.001: (goto_waypoint jerry sh6 sh1)  [4.000]
26.001: (unload_pallet jerry p2 sh1)  [1.500]
\end{lstlisting}
\caption{The HPlan with the action \textit{(set\_shelf Tom sh1)} performed at least 8 minutes later than it was originally performed.}
\label{fig:tw3plan}
\end{figure}

\subsection{Composition of Compilations}\label{comp:composition} 

Each compilation defined in this section can be used to answer one of the formal questions from the contrastive taxonomy that was identified in our user study.
However, through the iterative approach described in Definition~\ref{def:negprob} the set of questions that can be answered is not restricted to the formal questions found in the Contrastive Taxonomy.
Instead a composition of these compilations can be used to produce more complex constraints that answer a much wider set of questions. More complex questions that are not easy to specify without refinement can be posed through the iterative process of query and feedback.
Moreover, humans themselves have trouble understanding a decision from a ``one shot'' justification, they are more likely to comprehend a decision through a conversational process resulting in a more complete explanation~\cite{hil90}.

For example, consider the multiple questions asked in sequence $\text{q}_1, \text{q}_2, \ldots, \text{q}_n$ that have constraints $\phi_1, \phi_2, \ldots, \phi_n$. The user could instead have asked a single complex question $\text{q}_x$ that has the corresponding constraint $\phi_x$:
$$\Pi \times \phi_x = ((\Pi \times \phi_1) \times \phi_2) \ldots \times \phi_n$$
This compilation $\Pi \times \phi_x$ would have produced the same HPlan as the final HPlan resulting from the iterative process. However, this assumes that the user knows the question $\text{q}_x$ in advance. In practice, each question might have been prompted by the result of the previous iteration, allowing the user to refine their question during the process. 

This refinement also has the consequence that the user is able to pose questions about artefacts and processes of the plan that are not obviously representable in the model. As an example a user might want to know why the pallet \textit{p1} took too long to be transported from shelf \textit{sh3} to \textit{sh6}. This question refers to the time between two ground actions in the plan, and the vocabulary of the model does not allow a constraint on this time to be expressed. However, through the iterative process it is possible to incrementally converge to a set of constraints that force these two events to happen closer together in time. Moreover, it is possible to follow this process without explicitly and immediately defining the duration that the user considers to be "too long", instead allowing the user to refine their question as their understanding grows.

That these compilations can be used to produce more complex constraints that answer a much wider set of questions can be stated more strongly as: 
for every valid plan $\pi$ for a model $\Pi$, there exists a sequence of constraints, $\phi_1,\ldots, \phi_n$, such that $\pi$ is the only valid plan for $((\Pi\times\phi_1),\ldots\phi_n)$. 
Trivially, we can achieve any expected HPlan by iteratively applying the replace compilation shown in~\ref{comp:replace}. In practice our user study in Section~\ref{sec:userstudy} showed that by using a variety of questions, the users converged quickly on their desired plans.

\subsection{Justified User Suggestions}\label{comp:just}

For a planning model $\Pi$ with goals $G$ there can be many valid plans that satisfy $G$, which we call the space of plans for a planning model. Generating the plan that will best satisfy the user at each stage of the negotiation process is not guaranteed. Firstly, temporal planning tasks are intractable and in fact in the general case belongs to the complexity class EXPSPACE-complete~\cite{rin07b} and the introduction of numeric variables makes the problem undecidable~\cite{hel02}. Our approach is limited by these impediments, just as a human might try to explain a decision they have made. Secondly, even should an optimal plan be returned, it might not be the plan that most increases the user's understanding of the problem, or provides the fastest route to concluding a negotiation.

However, while it might not be possible to completely specify the metric of user satisfaction in a plan, it is possible to make some assumptions. One reasonable assumption is that the user wants to see their suggestion have an impact in the plan. When a user questions why an action was not used in the plan, a hypothetical plan containing that action would not be satisfactory if its effects are immediately undone, or it does not contribute towards a goal. Fink and Yang~\citeyear{fin92} use ``justified actions'' to refer to actions that are necessary for achieving a goal. That is to say an action $B$ is justified in a plan $\pi$ if there is a sequence of actions in $\pi$ where $a_1, ..., B, ..., a_n \models G$ and if we remove the action $B$ then, $a_1, ..., a_n \not \models G$. Similarly, a valid plan is \textit{perfectly justified} if it does not have any legal proper subplan that also achieves the goal.

Our compilations alone do not guarantee that the action suggested by the user is justified in the resultant plan. The resultant plan should show, if possible, the user's suggestion make a purposeful contribution to the satisfying the goals of the problem.
In future work we aim to build on the compilations strategy in Section~\ref{comp:keycausal} to develop compilations that ensure that user suggestions are used purposefully within the plan. That is, to enforce that the resultant plan is perfectly justified, or that at least the user suggestion appears in every valid subplan.

A second open question is whether the assumption is indeed reasonable. While it might seem clear that the user should be interested in their suggestion contributing towards the goal, it should also be considered that the goal $G$ does not necessarily capture all of the user's preferences and interests in the problem. As an example, the user might be interested in investigating the space of plans to determine if there remains enough flexibility to add additional exploratory actions, or achieve goals that they do not yet know how to concretely specify.
\section{Explainable Planning as an Iterative Process}\label{sec:iter}

In this section we present a framework within which we have implemented the iterative model restriction process described in Section~\ref{subsec:problem} and instantiated through the compilations described in Section~\ref{sec:comp}. We use an approach we call Explainable Planning as a Service (XAIP-as-a-service). This paradigm is motivated by Definition~\ref{def:negprob} and consists of an iterative conversational process between the user and the planning system. The user asks contrastive questions about a presented plan and receives explanations until the user terminates the process. Explainable Planning \textit{as a service} means implementing the approach as a wrapper around an existing planning system that takes as input the current planning problem and domain model, the current plan, and the user's question. It has the ability to invoke the existing planning system on hypothetical problems in order to address contrastive questions. In Section~\ref{sec:eval} we present the results of the user study conducted with this XAIP-as-a-service framework, alongside an evaluation of the computational costs and effectiveness of the compilations.

\begin{figure}[t!]
    \centering
    \includegraphics[width=0.7\linewidth]{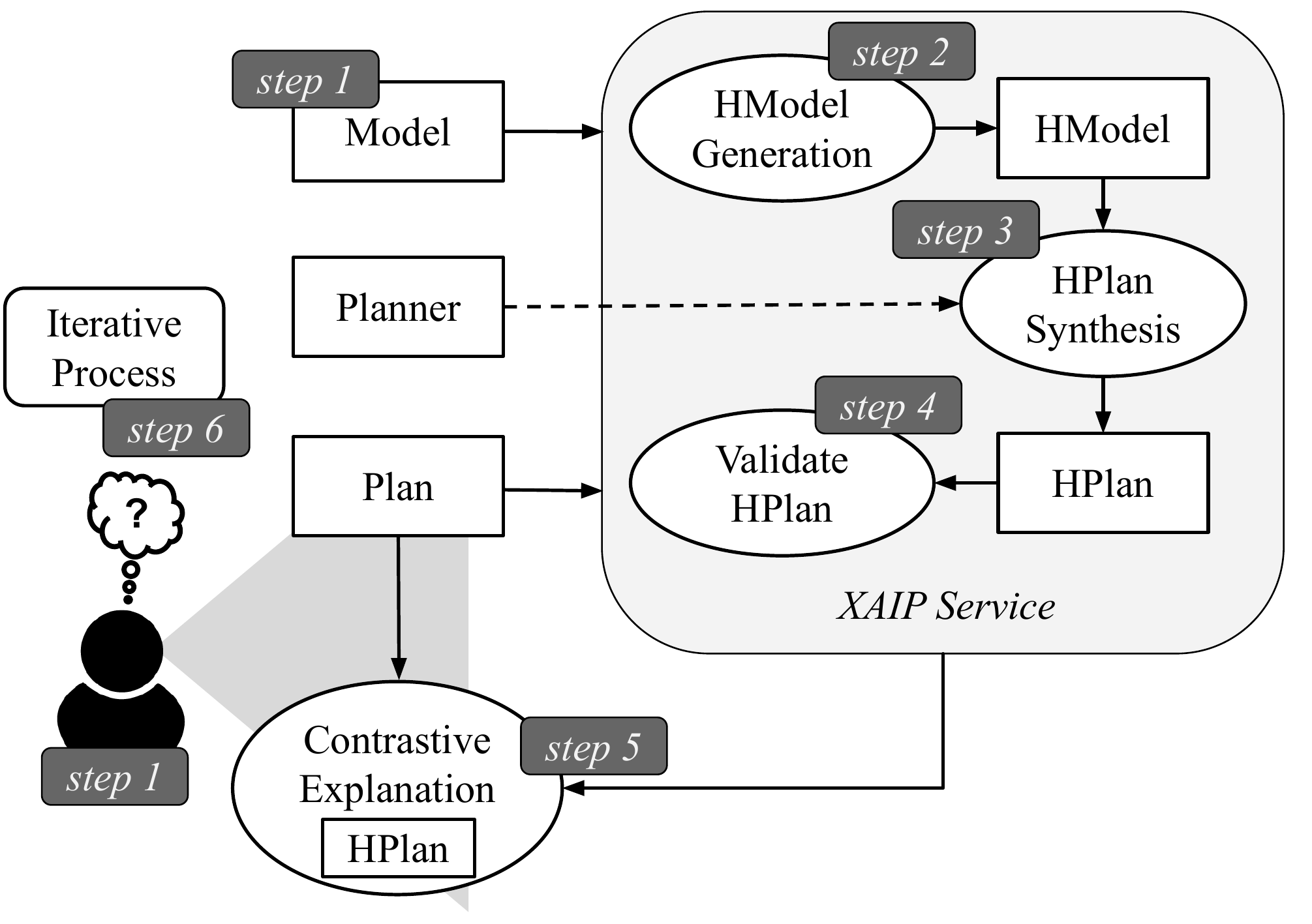}
    \\
    \caption{ Proposed approach for Explainable Planning as a service 
    }
    \label{fig:overview}
\end{figure}

The XAIP-as-a-service paradigm has the benefit that the known and trusted planner and model can be used to provide explanations. At each step a new hypothetical plan is created using the planner chosen by the user, and is validated against the user's original trusted model. As described in Definition~\ref{def:constraintaddition} a model restriction satisfies the condition that any plan for the restricted model is also a plan for the original model. Updates to the model serve to force decisions from the planner and so explore the consequences of those decisions. Figure~\ref{fig:overview} summarises the implementation described in Definition~\ref{def:negprob} and user interaction illustrated in Figure~\ref{fig:tree2}, following these steps:

\begin{enumerate}[align=parleft, label=\textbf{Step \arabic*:}, leftmargin=*]
    \item The \textit{XAIP Service} takes as input the planning problem and the domain, the plan, and a question from the user.
    \item The contrastive question implies a \textit{hypothetical model} characterised as an additional set of constraints on the actions and timing of the original problem.  These constraints can then be compiled into a revised domain model (HModel) suitable for use by the original planner.
    \item The \textit{original planner} uses the HModel as input to produce the \textit{hypothetical plan} (HPlan) which contains the user suggestion.
    \item The XAIP Service validates the HPlan according to the original model.
    \item The original plan and HPlan are shown to the user, with differences highlighted.
    \item The user may choose to repeat the process from Step 1, selecting the original model or any HModel and a new question.
\end{enumerate}

\subsection{Implementation details}\label{sec:imp}

\begin{figure}
    \centering
    \includegraphics[width=0.75\columnwidth]{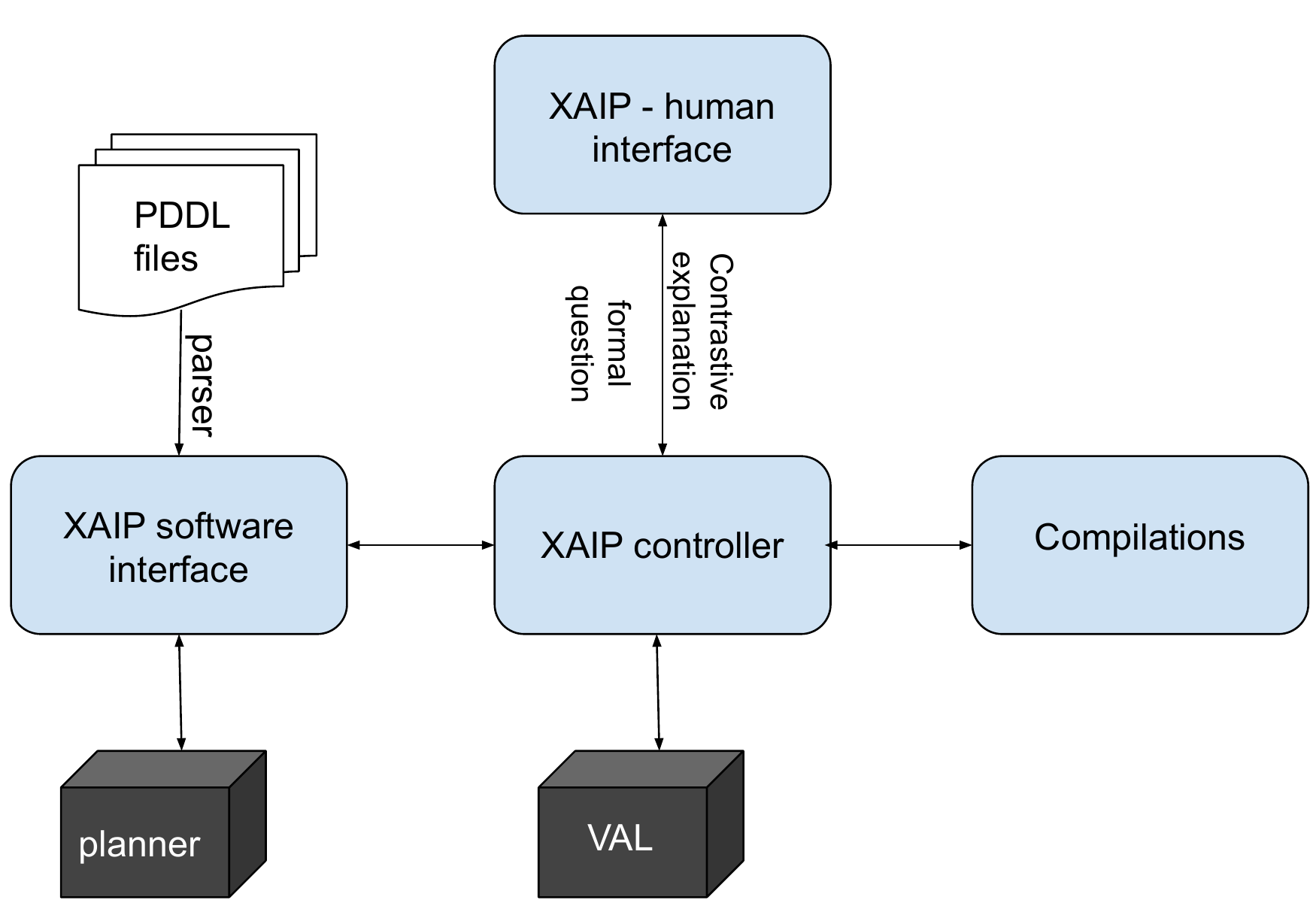}
    \caption{Architecture of the framework for Explainable Planning as a service. }
    \label{fig:framework}
\end{figure}

\begin{figure}
\centering
\begin{subfigure}[b]{.75\textwidth}
  \centering
  {\includegraphics[width=1\linewidth]{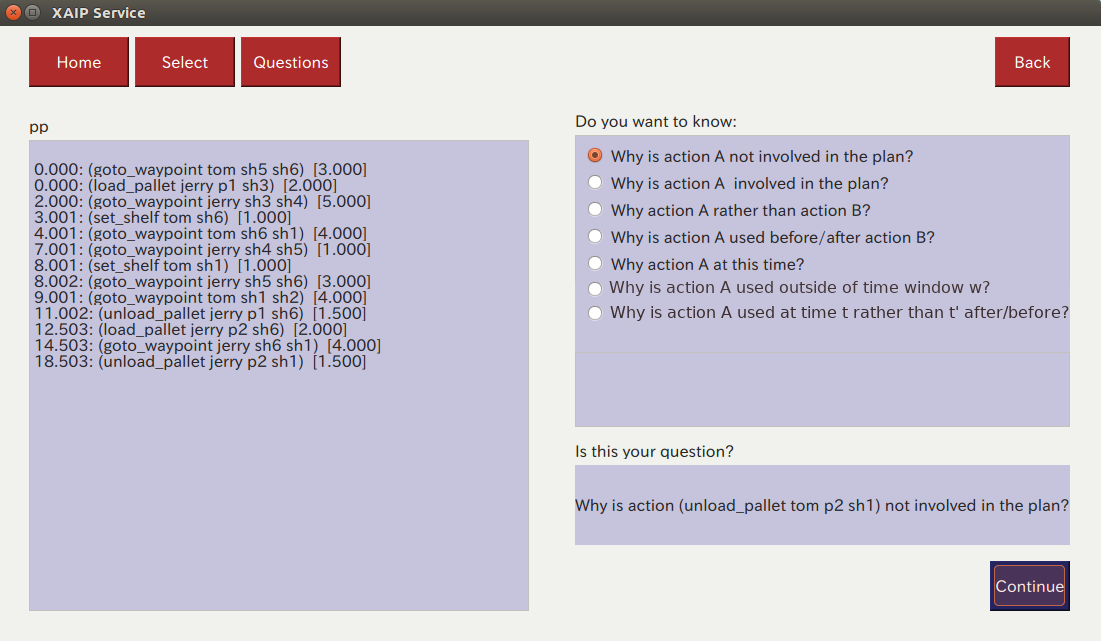}}
  \caption{Plan visualisation and question selection.}
  \label{fig:guia}
\end{subfigure}
\begin{subfigure}[b]{.75\textwidth}
  \centering
  \includegraphics[width=1\linewidth]{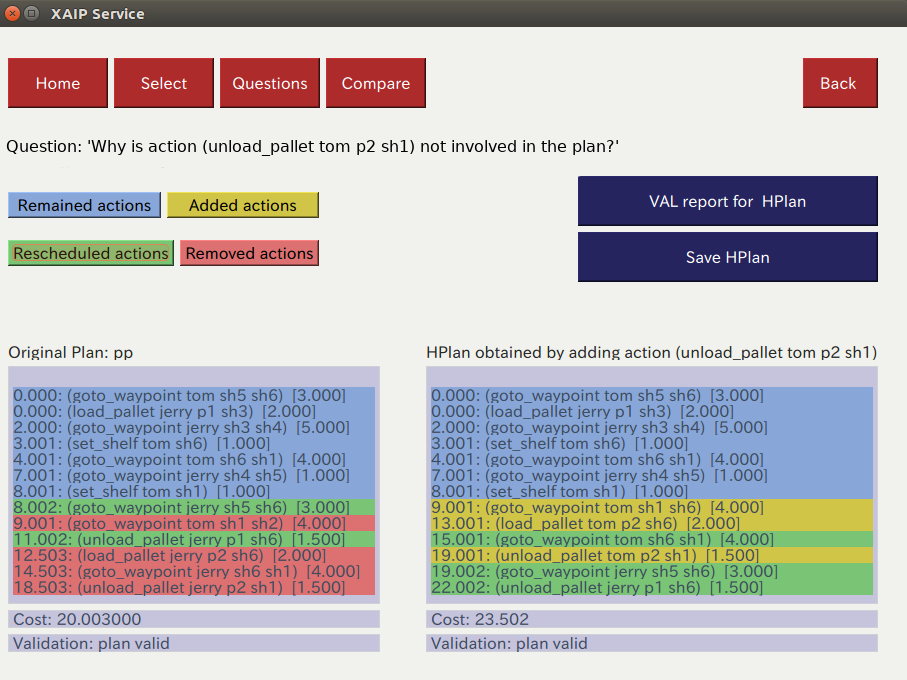}
  \caption{Explanation visualisation.}
  \label{fig:guib}
\end{subfigure}
\caption{Screenshots of the graphical user interface of the \textit{XAIP Service} framework. The first image displays the original plan, a user can formulate a question about the plan using the dialogue. The second image displays the side by side comparison of the original plan and the HPlan produced from the user's question. The differences in the plans are highlighted. Actions that are unchanged are coloured blue, those that are new in the HPlan are coloured yellow (only appear in the HPlan), actions are coloured green if they appear in both the plan and the HPlan but have different dispatch times, and actions are coloured red if they are removed from the plan (do not appear in the HPlan).}
\label{fig:gui}
\end{figure}

We implemented the XAIP-as-a-Service as modular framework for domains and problems written in PDDL2.1.~\cite{fox03}. This framework interfaces with any planner capable of reasoning with PDDL2.1, such as POPF~\cite{col10}, Metric-FF~\cite{hof03}, OPTIC~\cite{col12}, etc. The architecture of the framework is illustrated in Figure~\ref{fig:framework}. Interaction with a user is enabled through a graphical user interface, implemented with Qt-Designer. The modularity of the framework decouples the interfaces for providing user questions (\textit{Step 1}), synthesising the HModel (\textit{Step 2}), interfacing with the planner (\textit{Step 3}), and returning HPlans to the user (\textit{Step 5}).\footnote{All source code and example domain and problem files are open source and available online:\\ \url{https://github.com/KCL-Planning/XAIPFramework}.}

The process is controlled by the \textit{XAIP controller} module of Figure~\ref{fig:framework}. This module uses the interfaces of each other module of the framework described below. The controller is also responsible for validating hypothetical plans against the original domain (\textit{Step 4}), using the plan validation system \textit{VAL}~\cite{Dlong_val}.

\subsubsection{XAIP-Human Interface}\label{sec:xaip_human_interface}

The XAIP-human interface module of Figure~\ref{fig:framework} implements \textit{Step 1}, and \textit{Step 5} of the XAIP-as-a-service process. The module consists of a Qt interface through which the user is able to select an existing model (either the original model or a previous HModel), construct a question, and view the resulting HPlan.

The questions that can be constructed by the interface consist of those that are defined in the Contrastive Taxonomy in Table~\ref{tab:user_questions} in Section~\ref{sec:taxonomy}. A screenshot of the interface is shown in Figure~\ref{fig:guia}. In this screenshot the user has already selected a model and plan for which to ask a question, and selected the formal question ``Why is action A not used in the plan, rather than being used?'' (FQ1). The user has populated the details of the question so that the final question reads:
\begin{quote}
    ``Why is the action \textit{(unload\_pallet top p2 sh1)} not involved in the plan?''
\end{quote}

The interface presents the HPlan to the user, as shown in Figure~\ref{fig:guib}. In this plan comparison both plans are shown side-by-side with differences highlighted. These differences include added actions which were not present in the original plan, actions which have been rescheduled/reordered, and actions which have been removed. The user is also able to compare the costs of each plan, and view the validation report produced by VAL.
In Figure~\ref{fig:guib} the action that was suggested by the user, \textit{(unload\_pallet top p2 sh1)}, appears in the HPlan at time $19.001$.

\subsubsection{XAIP Software Interface and Compilations}

The \textit{XAIP software interface} module implements \textit{Step 3} of the XAIP-as-a-service process, interacting with the planner to produce hypothetical plans. This is done by parsing the original domain and problem and storing the resultant model in an internal knowledge base. This knowledge base contains a collection of models that can be queried or passed to the planner.

The \textit{Compilation} module implements \textit{Step 2} by providing an interface that given a formal question and model, applies the model restriction to produce the HModel. The Compilations module implements the model restriction in Section~\ref{sec:comp}. When triggered by the event of the user selecting a model and formal question through the XAIP Human Interface, the Controller will fetch the model from the XAIP Software Interface, pass the model and question to the Compilations module, and store the resulting model in the XAIP Software Interface Knowledge Base.

\subsubsection{Modularity}

The design of the framework's architecture allows the components to be independently adapted to better fit different XAIP scenarios. For example, the XAIP-Human Interface can be adapted to explanations for a robotic agent or an augmented reality setting~\cite{cha18b}. The explanation visualisation can be adapted to provide different information to the user, for example to illustrate discrepancies in the model highlighting what prevents the planner from solving the restricted model~\cite{sre19}.  The Compilations module can be replaced or extended to add new compilations that accommodate niche questions that are specific to the scenario.

An actualised example of a modification to the explanation framework is demonstrated by comparing the ethics of plans~\cite{kra20}. In this work Krarup et al. modified the architecture of the framework with an additional module called the Ethical Explanation Generator. With extra information about the ethical attributes of the model (the extrinsic value of actions, utilities of facts, etc.) and the moral principle, the Ethical Explanation Generator shows whether a plan is permissible under the principle and provides a causal explanation. The resulting extended framework allows users to compare the ethics of plans within the iterative process.
\section{Evaluation}\label{sec:eval}

Our evaluation falls into two parts: we evaluate the performance of the compilation of constraints by examining the planning time and plan quality produced for a large sample of problems, and we also present the user study that explores the value of the iterative process of plan explanation. The latter evaluation is based on observed interactions with an implemented system and is, therefore, more qualitative in style than the former evaluation. Nevertheless, both evaluations together serve to support our claims that the approach we have described provides a paradigm that allows users to usefully explore explanations of plans, by asking contrastive questions and being supplied plans in response to the constraints implied by those questions.

\subsection{Performance Evaluation}\label{sec:eval_perf}

Compilations can increase the difficulty of solving a problem so that it can no longer be solved in a reasonable time. For example, LTL constraints represented as B{\"u}chi-Automata and compiled into PDDL can scale very poorly and would not be appropriate for a real-time iterative dialogue with our system~\cite{ede06}. In order to evaluate the impact of the compilations listed in Section~\ref{sec:comp} we perform two experiments. The first is to evaluate the effect of single compilations on planning time, and the second is to determine the impact of multiple iterative compilations on planning time.

Explanation is a form of social interaction and and takes the form of a conversation~\cite{hil90}.
If it takes substantially longer to answer the explanatory question a user poses than to generate the original solution, it might be unreasonable to expect a user to want to wait for the explanation (depending on the context). In this case, the explanation process would be impractical in real world settings and that would undermine the value of the paradigm we have created of explanation as an iterative, conversational process. Moreover, it must not get exponentially harder to answer multiple iterations of questions.

The time to apply the compilations and generate the HModel is negligible for all the cases we consider so we do not take this into account in our evaluation.

We used four temporal domains from the recent ICAPS international planning competitions (IPC)~\cite{lon03} in our experiments. The IPC produces a new set of benchmark domains each year to test the capabilities and progress made by AI planners for different types of problems. We selected domains to be varied in what they modelled and the most interesting in terms of explainability. These are the \textbf{ZenoTravel}, \textbf{Depots} (IPC3), \textbf{Crew Planning} and \textbf{Elevators} (IPC8) domains. In both experiments we used the Crew Planning and Elevators domains, in the first experiment we used the Depots domain and used the ZenoTravel domain in the second. We explain the reason for the difference in the domains in the design of the second experiment in Section~\ref{sec:eval_iteration}.

ZenoTravel is a logistics domain which models a scenario in which a number of pilots have to deliver a number of packages by plane. The planes can travel at different speeds which consumes fuel at different rates. The pilots must fly their planes at the correct speeds to minimise the time whilst maintaining the fuel to successfully deliver all of the packages.

The Depots domain combines the transportation style problem of Logistics with the well-known Blocks domain. In this domain crates must be stacked in a certain order at their destinations. Trucks are used to move the crates between locations and hoists are used to stack the crates.

The Crew Planning domain is designed to plan the itinerary of a crew on the International Space Station over a period of days. The crew have to complete tasks critical to maintenance of the station such as configuring thermals and facilitating the delivery of payloads, whilst also performing the tasks necessary for survival such as eating and sleeping.

In the Elevators domain there are multiple elevators, with different speeds, that service portions of different building blocks. Each of the blocks share at least one mutual floor. The goal is to get a set of people to their desired floors using the elevators.

\subsubsection{Compilation Impact by Question}\label{sec:eval_comp}
\paragraph{Purpose}
We first designed an experiment to evaluate the impact each type of compilation in Section~\ref{sec:comp} has on the time taken to find a solution and the quality of the resultant solution. We designed this experiment to show that explanations can be produced in a reasonable time. 
We also wanted to see what effect compilations have on the quality of the solution. An explanation generated from an inefficient HPlan would not be satisfactory to the user. Although we cannot evaluate the quality of any given solution in the context of it's optimal solution, the large set of results for any problem will allow us to draw conclusions about possible inefficiencies.
We also looked to determine whether there were any questions, or question types, for which it is harder to produce HPlans and so took longer to find solutions. 

\paragraph{Design}
For each of the domains (Crew Planning, Depots, and Elevators) we selected four problems of varying complexity provided by the same IPC benchmark. We first used the planner to find the solutions to these as the control. Then, for each question type categorised in the Contrastive Taxonomy, we randomly generated the formal question and generated and solved the HModel, we repeated this ten times. All tests used a Core i7 1.9GHZ machine, and 16GB of memory. We used the POPF~\cite{col10} planner and recorded all solutions found in the time allocated to test the effect compilations have on optimisation and solution quality.
However, for the purpose of this experiment it is sufficient to evaluate if there are any obvious inefficiencies in our compilation approach, not to try to find the optimal plans for each constrained problem.

We conducted preliminary tests to determine the amount of planning time to allocate to each instance. We found that for each of the problems 3 minutes planning time was sufficient, other than problem 10 for the Depots domain which required 6 minutes.
To illustrate the efficiency of our compilations, the experiment required many tests of each type of compilation, so we chose the minimum sufficient planning time.

It was not practical to evaluate our approach with questions composed by humans. Therefore we randomly generated the questions used in our experiments. 
To ensure that the questions made sense, we had to take slightly different approaches to generating each question type. 
For each formal question other than FQ1 and FQ3, the actions were randomly selected from the original plan found from the appropriate model. 
We took the extra precaution, with FQ5, to ensure that the order of the selected actions in the original plan was the opposite of the new order enforced by the question. 
For the formal questions FQ4, FQ6, and FQ7, time windows were also generated. The lower bound was generated using a pseudo-random number generator, constrained to within the original plan time. The upper bound was formed by first generating a number between 1.5 and 4 and then multiplying the number by the duration of the selected action. 
This produced a spectrum of time windows from those that are very tight to those that are quite relaxed, which mimicked how a user might ask these types of questions.
To generate the formal questions FQ1 and FQ3 we had to create questions with actions that were not already present in the original plan. To do this we created a list of the possible grounded actions in the model and then randomly selected one of these grounded actions, that was not present in the original plan, to form the question. 
For FQ3 we also randomly selected an action from the original plan to replace. We then verified that the randomly selected (replacement) action was applicable in the state directly before the action chosen to be replaced. If the action was not applicable, a new action was generated and the process repeated until an applicable action was found. The rest of the compilation process then continued as normal.

The questions generated in this way might not be ones users would ask, being artificially constructed. However, evaluating how users interact with our framework was not the purpose of these experiments (that we consider in Section~\ref{sec:userstudy}), but the broad coverage of generated questions gives a reasonable assessment of the performance of the planner on compiled HModels. 

\begin{figure}
\begin{subfigure}{.5\textwidth}
  \centering
  \includegraphics[width=1.0\linewidth]{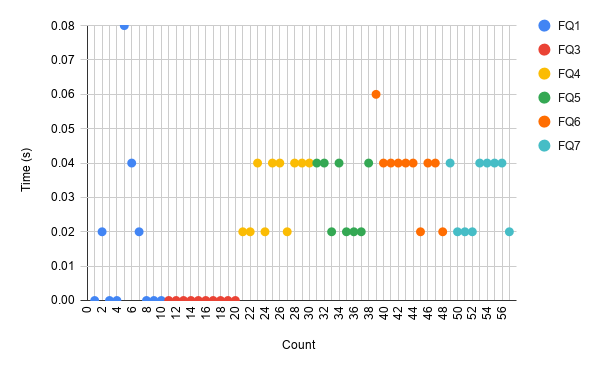}
  \caption{Crew Planning Problem 1, Literals 30, \\Planning Time 0}
  \label{fig:easy_sfig1}
\end{subfigure}%
\begin{subfigure}{.5\textwidth}
  \centering
  \includegraphics[width=1.0\linewidth]{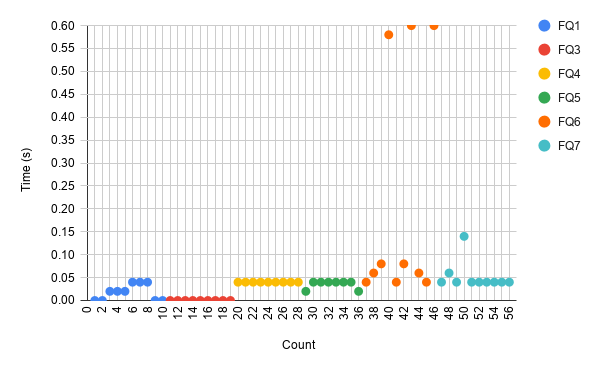}
  \caption{Crew Planning Problem 2, Literals 38, Planning Time 0}
  \label{fig:easy_sfig2}
\end{subfigure}
\begin{subfigure}{.5\textwidth}
  \centering
  \includegraphics[width=1.0\linewidth]{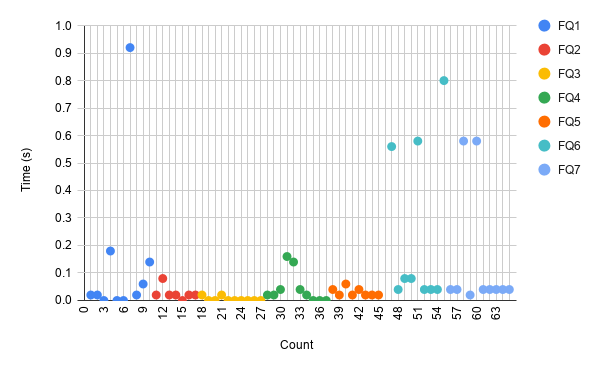}
  \caption{Depots Problem 1, Literals 44, \\Planning Time 0}
  \label{fig:easy_sfig3}
\end{subfigure}
\begin{subfigure}{.5\textwidth}
  \centering
  \includegraphics[width=1.0\linewidth]{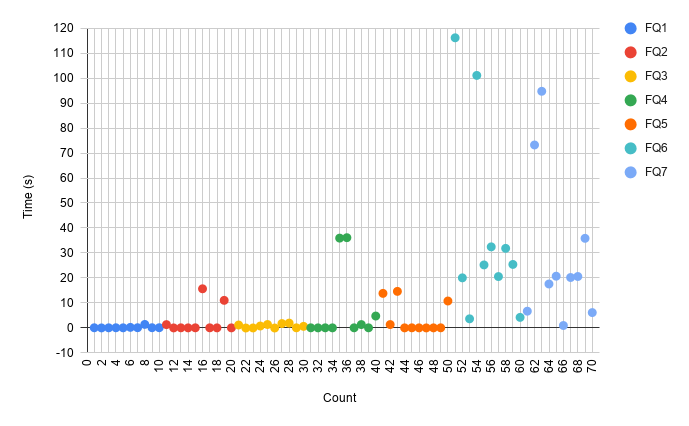}
  \caption{Elevators Problem 1, Literals 86, Planning Time 0.1}
  \label{fig:easy_sfig4}
\end{subfigure}
\caption{Scatter graph comparing the planning times of each compilation type over the simplest problems in each of the tested domains.}
\label{fig:performance_graphs_easy_scatter}
\end{figure}

\begin{figure}
\begin{subfigure}{.5\textwidth}
  \centering
  \includegraphics[width=1.0\linewidth]{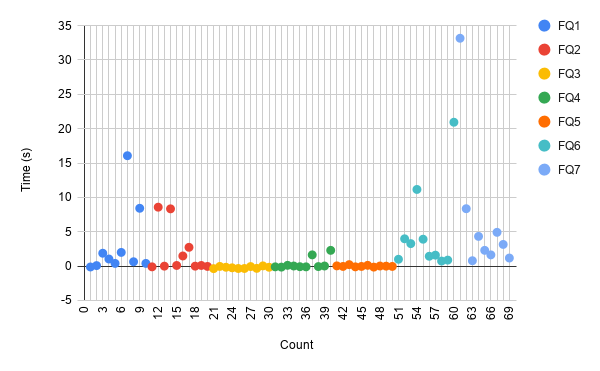}
  \caption{Elevators Problem 5, Literals 138, \\Planning Time 0.38}
  \label{fig:hard_sfig1}
\end{subfigure}%
\begin{subfigure}{.5\textwidth}
  \centering
  \includegraphics[width=1.0\linewidth]{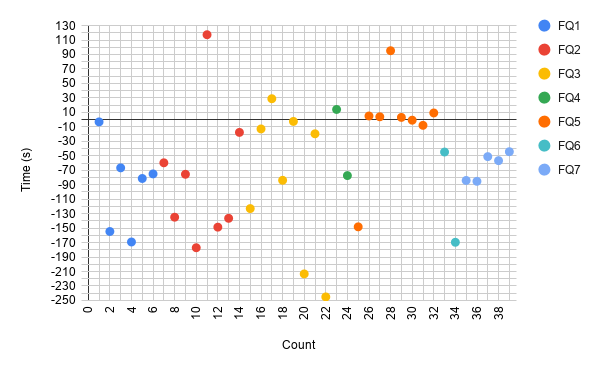}
  \caption{Depots Problem 10, Literals 192, Planning Time 245.06}
  \label{fig:hard_sfig2}
\end{subfigure}
\begin{subfigure}{.5\textwidth}
  \centering
  \includegraphics[width=1.0\linewidth]{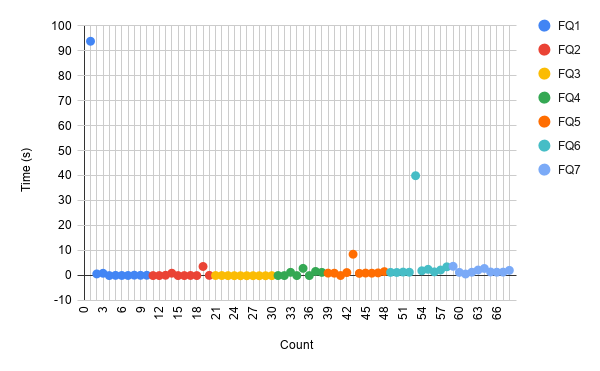}
  \caption{Depots Problem 13, Literals 224, \\Planning Time 0.12}
  \label{fig:hard_sfig3}
\end{subfigure}
\begin{subfigure}{.5\textwidth}
  \centering
  \includegraphics[width=1.0\linewidth]{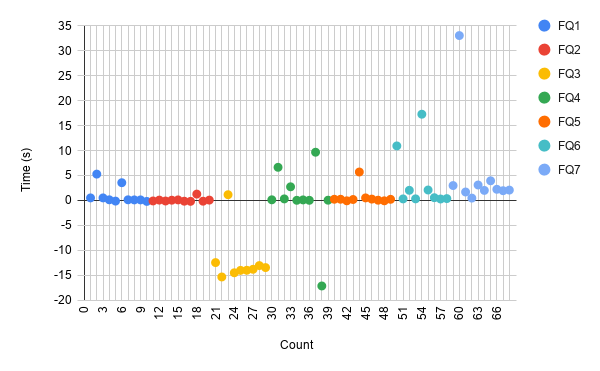}
  \caption{Crew Planning Problem 20, Literals 270, Planning Time 17.45}
  \label{fig:hard_sfig4}
\end{subfigure}
\caption{Scatter graph comparing the planning times of each compilation type over the hardest problems in each of the tested domains.}
\label{fig:performance_graphs_hard_scatter}
\end{figure}

\begin{figure}
\begin{subfigure}{.5\textwidth}
  \centering
  \includegraphics[width=1.0\linewidth]{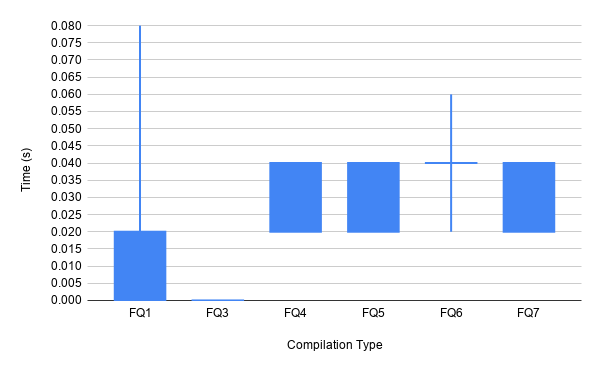}
  \caption{Problem 1, Literals 30, Planning Time 0}
  \label{fig:sfig1}
\end{subfigure}%
\begin{subfigure}{.5\textwidth}
  \centering
  \includegraphics[width=1.0\linewidth]{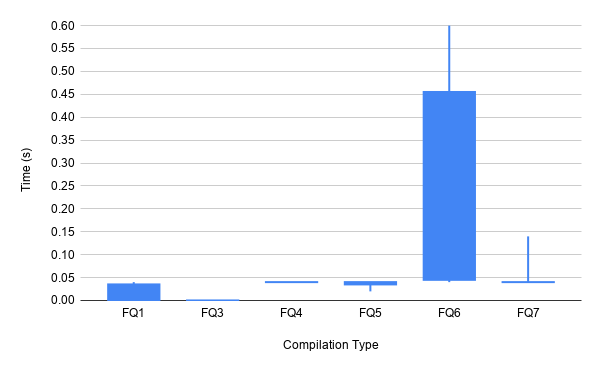}
  \caption{Problem 2, Literals 38, Planning Time 0}
  \label{fig:sfig2}
\end{subfigure}
\begin{subfigure}{.5\textwidth}
  \centering
  \includegraphics[width=1.0\linewidth]{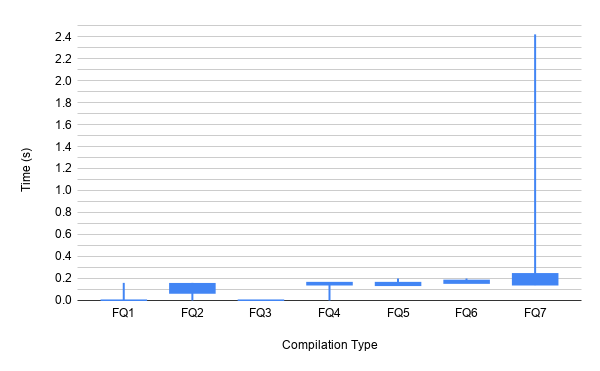}
  \caption{Problem 5, Literals 62, Planning Time 0}
  \label{fig:sfig3}
\end{subfigure}
\begin{subfigure}{.5\textwidth}
  \centering
  \includegraphics[width=1.0\linewidth]{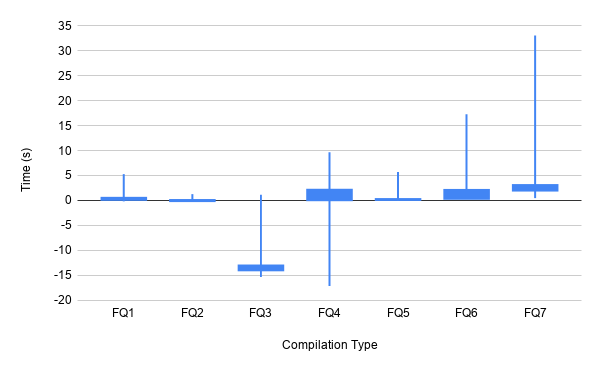}
  \caption{Problem 20, Literals 270, Planning Time 17.45}
  \label{fig:sfig4}
\end{subfigure}
\caption{Box and whisker plots comparing the planning times of each compilation type in the Crew Planning domain over four problems.}
\label{fig:performance_graphs_box}
\end{figure}

\begin{figure}
\begin{subfigure}{.5\textwidth}
  \centering
  \includegraphics[width=1.0\linewidth]{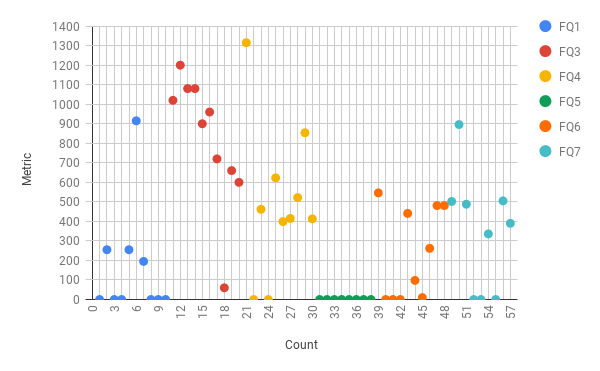}
  \caption{Crew Planning Problem 1, Literals 30, Metric 1440}
  \label{fig:easy_metric_sfig1}
\end{subfigure}%
\begin{subfigure}{.5\textwidth}
  \centering
  \includegraphics[width=1.0\linewidth]{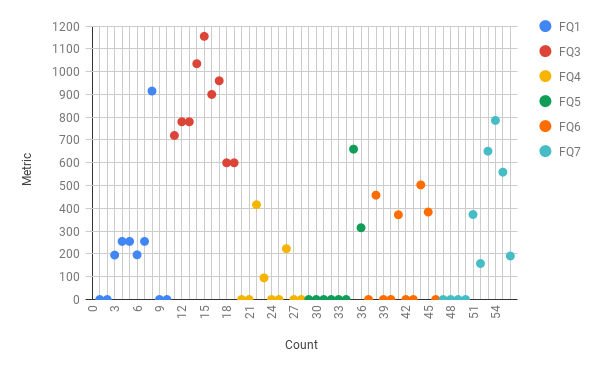}
  \caption{Crew Planning Problem 2, Literals 38, Metric 1440}
  \label{fig:easy_metric_sfig2}
\end{subfigure}
\begin{subfigure}{.5\textwidth}
  \centering
  \includegraphics[width=1.0\linewidth]{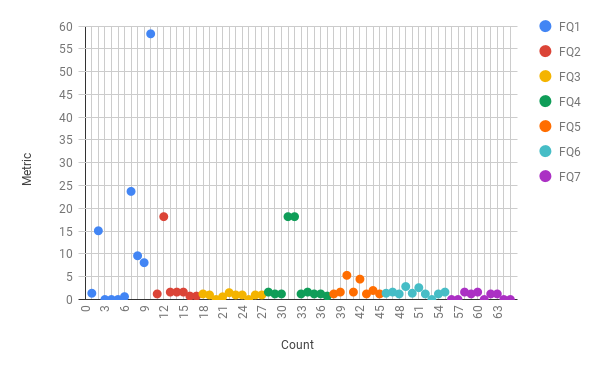}
  \caption{Depots Problem 1, Literals 44, Metric 53.182}
  \label{fig:easy_metric_sfig3}
\end{subfigure}
\begin{subfigure}{.5\textwidth}
  \centering
  \includegraphics[width=1.0\linewidth]{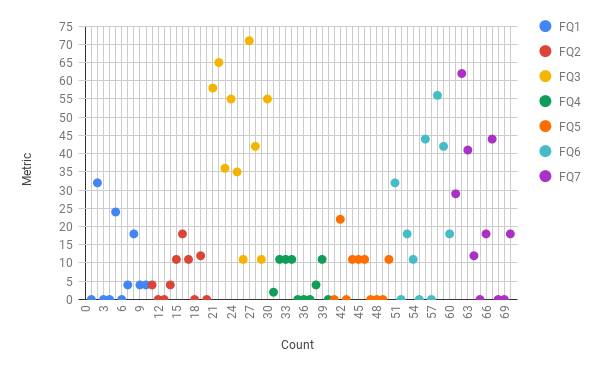}
  \caption{Elevators Problem 1, Literals 86, Metric 80.001}
  \label{fig:easy_metric_sfig4}
\end{subfigure}
\caption{Scatter graph comparing the metrics of each compilation type over the simplest problems in each of the tested domains.}
\label{fig:performance_graphs_easy_metric_scatter}
\end{figure}

\begin{figure}
\begin{subfigure}{.5\textwidth}
  \centering
  \includegraphics[width=1.0\linewidth]{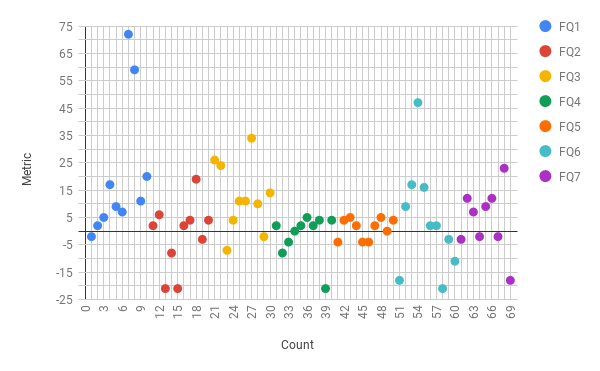}
  \caption{Elevators Problem 5, Literals 138, Metric 90.002}
  \label{fig:hard_metric_sfig1}
\end{subfigure}%
\begin{subfigure}{.5\textwidth}
  \centering
  \includegraphics[width=1.0\linewidth]{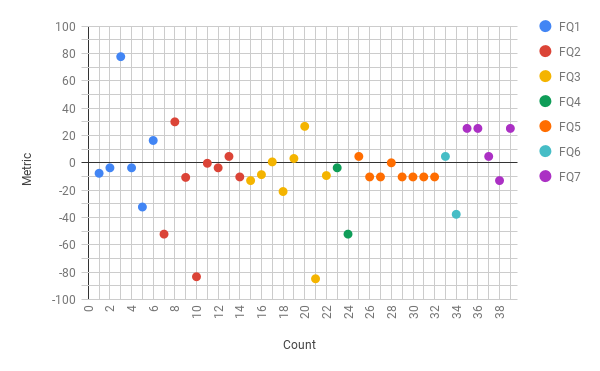}
  \caption{Depots Problem 10, Literals 192, Metric 256.171}
  \label{fig:hard_metric_sfig2}
\end{subfigure}
\begin{subfigure}{.5\textwidth}
  \centering
  \includegraphics[width=1.0\linewidth]{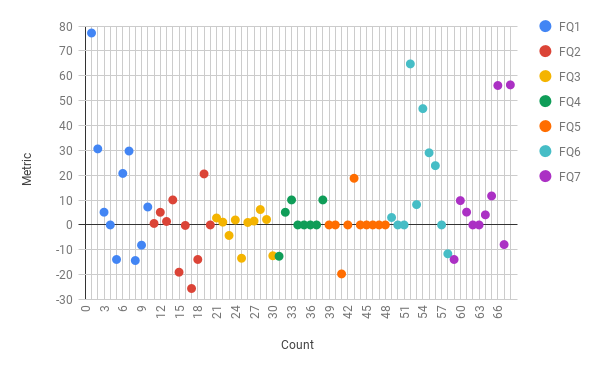}
  \caption{Depots Problem 13, Literals 224, Metric 91.601}
  \label{fig:hard_metric_sfig3}
\end{subfigure}
\begin{subfigure}{.5\textwidth}
  \centering
  \includegraphics[width=1.0\linewidth]{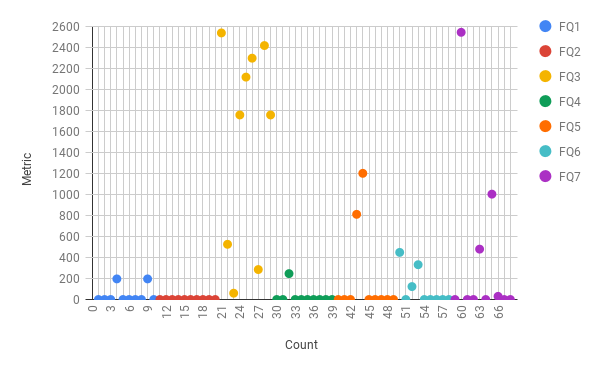}
  \caption{Crew Planning Problem 20, Literals 270, Metric 2880.001}
  \label{fig:hard_metric_sfig4}
\end{subfigure}
\caption{Scatter graph comparing the metrics of each compilation type over the hardest problems in each of the tested domains.}
\label{fig:performance_graphs_hard_metric_scatter}
\end{figure}

\paragraph{Results}
A subset of the results of this experiment is shown in Figures~\ref{fig:performance_graphs_easy_scatter},~\ref{fig:performance_graphs_hard_scatter},~\ref{fig:performance_graphs_box},~\ref{fig:performance_graphs_easy_metric_scatter}, and~\ref{fig:performance_graphs_hard_metric_scatter}. 
The results in these figures are a representative sample of the entire population of results and illustrate the performance characteristics we evaluated with this experiment. The full results of this experiment are available at~\url{https://tinyurl.com/xairesults}.

Figures~\ref{fig:performance_graphs_easy_scatter} and~\ref{fig:performance_graphs_hard_scatter} demonstrate that our compilation approach does not significantly impact planning time. 
These graphs include results from every domain we evaluated as well as multiple problems and show that on average the planning time is not critically affected over multiple domains and problem variants. 
We show the easiest and hardest domains and problems, to reveal how the compilations effected planning time at the extremes of the range of difficulty of the problems.

Figures~\ref{fig:performance_graphs_easy_scatter} and~\ref{fig:performance_graphs_hard_scatter} each contain four scatter graphs. The former containing the results of what we consider to be the easiest problems to solve in our test set, and the latter the most difficult. 
We classified the degree of difficulty for a problem as its size (number of literals in the problem) and the time taken to find any solution for the problem. However, the difficulty of a problem is only comparable for different problems in the same domain. Some domains are easier to solve than others, regardless of the problem size. 
Therefore, to keep the results representative we selected the easiest and hardest problems to solve for each domain, and the next easiest and hardest domain-problem pairs from any of the three domains.

Each data point on the graphs in Figures~\ref{fig:performance_graphs_easy_scatter} and~\ref{fig:performance_graphs_hard_scatter} corresponds to a compilation, the colour corresponds to compilation type categorised by the Contrastive Taxonomy, the key is displayed on each graph. The horizontal axis of each graph displays an arbitrary count to distinguish between each compilation. 
The vertical axis measures the difference between the time taken to find a plan for a compiled HModel and the time to find the original plan for the original model, in seconds. This means the zero on the vertical axis represents there being no difference in the time to find solution plans between the compiled model and the original model, a positive value means there was an increase in the time taken to find a solution for the compiled model, and the opposite holds for a negative value. 

As these plots are used to demonstrate that there is no significant impact on the planning time for constrained problems, we have not shown any results using further optimising search after the discovery of the first solution. As any optimisations will only increase the planning time, it is unfair to compare the planning time of a heavily optimised plan to one with no optimisations. For example, for a model $\Pi$ a planner might find the plan $\pi$ in 10 seconds with a metric of $M(\pi) = 10$ and then then no further plans within it's 3 minutes of allotted time. For a constrained model $\Pi \times \phi = \Pi'$ a planner might then take 9, 10, and 11 seconds to find plans with metrics of 12, 11.5, and 11 respectively, and nothing further in its 3 minutes. How then should the planning times of the two models $\Pi$ and $\Pi'$ be compared? They both have 3 minutes to find solutions, however the quality of the solutions compared to the optimal solution is not known. It might seem sensible to select the two plans with the closest metrics for comparison. However, the quality of the optimal solutions compared to either the original or constrained problems is not known, one of the discovered solutions could be optimal and the other very sub-optimal. To use a comparison that is well-defined, only the results for the first plan found in the graphs is included in Figures~\ref{fig:performance_graphs_easy_scatter},~\ref{fig:performance_graphs_hard_scatter}, and~\ref{fig:performance_graphs_box}. However, the full table of results, including optimisations, can be found at~\url{https://tinyurl.com/xairesults}.

The majority of points lie close to the horizontal axis showing that the compiled HModels in general are similar to the original models in terms of planning time. The median average increase in planning times for each of the domain-problem pairs are all below 4 seconds with one negative showing an improvement in planning time. This substantiates our claim that there is an insignificant impact on the planning time to solve constrained problems. On average, a user will have to wait less than 4 seconds longer than the time taken to solve the original problem to see the outcome of their question and receive an explanation. The highest and lowest increase in planning time both occur in the Depots domain problem 10, shown in Figure~\ref{fig:hard_sfig2}. The highest comes from a compilation of the formal question FQ2, removing an action, increasing the planning time by 117.4 seconds. The lowest increase in planning time, and in fact improvement in planning time, comes from a compilation of the formal question FQ3, replacing an action, improving the planning time by 245.02 seconds. 
The question types FQ6 and FQ7, the compilation of which is shown in Sections~\ref{comp:tw2} and~\ref{comp:delay} respectively, tend to negatively impact the planning time the most, with a median increase of 1.10 and 1.18 seconds. Whereas the compilations for the question types FQ2 and FQ3 have the least effect on the planning time with an average of 0.02 and 0.00 seconds, respectively. 
We report the median averages to ensure extreme values do not skew the results. 

The median effect on planning time across all problems ranges from -59.64 to 3.74 seconds. The domain-problem pairs that we consider to be easy have a range of 0.02 to 1.2 seconds, and the domain-problem pairs that we consider to be hard have a range of -59.64 to 0.925 seconds. Although this data suggests that compilations applied to the harder problems have a much higher chance of improving the planning time over the easier problems, actually the difficulty of the original problem does not have a significant effect on the planning time of the corresponding constrained problem. The results of a Mann-Whitney U test show that the sets of planning times from the easy and hard problems are statistically equal with $p < 0.05$. This shows that the impact of compilations on the planning time does not grow with the difficulty of the original problem.

Figure~\ref{fig:performance_graphs_box} contains four box-and-whisker plots, comparing the planning times of each compilation type in the Crew Planning domain. Each sub-figure displays the results for each of the problems we tested. This data shows that there is minimal difference between the types of compilations in their impact on the planning time.
These graphs show results from each of the problems for the Crew Planning domain, this exemplifies a typical use case of our approach where a user may have a domain for which they have multiple problems, requiring explanations for each.

Each box and whisker plot corresponds to a data set of 10 compilations of a specific type and problem.
The horizontal axis displays each of the compilation types labelled by their corresponding formal question.
Figures~\ref{fig:sfig1} and~\ref{fig:sfig2} do not have formal question FQ2, removing an action, because no plans could be found in the allocated time, we discuss why this is the case later. 
The vertical axis represents the same as the graphs in Figures~\ref{fig:performance_graphs_easy_scatter} and~\ref{fig:performance_graphs_hard_scatter}. 

Each box in the plot represents the interquartile range (IQR) of the difference in planning times; that is, the middle 50 percent of planning times for HModels generated from one compilation type. The whiskers represent the largest and smallest difference in the planning times. The results suggest that the impact the compilations have on planning time is quite inconsequential, and that there are not any compilation types that are substantially more difficult.

The planning times for HModels generated from each compilation type are consistent across their problems. This can be seen with the overlapping interquartile ranges on most data sets. This shows that there is little variation in planning time between the types of compilations and seems to suggest that the difficulty of the original problems impacts the planning time more significantly than the type of compilation.

The interquartile range of the data sets is generally small, showing there to be little variation in the planning time for each compilation type. 
The IQRs of the data sets are also grouped around the horizontal axis showing that there is not a large increase or decrease in planning time for the majority of the compilations across the problems. 
A compilation for a question type FQ7 in Figure~\ref{fig:sfig4} gives the greatest increase in planning time of 33.01 seconds. However, this is an extreme value for this data set as can be seen from the IQR of 2~--~3.08. There are a few other significant changes on planning time from compilations. For example, for a question type FQ1 in Figure~\ref{fig:sfig1} there is an increase of 0.08 seconds. This is quite substantial considering that the original planning time was essentially 0 seconds. However, in practice the increase in planning time is negligible. For each of these significant changes in planning time, the IQR of the data set shows that it is an extreme value.

The largest IQR is for FQ4 in Figure~\ref{fig:sfig4} of 0.055~--~2.145. This is expected, because problem 20 is the hardest to solve for this domain. The other ranges in this problem are similar and also show little negative impact in the planning time.
The data set with the largest interquartile range compared to the other compilations performed in the problem is FQ6, which corresponds to the compilation shown in Section~\ref{comp:tw2}, in Figure~\ref{fig:sfig2} with an IQR of 0.045~--~0.455. This stands out compared to the other results in the plot where the ranges are very small, and the values show close to zero, however, in practice an increase in planning time of 0.045 to 0.455 seconds is still negligible. 

the results for problem 20, shown in Figure~\ref{fig:sfig4} show that for some compilations there was an improvement in planning time. For FQ4 this seems to be an extreme case where only the lowest planning time was an improvement of 17.09 seconds. Whereas, for compilations of the question type FQ3, the majority improved the planning time. In fact, across all problems the compilations for FQ3 had the least negative impact on planning time.


Figures~\ref{fig:performance_graphs_easy_metric_scatter} and~\ref{fig:performance_graphs_hard_metric_scatter} show the impact of the compilations on the solution quality for the easy and hard problems we defined earlier. In each of the three domains used in our experiments, the metric for quality is defined to be the total duration of the plan, keeping in mind that actions can be performed in parallel. The horizontal axis is the same as in Figures~\ref{fig:performance_graphs_easy_scatter} and~\ref{fig:performance_graphs_hard_scatter} whereas the vertical axis measures the difference between the metric for the plan for a compiled HModel and the metric for the original plan for the original model. Zero on the vertical axis represents no difference in metric for the plans from the original and compiled models, a positive value indicates the metric for a compiled model was higher, and vice versa for a negative value.

As opposed to the results comparing the impact of the compilations on planning time, these results do contain the most optimised plan. This is because each problem had the same amount of time within which to find a solution, including the original problem. Although the ultimate planning time for two problems may have differed, they both had the same opportunity to improve. Therefore, we consider the quality of two plans found in the same allotted time comparable. 

Nonetheless, as we observed earlier, the constraint added in response to a question could increase the metric for the solution significantly, but still be optimal under the new constraint. However, another constraint added to the same problem could marginally increase the metric, but be sub-optimal. From the spread of the values in the results, potentially inefficient solutions are recognisable. Data points that lie in the same metric range are likely caused by constraints that limit the search space of the problem, causing better quality solutions to be pruned away. Whereas lone data points such as for FQ1 in Figure~\ref{fig:easy_metric_sfig3} may indicate inefficient solutions.

The results show that the majority of the compilations do not impact the metric significantly. Six of the seven compilation types have a median average increase in metric of less than 8.5, whilst the seventh has an average increase of 21.502. A compilation of the question type FQ3 for problem 20 of the Crew Planning domain had the largest impact on the metric, with an increase of 2541, shown in Figure~\ref{fig:hard_metric_sfig4}. A compilation of the question type FQ3 also lead to the biggest decrease in metric, an improvement of 84.83 to the original solution of problem 10 of the Depots domain, shown in~\ref{fig:hard_metric_sfig2}. The compilation for FQ3 on average lead to the largest increase in the metric, whereas the compilation for FQ2 had the lowest.

The compilations applied to the easier problems had no improvements on the metric. This could be due to the solutions to the original problems being optimal. The compilations when applied to the easier problems had a worse effect on the metric than the harder problems, with a median increase of 3.681 compared to 2.002.

\begin{table}[t]
\centering
\begin{tabular}{| *{13}{c|} }
    \hline
\multirow{2}{*}{}
    & \multicolumn{4}{c|}{Crew Planning}
            & \multicolumn{4}{c|}{Depots}
                    & \multicolumn{4}{c|}{Elevators}               \\
   \cline{2-13}
      &   1  &   2  &   5  &   20  &   1  &   3  &   10  &   13 &  1  & 2  & 3 & 5 \\
    \hline
FQ1   &  0 & 0 & 0 & 0 & 0 & 1 & 4 & 0 & 0 & 0 & 0 & 0 \\
    \hline
FQ2   &  10 & 10 & 3 & 0 & 3 & 2 & 2 & 0 & 0 & 0 & 0 & 0 \\
    \hline
FQ3   &  0 & 0 & 0 & 0 & 0 & 0 & 0 & 0 & 0 & 0 & 0 & 0    \\
    \hline
FQ4   &   0 & 1 & 0 & 0 & 0 & 4 & 8 & 2 & 0 & 2 & 1 & 0  \\
    \hline
FQ5   &   2 & 2 & 1 & 0 & 2 & 0 & 2 & 0 & 0 & 0 & 0 & 0  \\
    \hline
FQ6   &     0 & 0 & 0 & 1 & 0 & 2 & 8 & 0 & 0 & 2 & 0 & 0   \\
    \hline
FQ7   &    1 & 0 & 0 & 0 & 0 & 0 & 5 & 0 & 0 & 4 & 0 & 1    \\
    \hline
\end{tabular}
\caption{Table showing the number of problems for which a solution could not be found grouped by question type. The table is divided into the domain type and then sub-divided by the problem number.}
\label{tab:compilation_fails}
\end{table}

Although any constrained problems that were provably unsolvable were discounted and repeated, we did not repeat the tests for problems where the planner failed to find a solution. However, because there was no data for these problems the results were not displayed in the graphs. Table~\ref{tab:compilation_fails} displays the number of problems for which a solution could not be found within the allotted planning time. Overall, 86 of the 840 constrained problems failed to be solved within the required time. 29 of the 86 were derived from compilations applied to problem 10 for the Depots domain. This problem took the longest time to solve at 245.06 seconds, this being closer to the maximum planning time could be the reason for the failures in finding solutions. However as stated above there were many compilations that improved the planning time for this problem, so it was possible to solve (efficiently) with the correct constraints. 30 of the 120 HModels produced to answer questions of the type FQ2 were not solvable within the allotted time. This question is unlike the others as the constraint it produces enforces that an action cannot be present in a solution, rather than forcing it to. Therefore, the number of these questions that can be asked about a plan is limited by the number of actions that appear in the plan. The limited choice on questions may have impacted the ability to find a solution, for example the plan for problem 1 for the Crew Planning domain had only 12 actions. If any of the actions were landmarks or crucial for achieving a goal in the plan, then removing these actions would have a large consequence the resultant solution, potentially even making the problem unsolvable. Although none of the constrained problems were found to be provably unsolvable, removing an action from the larger problems with larger plans had less of an impact. 
The process for proving if a problem is unsolvable is difficult, therefore it is infeasible to definitively determine if these problems are provably unsolvable or not.

\paragraph{Analysis}
Many of the compilations when applied to the harder problems lead to better solution plans than the originals. This is an important feature of the plan negotiation problem, where a user can suggest a counterfactual that leads to a better plan. This observation is very important: it demonstrates that the original plan is not optimal and that the addition of constraints in these cases actually narrows the search space in a way that reduces the work required to find a good quality plan in the remaining space. It is perhaps surprising that automatically generated questions that do not target observed weaknesses in an original solution should so often lead to improved plans.
These results highlight the difficulty in finding optimal plans, the necessity to be able question unconvincing plans, and the effectiveness of our compilation approach in finding suspected improvements in plans.

The spread of results on metrics is noticeably larger than for the results on planning time. We believe that this is because the performance of the planner in finding a first solution to a problem is most significantly affected by the domain and the size of the problem being solved, neither of which is significantly altered by the compilation of constraints. In contrast, the quality of the best plan that can be found in a given time can be very much affected by the constraints in the problem: high quality solutions can be excluded by the addition of constraints, and poor quality solution branches can be pruned by the addition of constraints. This was discussed in further detail in Section~\ref{sec:comp}, we see evidence for both patterns of behaviour in these results, which substantiates the discussion.

The compilations applied to problem 10 for the Depots domain, the results of which are shown in Figure~\ref{fig:hard_sfig2}, produced unusual results. The planning times are considerably more varied than the results found for any other domain-problem pairs. The compilations also improved the planning time more than any other domain-problem pair. A possible reason for this is due to the planner having difficulties finding a solution to the original problem because of failing to select the choice branch leading to a simpler solution. For example in the search space there could be a more complex path which which the planner is biased to go towards through a misleading heuristic. The constrained HModels produced for this problem might not have this issue. The heuristic for the new model could more accurately lead the search to a goal, or the complex parts of the search tree could be pruned away in the new model entirely. 

The compilation for the formal question FQ3, shown in Section~\ref{comp:replace}, had the lowest impact on the planning time out of any of the question types. This is likely due to the nature of the question requiring a part of the plan be fixed. Therefore unlike the other compilations, the compiled problem is smaller than the original. Although this does not guarantee that the problem will always be easier to solve, as the specifics of the replacement performed by the compilation could have a substantial effect on the difficulty of the constrained problem.

\subsubsection{Performance of Iterated Compilations}\label{sec:eval_iteration}

We now present experiments exploring the effects of iterating multiple compilations of constraints. The iterated model restrictions that underpin the interaction we describe in Section~\ref{subsec:problem} depends on the planner meeting the demands of planning for models in which multiple constraints have been compiled (using the approach described in Section~\ref{sec:comp}). As we have already observed, the addition of constraints to a model can, in general, be expected to make the problem harder to solve. A well-known phenomenon affecting combinatorial optimisation problems is the {\em phase transition}~\cite{phasetrans}: members of a family of combinatorial problem include instances that are very easy to solve and other instances that are so over-constrained that it is trivial to determine that they are unsolvable. As constraints are added to the former, or removed from the latter, instances are created that are progressively more difficult to solve or more difficult to show unsolvable, respectively. Between these two advancing problem sets lies a transition from solvable to unsolvable and the problems at this boundary are typically the most difficult to tackle (which ever way the resolution lies). Thus, as we iteratively add constraints to a problem, we are pushing towards the phase transition where the problems are likely to become harder to solve. In these experiments we seek to determine the extent to which that expectation affects the performance in practice.


\paragraph{Purpose}
The second experiment was designed to evaluate the competence of the compilations when used within the iterative approach to explanations. For a user to engage in a conversational process with the explanation system they must receive efficient responses to their questions. As shown in the first experiment above, each type of compilation generally scales well with the complexity of the original problem. However, the results of the first experiment do not give any insight into how the compilations interact or interfere with one another and whether it is reasonable to expect a planner to produce solutions for more precisely constrained models. Therefore, a second experiment was designed to evaluate the impact of iterative compilations on the planning time and the quality of solutions.

\paragraph{Design}
In this experiment we used the same domains as the first experiment but instead of the Depots domain the ZenoTravel domain was selected from IPC-3. This is because there was not enough feasibly solvable problems provided by the benchmarks for the Depots domain for the breadth of this experiment. We chose the ZenoTravel domain because it belongs to the same set of benchmarks as the Depots domain and there are clear justifications for the need of explanations in the domain. For each domain we selected ten problems of varying complexity provided by the IPC benchmarks. We selected problems with the same range of complexities across each of the domains. For the Crew Planning domain this was problems 1 to 10, for Elevators 1 to 9 and 14, and problems 3 to 12 were selected for the ZenoTravel domain. We did not select problems 1 or 2 in the ZenoTravel domain because they were too easy to solve, and we did not select problems 10 to 13 in the Elevators domain because they could not be solved within the designated time.

We first solved each of these domain-problem pairs to get the original plans that are used as the control and to generate the first set of questions. We then selected a question type from the Contrastive Taxonomy at random, and generated an appropriate question. We then compiled this question into the original model to generate the HModel and used a planner to find the solution HPlan. We repeated this step for a total of twelve times, but each time generated a question from the last HPlan and compiled the question into the last HModel. The results from the user study in Section~\ref{sec:userstudy} suggest that users only ask five questions on average, however for the sake of robustness we simulated twelve for each problem. We generated questions using the same approach as the first experiment and disregarded questions that lead to provably unsolvable models.

All tests used a Core i7 1.9GHZ machine, limited to five minutes and 16GB of memory. We increased the planning time from the first experiment by two minutes to offer a larger window through which to view any growth trends in the planning time for models with iterated constraints. We used the POPF~\cite{popf} planner and recorded all solutions found in the time allocated to test the effect compilations have on solution quality with optimisation.

The compilation for the formal question FQ3, was not used in this experiment. This is because, by the nature of the question, the part of the plan up until the action that is being replaced is fixed. Unlike any other compilation, for all subsequent questions, the compilation is applied to an HModel that has a partially solved problem expressed as its initial state.
Therefore, the FQ3 compilation is the only one that reduces the size of the problem to be solved, having a distorting effect on the impact of other compilations.

\begin{table}[t]
\centering
\begin{tabular}{lrrrr} 
 \toprule
 Problem  &  &  & \\
 Number &     Crew Planning  &    Elevators  &   ZenoTravel    \\
 \midrule
 1, 1, 3    &   12           &  12           & 7\\
 2, 2, 4    &   8            &  12           & 12\\
 3, 3, 5    &   9            &  12           & 12\\
 4, 4, 6    &   12           &  4            & 12\\
 5, 5, 7    &   12           &  12           & 12\\
 6, 6, 8    &   12           &  9            & 12\\
 7, 7, 9    &   12           &  12           & 10\\
 8, 8, 10   &   12           &  12           & 9\\
 9, 9, 11   &   12           &  12           & 6\\
 10, 14, 12 &   12           &  2            & 4\\
 \bottomrule
\end{tabular}
\caption{Table showing the largest number of cumulative compilations applied to each domain-problem pair that was still solvable within the five minutes of allocated planning time. The Problem Number column denotes the problem for each domain in order, for example the bottom row shows the results from problem 10 for the Crew Planning domain, problem 14 for the Elevators domain, and problem 12 for the ZenoTravel domain.}
\label{tab:iterations_success}
\end{table}

\begin{figure}
\begin{subfigure}{.5\textwidth}
  \centering
  \includegraphics[width=1.0\linewidth]{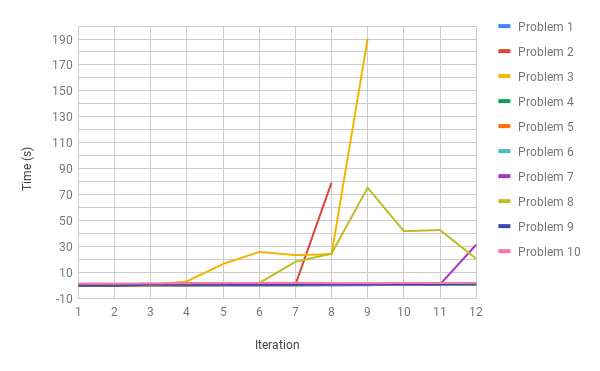}
  \caption{Crew Planning Domain Problems 1 - 10}
  \label{fig:second_test_time_sfig1}
\end{subfigure}%
\begin{subfigure}{.5\textwidth}
  \centering
  \includegraphics[width=1.0\linewidth]{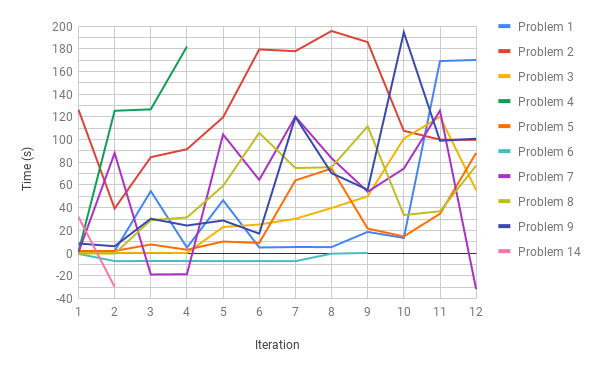}
  \caption{Elevators Domain Problems 1 - 9, 14}
  \label{fig:second_test_time_sfig2}
\end{subfigure}
\begin{subfigure}{.5\textwidth}
  \centering
  \includegraphics[width=1.0\linewidth]{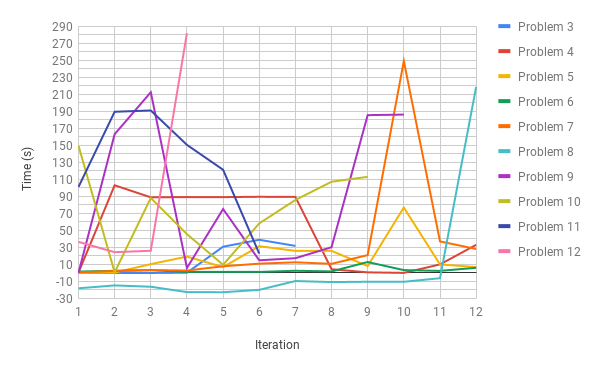}
  \caption{ZenoTravel Domain Problems 3 - 9, 10 - 12}
  \label{fig:second_test_time_sfig3}
\end{subfigure}
\caption{Line chart displaying the change in planning time over multiple iterations of compilations applied to 10 problems for 3 planning domains.}
\label{fig:performance_graphs_second_test_time}
\end{figure}

\begin{figure}
\begin{subfigure}{.5\textwidth}
  \centering
  \includegraphics[width=1.0\linewidth]{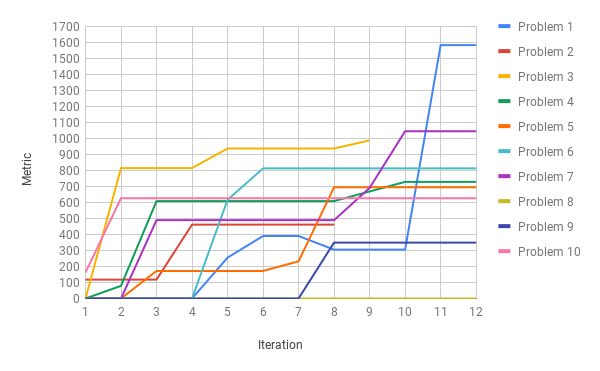}
  \caption{Crew Planning Domain Problems 1 - 10}
  \label{fig:second_test_metric_sfig1}
\end{subfigure}%
\begin{subfigure}{.5\textwidth}
  \centering
  \includegraphics[width=1.0\linewidth]{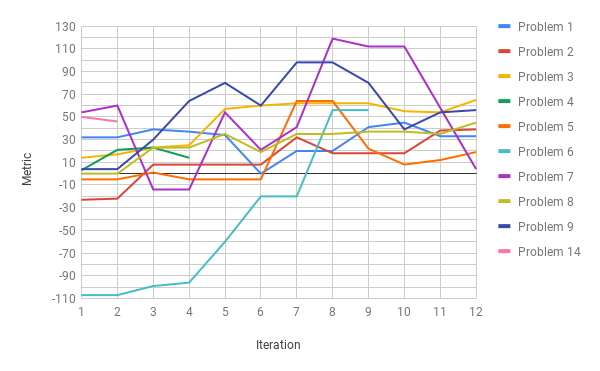}
  \caption{Elevators Domain Problems 1 - 9, 14}
  \label{fig:second_test_metric_sfig2}
\end{subfigure}
\begin{subfigure}{.5\textwidth}
  \centering
  \includegraphics[width=1.0\linewidth]{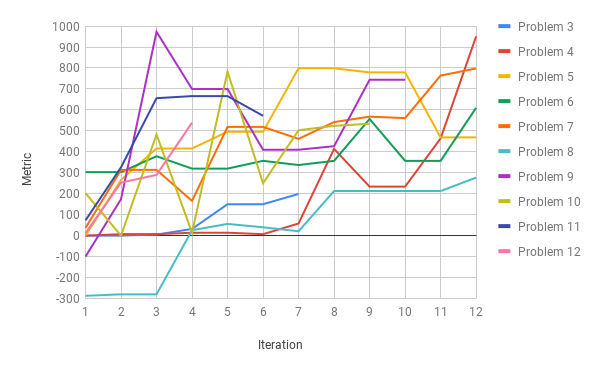}
  \caption{ZenoTravel Domain Problems 3 - 9, 10 - 12}
  \label{fig:second_test_metric_sfig3}
\end{subfigure}
\caption{Line chart displaying the change in planning quality (metric) over multiple iterations of compilations applied to 10 problems for 3 planning domains.}
\label{fig:performance_graphs_second_test_metric}
\end{figure}

\paragraph{Results}
    
The results of this experiment are shown in Table~\ref{tab:iterations_success} and Figures~\ref{fig:performance_graphs_second_test_time} and~\ref{fig:performance_graphs_second_test_metric}. 

Table~\ref{tab:iterations_success} shows the number of iterations of compilations successfully applied to a problem. For example, the HModel formed from the constraints derived from 8 questions compiled into problem 2 for the Crew Planning domain was able to be solved within 5 minutes, the 9th question and compilation produced an HModel where no plan could be found within 5 minutes. The majority of the problems were still solvable after the full 12 iterations of compilations were applied, with 20 of the 30 producing plans for the most constrained problems. Of the problems that failed to find plans before the full number of iterations, only three of these were below the average number of questions users asked about plans, as found in our user study. 

Figure~\ref{fig:performance_graphs_second_test_time} displays the change in planning time as the iterations of compilations applied to problems increases. This Figure contains three line charts, one for each domain used in the experiment. Each line corresponds to a planning problem, the key is shown on the right of each chart. The horizontal axis displays the iteration of the compilation applied, for each subsequent iteration, the compilation is applied to the previous HModel so each iteration is more constrained than the last. The vertical axis displays the difference in planning time in seconds. For each of the plots the difference is relative to the planning time for the original problem, so the zero on the vertical axis represents there being no difference in planning time between the compiled model (at any iteration) and it's corresponding original problem, a positive value means there was an increase in the time taken to find a solution for the compiled model, and the opposite holds for a negative value.

The results displayed in Figure~\ref{fig:performance_graphs_second_test_time} show various trends on the impact iterative compilations have on the planning time. There are problems where the impact is negligible from start to finish, such as in problems 1, 4, 5, 6, 9, and 10 for the Crew Planning domain, and problems 5 and 6 for the ZenoTravel Domain. There are problems where the planning time drastically increases for a single iteration, this can lead to no solution being found within the allocated time for the HModel produced from the next iteration, such as in problem 3 in the Crew Planning domain, and problem 12 for the ZenoTravel domain. 
However, sometimes the planning time decreases again in subsequent iterations, for example in problem 7 in the ZenoTravel domain. This also happens multiple times in problems 7, 8, and 9 in the Elevators domain where the planning times fluctuate between drastic increases and decreases in planning time from one iteration to the next. In other cases the planning time can increase, stay level for some iterations, and then decrease again such as in problem 2 in the Elevators domain and problem 4 in the ZenoTravel domain.
In some cases the compilations can decrease the planning time such as in problems 6 and 8 in the Elevators and ZenoTravel domains respectively which both have improvements in the planning time over several iterations. 
Although, for most of these plots there tends to be no correlations or a slightly positive correlation between the number of iterations and the planning time. This is quite easy to see in Figures~\ref{fig:second_test_time_sfig1} and~\ref{fig:second_test_time_sfig3} where the increase in planning times lies below 30 seconds for the final iteration for the majority of problems.
The Elevators domain has a more positive correlation between the number of iterations and the planning time than the other domains, where the majority of problems finish with an increase in planning time of between 50 and 100 seconds. Although it does also have an example of a significant improvement in planning time for even the most constrained HModel in problem 7.

Figure~\ref{fig:performance_graphs_second_test_metric} displays the change in the quality of the solutions as the iterations of compilations applied to the problem increases. This figure contains three line charts for each domain in the experiment. The charts are set up the same as those in Figure~\ref{fig:performance_graphs_second_test_time}, but the vertical axis displays the difference in the quality of the solutions as measured by the metric defined in the domains. 

The results displayed in Figure~\ref{fig:performance_graphs_second_test_metric} show that the quality of the solutions vary over the number of compilations applied to a problem. The quality of the solutions tends to get worse over the full number of compilations applied in the Crew Planning and ZenoTravel domains. At a more granular level, the metric increases and decreases from one iteration to the next quite drastically for some problems, for example in problems 9 and 10 in the ZenoTravel domain. However, some others have a more steady increase, such as in problem 7 in the ZenoTravel domain. For all problems, but problem 8, in the Crew Planning domain there is one compilation that causes a drastic increase in the metric, the subsequent compilations then have much less of an impact. Problem 8 in the Crew Planning domain is only problem where each iterative compilation had no impact on the metric. The compilations have a much lower impact on the metric for the problems in the Elevators domain. Again, there are small fluctuations in the metric from one iteration to the next, but the overall change in the metric for the majority of the problems is minimal. This can be observed by noticing that the majority of the problems end with a insignificant increase in metric of between 4 and 65.

\paragraph{Analysis}
The results show that the majority of the constrained problems produced from a large number of iterations of compilations are still solvable and within a reasonable time. There were a few instances where a problem was constrained in such a way that it became unsolvable within the 5 minutes of allotted planning time. The results indicate that this is not because of incremental increases in the difficulty over the number of compilations applied to a model, but because of a single compilation that makes the subsequent HModel significantly more difficult to solve. This is likely due to the same reasons that have been discussed in the introduction to Section~\ref{sec:comp}.

The results show that problems do not get significantly harder to solve as the number of constraints applied to the problem increases. This can be shown by running a Wilcoxon Signed-Rank test on the differences between the planning times for consecutive iterations of HModels. For each problem two populations were created from the data, $\mu_1$ and $\mu_2$ where $\mu_1 = \{0, p_1, \dots, p_{n-1}\}$ and $\mu_2 = \{p_1, ..., p_n\}$ and $p_i$ is the planning time for a model at iteration $i$ and $n$ is the number of iterations. 
We performed a Wilcoxon Signed-Rank test with the null hypothesis $H_0: \mu_1 = \mu2$ and alternate hypothesis $H_1: \mu_1 \neq \mu_2$.
The results were not significant for 26 of the 30 problems with $p = 0.05$. This shows that for these 26 problems the compilations did not significantly worsen the planning time as the iterations progressed. Four of the problems were statistically different and so we could reject $H_0$ and accept $H_1$ with $p = 0.05$. More specifically a one tailed test showed that for these four problems we could accept the alternate hypothesis $H_2: \mu_1 < \mu_2$. Even though for these four problems the compilations worsened the planning time as the iterations progressed, this does not mean that they got exponentially harder. 


The results on the impact in the metric, as discussed in the results, mostly show that the greatest increase in the metric comes from a single compilation. This is likely due to the constraint compiled into the model constraining the search space in such a way that the better solution plans are no longer valid, or are much harder to find. This also makes sense when noticing that subsequent compilations rarely improve the metric once it has been drastically increased. Any constraint compiled into an HModel produces a new HModel that also has the constraints that were applied previously. If one of these constraints limits the space of valid plans by removing a plan with a better quality, then any subsequent constrained HModels will also be limited in this way. This is also supported by, more severely, the failure to find solutions for some sufficiently constrained problems within the time allotted. Unrelated to the number of constraints that have been applied, a problem can go from solvable to unsolvable with a single additional constraint. This leads us to believe that the number of constraints is of less importance than the ultimate constrained problem, whether this be the result of one compilation or of multiple.

\subsection{User Study} \label{sec:userstudy}


We designed a comparative user study to evaluate the effectiveness of our XAIP-as-a-Service framework and the iterative model restriction approach that it utilises. We designed the study based on recommendations for metrics on Explanation Satisfaction by Hoffman et al.~\shortcite{hoff18b}. Explanation Satisfaction is a measure of the degree to which users feel that they understand the AI system or process being explained to them and is a contextualised, a posteriori judgment of explanations. The rest of this section describes the design of our user study and discusses the results it produced.

\begin{table}[h!]
    \begin{tabular}{|p{0.8cm}p{12cm}|}
    \hline
      \textbf{EQ1:} & ``How well did the XAIP system help you to understand the plan?'' \\
        \hline
\textbf{EQ2:} & ``What do you expect from a good explanation?'' \\
\hline
\textbf{EQ3:} &  ``What would make an explanation satisfactory?'' \\
\hline
\textbf{EQ4:} &``In what ways does it help if explanations are contrastive?'' \\
\hline
    \end{tabular}
    \caption{Open-ended questions for users}
    \label{tab:user_questions}
\end{table}

\begin{figure}[!b]
    \centering
    \includegraphics[width=0.43\columnwidth]{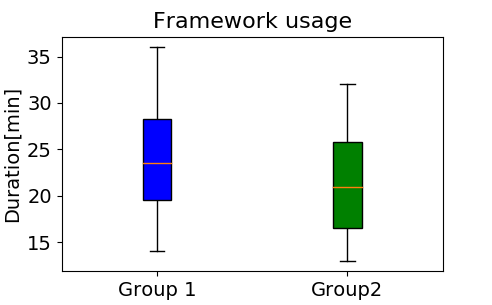} 
    \includegraphics[width=0.56\columnwidth]{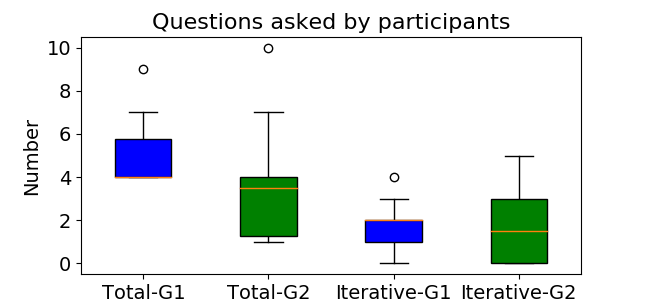}
    \caption{Standard deviation of the time participants spent understanding the plan and using the framework (left). Standard deviations of total number of questions that participants asked and of number of questions that were asked  as iterative questions (right).}
    \label{fig:stats}
\end{figure}

\begin{figure}[!b]
    \centering
    \includegraphics[width=1.0\columnwidth]{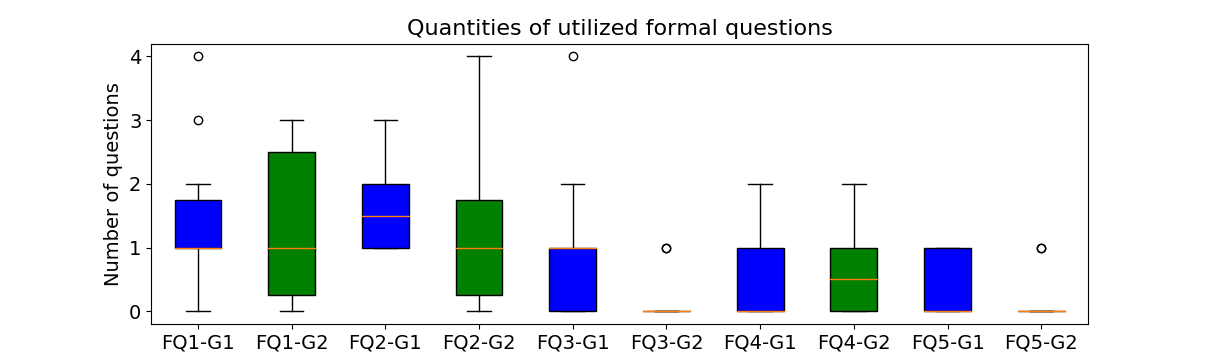} 
    \caption{Standard deviation of the number of formal questions (defined in Section~\ref{sec:taxonomy}) utilised by study participants. G1 label and blue colour denote results for Group 1, while green and G2 for Group 2.}
    \label{fig:statsFQ}
\end{figure}

\begin{figure*}[t!]
    \centering
    \includegraphics[width=0.85\textwidth]{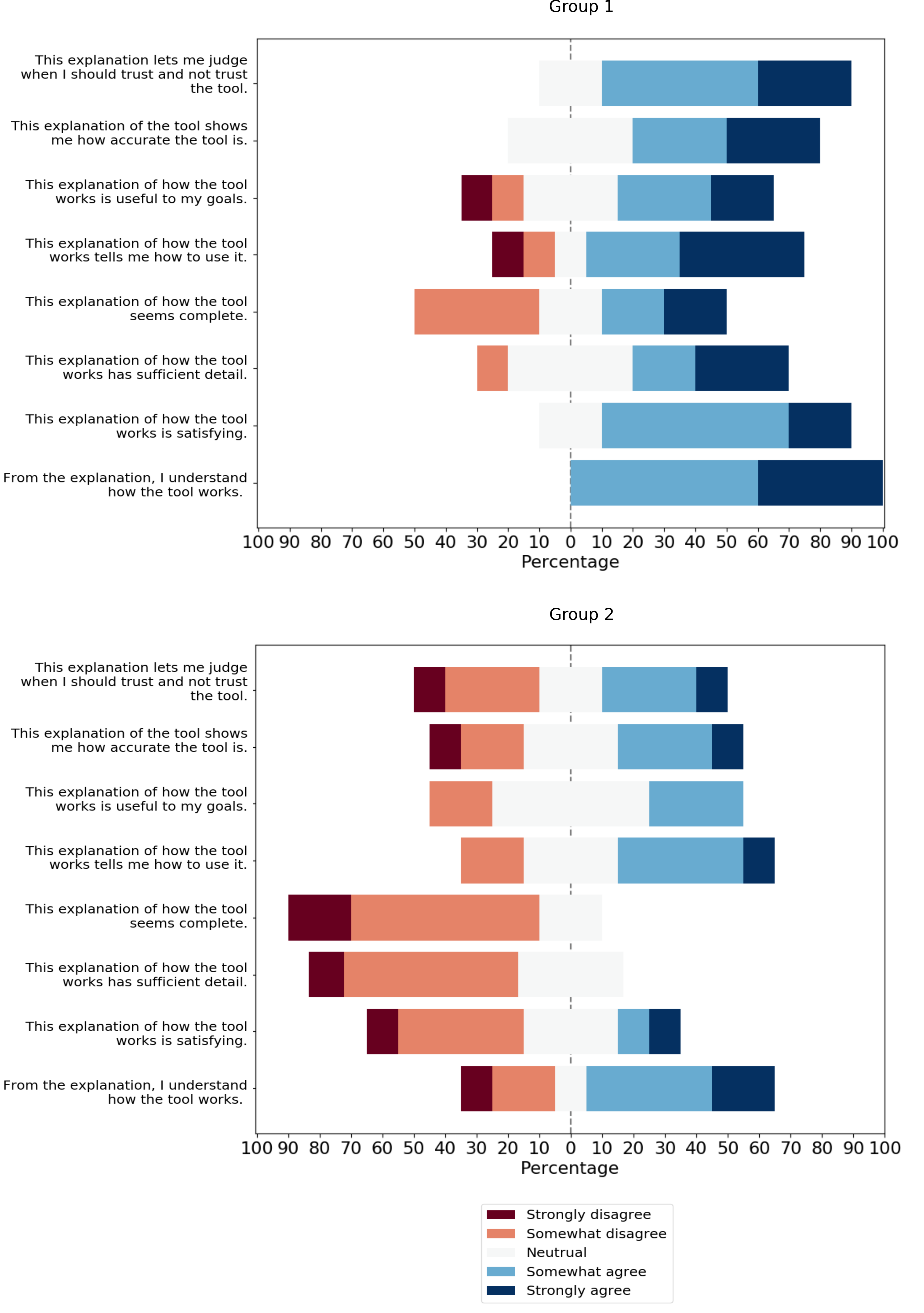}
    \caption{Explanation Satisfaction Scale results collected in the user study. Likert bar plot visualises distributions of users answers on their attitudes towards statements about explanations on the left side of the figure.}
    \label{fig:sat_scale}
\end{figure*}

\paragraph{Design}

A comparative study, also called a true experiment, is a method of data collection designed to test hypotheses under controlled conditions in behavioural research. True experimental designs involve 
manipulation of the independent variable by exposing participants to different conditions by varying this variable. In the experiment that we designed, participants were randomly divided into two groups - the experimental group who interacted with the original framework, and the control group who interacted with a simplified framework.

Participants were using the XAIP as a Service framework in order to understand the plan in Figure~\ref{fig:plan}. They were asked to imagine themselves as employees of the warehouse that we use as our running example described in Section~\ref{sec:running_example}. The chosen domain was simple enough so participants could identify themselves in the roles, and complex enough to judge the XAIP system. 

20 participants were divided in two groups in order to compare user experience. The experimental group \textbf{Group~1}~(G1) had the opportunity to use the framework's full capabilities and get explanations in the form of highlighted comparisons between plans, as disc in Section~\ref{sec:xaip_human_interface} and illustrated in Figure~\ref{fig:guib}. The control group \textbf{Group~2}~(G2) used the framework where the $comparison$ of plans (with coloured differences and highlighted costs) was disabled. As the explanation, they only got the HPlan next to the original plan. Both groups were asked to use the framework until they were satisfied with their understanding of the system and the plan or until they found interaction with the framework useful. We did not attempt to track how the model of the users evolved whilst using the system, or what triggered their termination with the system, but to determine the overall experience the users had in using they framework.

To evaluate understanding of plans that participants gained using the framework, they rated the system using the Explanation Satisfaction Scale, which is a 5-point Likert scale for Explainable AI systems designed by Hoffman, Klein, and Mueller~\citeyear{hoff18a}. 
The scale measures an attitude on a continuum from `strongly agree' to `strongly disagree' corresponding linearly to values from 1.0 to 5.0. Additionally, we collected qualitative information about the experiences and expectations Group 1 had whilst partaking in the study by asking them four open-ended questions as defined in Table~\ref{tab:user_questions}.

\paragraph{Data} We conducted a study with 20 volunteers (5 students, 4 engineers, 4 software developers, 3 researchers, 2 assistant professors, a chemist and a copywriter) divided in two groups with 10 persons each. Participants’ ages ranged from 23 to 43 years, with 35\% identified as female and 65\% identified as male. The average time G1 spent with the plan and the framework is 24.2 minutes, and on average, they asked 5.1 questions. The average time G2 spent with the plan and the framework is 21.1 minutes, and on average, they asked 3.7 questions as it can be seen  in Figure~\ref{fig:stats}. The maximum number of question asked was 10, while the minimum number was 1. The most commonly utilised formal questions were types FQ1 and FQ2 which require removing actions from the plan or adding new actions to the plan. Quantity distributions of each formal question asked are given in Figure~\ref{fig:statsFQ}.

\paragraph{Quantitative results} The results of the evaluation with the Explanation Satisfaction Scale are shown in Figure~\ref{fig:sat_scale}. The median of all responses for Group 1 is 4.0, and the interquartile range (IQR) is 2.0, which corresponds to the overall attitude ``somewhat agree'' that information gained through usage of  the framework is satisfying. Some of the participants did not find the plan comparisons useful to their goals, and some of them somewhat disagreed that the explanations as presented are complete. The median of all responses for Group 2 is 3.0, and IQR is 2.0, which corresponds to the overall attitude ``neutral'' that explanations obtained through interaction with the framework are satisfying. We also performed t-test with the results of both groups and got the values $t=5.57$ and $p=1.038\mathrm{e}{-7}
$ telling us that there is a significant difference in explanation satisfaction of two groups. Generally, the users found plan comparisons as in Figure~\ref{fig:guib} useful for understanding how the AI planning system works and that the process of iterative model restriction is satisfying. Also, they agreed that explanations in this way could be useful for improving judgement about whether or not to trust the system. 

\paragraph{Qualitative results}6 out of 10 participants of Group 1 filled out all answers for the questions in Table~\ref{tab:user_questions} and, overall, there was an 82.5\% response rate to the questions. In the text below, we denote the study participants as P1-P10. 
By analysing the collected data, we were able to determine several patterns and themes amongst users' responses. 

Participants agreed that the framework is helpful and that they had ``no problem understanding the plan thanks to the system''(P3). Also, they found the presentation of plan comparisons useful ``to understand why ... something (is) possible or not''(P2),  to see ``how the changes ... affect the original plan''(P3) and to aid in ``allowing to reach some conclusions''(P10). 

However, for a good explanation, users expect more details ``explaining the rules for the plan''(P6) and  ``showing the logic behind the reason that a specific action has been decided''(P7).
Also, they ``expect that it (good explanation) would explain each step of the plan in an understandable way, to explain every change that has been made and how does that change affect the plan''(P4) and  they ``expect an argument for the usage of one thing over anything else''(P9). This theme also complements their attitude towards explanation completeness as revealed in the Explanation Satisfaction Scale results (Figure~\ref{fig:sat_scale}).

Additionally, participants reflected on the explanation presentation. Even though they think ``it (is) great to compare plans and see what makes the difference''(P5), they also expect more user-friendly presentation of final explanation such as ``in the format of a sentence''(P8) or ``a visual representation of what the robots are doing''(P9).

\section{Related Work}
\label{sec:rel}

As discussed earlier, 
several researchers (e.g: Mueller et al.~\citeyear{mue19}) have drawn a distinction between local and global questions and the corresponding explanations. 
The study in Section~\ref{sec:taxonomy} showed that local questions are significant for XAIP, and that many of these questions are contrastive in nature.
In this section we present an overview of additional related research on explanation with a focus on local questions and explanations.  We start with  a brief description of relevant background ideas from Philosophy, Psychology and Cognitive Science, which have contributed to current views on explanation.  We then focus in on more recent related work on local explanation from XAI, and finally on work in XAIP, which is most closely related to the work described here.


\subsection{Philosophy, Psychology and Cognitive Science}

There is a vast body of literature in the fields of philosophy, psychology, and cognitive science on the topic of explanation.  
Much of the early work in this area focused on \textit{causal explanation}, 
namely the idea that questions could be answered by identifying and elucidating the causes for a particular event or result being questioned. Hume~\citeyear{hum48} noted that if there is a cause between two events, the first is always followed by the second. 
Lewis~\citeyear{lew74} expanded on this, arguing that we should understand Hume's definition wherein if an event \textit{C} causes the event \textit{E} and, if under some hypothetical case, \textit{C} did not happen then neither would \textit{E}.
Lewis~\citeyear{Lewis1986-LEWCE} then argued that to explain an event one must provide some information about its causal history. Although, this may be enough to explain why an event happened, it does not answer all questions about an event. One could not use the causal history of an event alone to answer if it was the best outcome, what would happen if the event did not occur, or if another event occurred in it's place.
As a consequence, more recent work on explanation has introduced and considered notions of contrastive questions and explanations. 
In particular, philosophers, such as van Fraassen~\citeyear{van80}, have argued that ``why"-questions can be implicitly or explicitly understood as: ``why is A better than some alternatives?". This is what we describe as a local contrastive question, where the questioner wants to understand why a plan is good, or why a particular decision was made. These questions can be answered with contrastive explanations.

Van Bouwel and Weber~\citeyear{bou02} defined four types of explanatory questions, three of the four are explicit local contrastive why questions that call for explaining the differences between either real or hypothetical alternatives. They argue that the final explanatory question is not contrastive, and is asked when the user wants a global understanding of the properties of objects.
However, most other researchers think all ``why" questions ask for contrastive explanations, whether the contrast case is implicitly or explicitly stated~\cite{hil90,lipton_1990,lom12}.
Hilton~\citeyear{hil90} recognised that one does not explain an event, but instead explains why the event occurred in one case but not in a counterfactual (hypothetical) contrast case. This is the basis of contrastive explanations.
Lipton~\citeyear{lipton_1990} argues that the cognitive burden of complete explanations is too great. He goes further by demonstrating that explaining a contrastive question is less demanding because it is enough to show what is different between the two cases instead of a full causal analysis. 
Miller~\citeyear{mil18} extended structural causal models~\cite{hal05c,hal05e} to contrastive explanations based on Lipton's Difference Condition. Lipton's difference condition states that to explain why \textit{P} rather than \textit{Q}, one must cite a causal difference between \textit{P} and not-\textit{Q}, consisting of a cause of \textit{P} and the absence of a corresponding event in the history of not-\textit{Q}~\cite{lipton_1990}.
Hilton~\citeyear{hil90} argues that explanation is a conversation. We have adopted this strategy by treating explanation as an iterative process where the human continues to ask contrastive questions as a means of understanding plans, and imposing additional constraints on the planning problem.

\subsection{Explainable AI}
Meuller et al.~\citeyear{mue19} provide an overview of the landscape of research into Explainable AI (XAI). This work spans several decades, and includes work carried out with intelligent tutoring systems, XAI hypotheses and models, and explanation in expert systems. The early work on explanation in expert systems provided causal explanation for conclusions, often in the form of chains of rules contributing to the conclusion~\cite{van78}. 
Recently, there has been a resurgence of interest in explanation in XAI, both when the model is and is not interpretable.  This is, in large part, due to the difficulty of understanding the results of deep learning systems~\cite{rud19}.
Madumal et al.~\citeyear{mad19} use structural causal models to answer \textit{``why?"} and \textit{``why not?"} questions in Reinforcement Learning systems. They learn approximate causal models for counterfactuals to explain the local contrastive question \textit{``why not action B (rather than A)?"}, by simulating what would happen if B was performed, and providing a contrastive explanation showing the difference between the causal chain of \textit{A} and \textit{B}. They provide minimally complete explanations to answer questions of the form \textit{''why action A?"}. However, the explanation is more of a justification that the action \textit{A} contributes to the goal in some way based on the global model, rather than explaining why the decision was made to perform the action as opposed to some other action or not at all. We argue that users need explanations to a wider set of questions to understand the reasoning behind a particular decision.

As a user may not have any knowledge of the inner workings of a black box system, researchers have focused on providing global explanations for how a system came to a particular outcome~\cite{kri99,joh09,aug12}.
However, researchers have recently tried to tackle the problem of providing local explanations for why a black box system arrived at an outcome. For example an image classifier might recognise an image as a bird; Ribeiro et al.~\citeyear{rib16} provide a way to highlight features of the image to help justify its decision, for example highlighting the birds beak and wings. However, Hoffman et al.~\citeyear{hof18c} have argued that neither of these approaches is enough and that users must be able to ask contrastive why questions through a process they call explanation as exploration to best understand black box models. Our approach of explanation as an iterative process is similar to what they propose.

Hoffman et al.~\citeyear{hof18} give a thorough overview on evaluating explanations. They focus mainly on techniques that can be used to measure XAI constructs. These include measuring explanation goodness and satisfaction, mental models, curiosity, and trust. We adopted the proposed metrics for evaluating explanation goodness and satisfaction in our experimental evaluation in Section~\ref{sec:userstudy}.

There have also been recent efforts within XAI on \emph{human-centered} explainability \cite{miller2017inmates,mil19,lim09}. 
Until recently, there has been only limited investigation to determine what users actually want explained.
Haynes, Cohen, and Ritter~\citeyear{hay09} proposed a taxonomy, based on that of Graesser et al.~\citeyear{gra92} and Lehnert~\citeyear{leh78}, that they empirically tested using a pilot simulation tool. In this case, they found that \textit{definition} questions were the most common, as the users were tasked with operating an unfamiliar tool. Their end-goal was to understand how to use the tool rather than a plan.

Lim, Dey, and Avrahami~\citeyear{lim09} looked at the following questions, ``why", ``why-not", ``how", and ``what if". They found that ``why" and ``why-not" questions led to the best improvement in understanding amongst users.
Penney et al.~\citeyear{pen18} used this taxonomy to study how users foraged for information in the game StarCraft.
For assessing an agent, they found that ``what" questions were asked most frequently (over 70\%), followed by ``why" questions. They reasoned that, in their domain, the most common questions related to uncovering hidden information about the current and past states.

In contrast, the taxonomy presented in Section~\ref{sec:taxonomy} shows that, when presented with a plan, user questions are more commonly the ``why'' questions that relate to actions.


\subsection{Explainable AI Planning}

In the specific context of AI Planning, explanation has received attention from a number of researchers. We consider relevant related work in this section, focusing on three issues: the role of model reconciliation, the role of contrastive explanations and the explanation of unsolvable planning problems.

\subsubsection{Model Reconciliation}

Chakraborti et al.~\citeyear{cha17} adopt the position that explanation is a {\em model reconciliation problem (MRP)}~-- namely, that the need for explanation is due to differences between the agent's and the human's model of the planning problem.  The planning system therefore ``suggests changes to the human's model, so as to make its plan be optimal with respect to that changed human model''.
In that work, Chakraborti et al. describe an approach for generating minimally complete and monotonic explanations that update the users model, so that it will accept a plan as correct. The approach assumes that the user model is known. 
In contrast, Sreedharan~\citeyear{sre18} generates conformant explanations that are applicable to a set of possible user models in cases where the user's model is not precisely known. 
Both approaches consider only {\em optimal} solutions for classical planning problems. In general, the assumption that plans must be optimal in order to support explanation is troublesome. Optimal planning for temporal and numeric problems (which we consider here) is undecidable. In addition, the metric or preferences by which plans should be assessed might differ between the human and agent, as Smith~\citeyear{smi12} suggests. Of course, the preferences and optimisation criteria could be considered as part of the model, in which case the {\em model reconciliation} perspective would still be appropriate.  However, to date, these aspects of the model have not received much attention.

As in the model reconciliation work, we suppose that the user and agent may have different models. We also suppose that they may have different preferences, and different computational abilities.
However, we don't assume that either the agent or the human have direct knowledge about the other's model or capabilities, 
or that they do optimal planning.
Instead, we have considered the problem of plan explanation to be one of {\em model restriction} 
where the human can impose additional constraints on the planning problem (through contrastive questions).
As we noted in Section~\ref{subsec:problem}, model restriction could be considered as a special case of model reconciliation, 
in which the planning agent's model is temporarily revised by imposing the constraints implied by a contrastive question.
However, these constraints do not result in permanent changes to the agent's model of the world, or model of the planning domain~--
they are temporary restrictions on the plan trajectory.
It's also true that the human's interaction and questioning might result in 
evolution of their own model of the planning domain or problem. 
However, we do not assume that the agent has any knowledge of the human's model, 
and we do not attempt to model the human's learning process.

A complementary line of work is that of generating plans that are more \textit{explicable} within the framework of a human user's model. Zhang et al.~\citeyear{zha17} proposed generating explicable plans by postulating that humans understand plans by associating the actions in the plan with abstract tasks. They learn what abstract tasks humans associate to actions and use this to produce new more explicable plans. 
Kulkarni et al.~\citeyear{kul16} model explicability by first computing the distances between the plans generated by a planning agent and plans expected by the user. Human subjects are then asked to label the actions in the agent plans as explicable. These results are used along with the plan distances to form a regression model called explicability distance. Explicability distance is then used as the heuristic to search for explicable plans. It is not always possible to produce explicable plans if the agent model does not allow for what the user expected and explanations are needed. 

Work on generating explicable plans can be seen as complementary to work on explanation; explicable plans should reduce the need for explanation, but do not eliminate it. To date, this work has assumed different models for the planner and the human.  As a result, an explicable plan is one which is reasonable in the human's model, so that a planner can use that model to generate explicable plans. This work has not seriously addressed situations in which the models are the same, but the plan is simply too complicated for the user to understand easily or quickly. In this case, a more explicable plan would presumably be one which is smaller and simpler for the human to reason about. This work has also not seriously addressed situations where the human and agent have the same models of actions and of the world state, but different preferences or optimisation criteria. In this case, an explicable plan would be one which is good according to the user's preferences and optimisation criteria.


\subsubsection{Contrastive Explanations of Plans}

%

In addition to the work previously cited~\cite{fox17}, highlighting the role of contrastive questions in XAIP, 
Eifler et al.~\citeyear{eif20} approach answering local contrastive questions by explaining the reason that a contrast case \textit{B} was not in the plan, or a feature of the plan, by using the properties that would hold if \textit{B} {\em were} the case.  This is in contrast to our approach in which we give a specific plan trace containing \textit{B}.
Kim et al.~\citeyear{kim19} detail a general approach for generating differences between plan traces using Bayesian inference with search for inferring contrastive explanations as linear temporal logic specifications. These resulting differences can then be used to generate contrastive explanations. However, they do not consider how to generate the different plan traces, or how they can be used to answer specific questions.
Kasenberg et al.~\citeyear{kas19} focus on justifying an agent's behaviour based on deterministic Markov decision problems. They construct explanations for the behaviour of an agent governed by temporal logic rules and answer questions including contrastive \textit{``why"} queries. However, they only recognise implicit contrastive questions of the form \textit{``why A?"} (or \textit{``why }$\neg$\textit{A?"}) which they explain by citing rules and goals dictating that they had to (or could not) perform \textit{A}. 
Whereas we argue that users must be able to ask explicit contrastive questions of the form \textit{``why A rather than B?"} which can usually only be answered by generating a hypothetical plan where \textit{B} is included and \textit{A} is absent.
Bercher et al.~\citeyear{ber14b} also provide explanations for implicit contrastive \textit{why} questions in a system that helps users to assemble a home theatre. The explanations consist of the set of reasons that an action is present in the plan. Chakraborti et al.~\citeyear{cha19b} show how these kinds of explanations can be minimised by selecting the most relevant explanatory content. Like the previous work by Kasenberg et al., 
these approaches do not utilise more comprehensive techniques for counterfactual reasoning in order to construct alternative plans, and compare them with the original.

\subsubsection{Unsolvability}
A special kind of ``why'' question is: ``why didn't you find a solution to this problem?'' 
While there has been recent work on the generation of unsolvability certificates for planning problems~\cite{eri17,eri20},
these are not very satisfying as explanations. 
Gobelbecker et al.~\citeyear{gob10} argue that excuses must be made for why a plan cannot be found. These are counterfactual alterations to the planning task such that the new planning task will be solvable. They provide an algorithm to produce these excuses in a reasonable time.
Sreedharan et al.~\citeyear{sre19} address the problem of explaining unsolvability by considering relaxations of the planning problem until a solution can be found, and then looking for landmarks of this relaxed problem that cannot be satisfied in less relaxed versions of the problem. The unsatisfiability of these landmarks provides a more succinct description of critical propositions that cannot be satisfied. Eifler et al.~\citeyear{eif20} has taken a somewhat different approach by deriving properties that must be obeyed by all possible plans. Although this is not the focus of this work, these too could serve as explanations in cases of unsolvability. 

We have not attempted to provide explanations for unsolvability of planning problems in this paper.
However, since we allow for the user's and agent's models and computational abilities to differ,
it is certainly possible that the model restriction imposed by a contrastive question may result in a planning problem that is unsolvable by the planning system. Addressing this issue is left to future work.

\section{Conclusion}\label{sec:conc}

In this paper we considered the problem of plan explanation to be an iterative process, 
in which the user repeatedly asks questions that are contrastive in nature.
To motivate our focus on contrastive questions, in Section~\ref{sec:taxonomy} we presented a user study examining the kinds of questions users asked about problems in three small planning domains. 
This study demonstrated that the vast majority of user questions were indeed contrastive. 
We categorised these questions into a taxonomy of 7 different types, which served as the focus for the remainder of the paper. 


Each contrastive question, such as ``{\em why did you do A rather than B?}'' leads to a ``constrained'' or {\em restricted} planning problem -- in this case, one in which ``B'' must be in the plan instead of ``A''.  
This restricted planning problem must then be solved by the planning system in order to compare and contrast the user's proposed alternative solution (foil) with the original plan generated by the planning system. 

Through this iterative process, the user is able to explore the space of possible solutions.  
This may result in the user increasing their understanding of the problem and the solutions being proposed by the planning system, 
but may also lead to improvement of those solutions, as a result of the constraints imposed by the user's questions.  
Ultimately, it is up to the user to decide when they are satisfied with the resulting plan.
This process increases transparency for the user, and ultimately leads to greater trust and understanding of the planning problem and the possible solutions.

\subsection{Contributions}

There are several contributions in this paper:
\begin{itemize}
    \item In Section~\ref{sec:taxonomy} we hypothesised that users are more likely to ask contrastive why questions about plans. We supported this hypothesis with a user study, and formalised the results into a contrastive taxonomy of questions.
    \item In Section~\ref{sec:plans} we formalised the iterative process as being one of {\em model restriction}, where each contrastive question leads to a hypothetical problem characterised by restrictions imposed on the model of the planning problem.
    \item In Section~\ref{sec:comp} we showed how these restrictions can be compiled into a PDDL2.1 model for the contrastive questions in our taxonomy. 
    \item In Section~\ref{sec:iter} we presented the implementation of this framework as a service - namely as a wrapper around an existing temporal/numeric planning system, with a simple user interface for comparing plans. 
    \item In Section~\ref{sec:eval} we evaluated the computational consequences of model restriction, and solution quality of the plans generated from those restricted models, and efficacy of the explanation framework as a whole.
\end{itemize}
   

In Section~\ref{sec:eval_perf} we evaluated the impact of adding constraints on the efficiency of the plan generation process for the constraints imposed by the different types of contrastive questions.  
The results show that for the majority of problems and questions, the planning time for restricted planning problems is quite similar to that for the original unconstrained problem.
As we noted, adding constraints to a planning problem reduces the number of possible solutions and could make it more difficult to find a plan solution.  
However, the constraints can also rule out significant portions of the plan search space, 
making it easier for the planner to find a solution. 
Predicting the performance impact for a particular question and problem is therefore not easy or obvious, 
but in general, performance does not appear to be a significant issue.
The results in Section~\ref{sec:eval_iteration} shows that the number of constraints is of less importance on the planning time than the ultimate constrained problem, whether this be the result of one compilation or of multiple. 
This demonstrates that the difficulty in answering questions does not necessarily increase with the number of questions asked, and therefore supports our iterative model restriction approach.

We also evaluated the impact of constraints on plan quality.  
For easier problems the compilations had little impact on the quality of plans produced by the planner, but for harder problems, the addition of constraints sometimes resulted in significant improvement in plan quality.
This is an important feature of the plan negotiation problem, where a user can impose restrictions that lead to a better plan.
This demonstrates that the original plan was not optimal and that the addition of constraints can actually narrow the search space in a way that guides the planner toward better quality solutions.

Through a comparative user study, described in Section~\ref{sec:userstudy},
we evaluated the effectiveness of the framework in improving the user's understanding of the proposed plan. 
The results indicate that the iterative framework and plan comparison helps users to understand plans better. 
In particular, it helps them to understand why actions are in the plan, 
why actions are in a particular order, and how changes affect the plan.
However, the study also pointed out some of the weaknesses in the current system, 
namely in the quality of the explanations. 
User's expected a more in-depth explanation of the differences between plans, 
rather than just a simple highlighting of the differences.
We discuss this further in the next section.

\subsection{Future Work}
All of the below issues present interesting technical challenges that warrant further investigation.

\subsubsection*{Compiling Constraints}


Current PDDL languages do not have the ability to express constraints on action inclusion, exclusion, or ordering, and do not allow us to place more complex constraints on \textit{how} something is achieved or on plan structure. For example, the question ``Why did you use action A rather than action B for achieving P?'' requires planning with the HModel where B is required to be in the causal support for achieving P, but A is not in that causal support. This is substantially more difficult than just excluding A from the plan and forcing B into the plan. This has been touched upon in the paper with a first step towards a solution in Section~\ref{comp:keycausal} and a further discussion of the nuance of this issue in Section~\ref{comp:just}. However, we believe for a robust solution, LTL will likely play a key role in defining the semantics of any new language which enables the expression of such constraints.

As we discussed in Section~\ref{comp:composition}, there are possible questions that cannot obviously be represented in the vocabulary of the planning model. For example, the user might ask ``why did it take so long to accomplish A?'' Such a constraint requires a richer language that allows one to impose trajectory constraints on the plan itself, e.g. requiring that A be achieved before some time limit. Such trajectory constraints might be expressible in PDDL 3.0, but this issue requires further investigation. These types of questions might be unexplicitly expressible through a series of questions which give the same effect, however this currently requires a user to infer which questions should be asked and in which order to achieve this. In the future we would like to automatically infer these more robust questions from natural language. We believe that both issues highlighted in this section are needed for automatic inference of complex questions, and that a contrastive language that is both representable, expressible, and actionable will be a crucial part of any solution.

\subsubsection*{Explanation}

Currently, our explanations for contrastive questions consist of comparing two plans side-by-side, highlighting the action differences between them.
However, there is clearly room to do much more. We do not take advantage of the causal structure of the plans within our explanations. 

As we indicated above, in the user study, the users wanted deeper explanations of the differences between plans. Considering the causal structure of the two plans and how they differ appears to be part of the solution. For example, a plausible explanation might be ``because you asked me not to use action A, I had to use another action to achieve P, and action B appeared to be the best choice.''  However, this approach is likely only part of the solution -- abstraction may also play a key role.  
For example, abstracting away some of the predicates, variables, and actions that are the same in both plans would allow the explanation to focus on the key differences between the two plans.  This is related to the problem of explaining unsolvability discussed below.

\subsubsection*{Unsolvability}

By adding constraints to a planning problem, it's possible that the problem will become unsolvable.
Sreedharan et al.~\citeyear{sre19} address the problem of explaining unsolvability by considering relaxations of the planning problem until a solution can be found, and then looking for landmarks of this relaxed problem that cannot be satisfied in less relaxed versions of the problem. The unsatisfiability of these landmarks provides a more succinct description of critical propositions that cannot be satisfied. This approach is appealing, but for temporal/numeric planning problems, it becomes more challenging. This is because the number and character of possible relaxations increases dramatically. For example, in addition to abstracting away certain predicates, one could consider abstracting away some of the numeric variables, or relaxing certain action preconditions, or allowing actions to have arbitrary durations. A more targeted approach would be to consider relaxing the constraints imposed by the user's question until the problem becomes solvable again. 
The removed constraints could then be analysed to determine how they prevent milestones from being achieved. 
A satisfying explanation for the question ``Why did you not do B?'', might be something like ``Action B takes too long and makes it impossible for me to achieve (milestone) M''. The advantage of this more targeted approach is that it reduces the number of abstractions that need to be considered -- only the actions involved in the user imposed constraints, and the variables and predicates they reference, need to be considered for abstraction.



\section*{Acknowledgements}
This work was partially supported by 
{EPSRC}{} grant 
{EP/R033722/1} for the project Trust in Human-Machine Partnership (THuMP) and 
AirForce Office of Scientific Research award number FA9550-18-1-0245.


\appendix
\section{Contrastive Taxonomy - User Study and Analysis}
\label{appendixA}

\subsection{Purpose} \label{ref:methodology}
We conducted a study to test our hypothesis that users would ask more local, contrastive \textit{why} questions than global \textit{how}, or \textit{what} questions.

Our null hypothesis and alternate hypothesis, $H_0$ and $H_a$ are as follows:

\begin{quote}$H_0$: Users ask an equal distribution of \textit{why}, \textit{how}, and \textit{what} questions about planning scenarios, when the model is well known.\end{quote}

\begin{quote}$H_a$: Users ask more \textit{why} questions than \textit{how} or \textit{what} questions about planning scenarios, when the model is well known.\end{quote}

We also wanted to support our compilation approach by creating a taxonomy of the user questions we procured, called the \textit{Contrastive Taxonomy}.


\subsection{Methodology}
%
We designed a study to elicit questions from users about plans. We recruited participants, from a website (\url{https://wwww.prolific.co)} that specialises in sourcing eligible subjects, each of which were compensated \pounds{10} for their time. 
We selected a sample size of 15 which is a typical number for this type of study~\cite{cha18,kul19}.
The participants were from different, non-planning related, backgrounds and professions between the ages of 21 and 39 years old. In the study, participants were presented with three planning scenarios which were described to them in detail. 
They were first asked to watch a short video~\cite{pla19} showing an animation of a planning problem being performed~\footnote{https://youtu.be/MSCakpJUcpc}. 
Once the participants were familiar with the content of the animation
they were asked to re-watch the video
and write down any questions they had, and (if applicable) a reason for why they asked the question. For each question the participants were told to note down the time during the video that caused it to be asked. 
The participants were asked to do this twice more with two videos of new planning problems. However, this time they were told to only ask questions that were specific to decisions, or actions that were made in the plan. Participants were asked to re-watch each video until they could not produce any new questions.
From this data, we performed a content analysis to extract a taxonomy of the types of questions that people answered in our scenarios.

We used three different scenarios in the study: (1) a family of five must sail to the other side of a river, with some constraints placed on sailing the boat; (2) a logistics problem where six packages have to be delivered to specific locations using trucks and aeroplanes; and (3) a robot must place different objects into positions on a grid. We chose these domains because they are simple to understand and reason about without much participant training, and are varied in what they model.


\begin{table}[t]
\centering
\begin{tabular}{ l p{13cm} r} 
 \toprule
 & Question Type& \#\\
 \midrule
FQ1 & Why is action A not used in the plan, rather than being used?        &   17\\
FQ2 & Why is action A used in the plan, rather than not being used?        &   75\\
FQ3 & Why is action A used in state S, rather than action B?              &   35\\
FQ4 & Why is action A used outside of time window W, rather than only being allowed within W?        &   6\\
FQ5 & Why is action A not performed before (after) action B, rather than A being performed after (before) B?         &   10\\
FQ6 & Why is action A not used in time window W, rather than being used within W? &   2\\
FQ7 & Why is action A used at time T, rather than at least some time T' after/before T?       &   6\\
FQ8 & Non-contrastive & 17\\
 \bottomrule
\end{tabular}
\caption{Frequency of questions categorised by the Contrastive Taxonomy, the types are numbered for ease of reference.}
\label{table1}
\end{table}


\begin{table}[t]
\centering
\begin{tabular}{lrrrr} 
 \toprule
 Question  &  &  & \\
 Type &     Video 1  &    Video 2  &   Video 3    \\
 \midrule
 What?         &   2  &  1    & 3\\
 How?          &   0    &  3    & 2\\
 Why?          &   65 &  50 & 42\\
 \bottomrule
\end{tabular}
\caption{Frequency of questions by video categorised by Miller's taxonomy.}
\label{table2}
\end{table}


\subsection{Results}
The results of our study are shown in Tables~\ref{table1} and \ref{table2}. Table~\ref{table1} shows the the number of questions in each category in our taxonomy of formal questions (FQ). Table~\ref{table2} shows the number of questions in each category in the taxonomy proposed by~\citet{mil19} and compares the questions asked in Video 1, where users were asked to propose any questions, and the videos 2 and 3, where the questions had to be related to the plan. 

We categorised the questions into the Contrastive Taxonomy by splitting the question into the fact and the, sometimes implied, contrast case. 
For example, take the question, posed by a participant about the first planning situation:
\begin{quote}
``Why did Son swap with Fisherman?... Fisherman should [have] pick up Daughter''    
\end{quote}
The fact is that the Son swapped places on the boat with the Fisherman, and the contrast case is that the Fisherman should have picked up the Daughter.
If, like in this example, the contrast case was an explicit action which the user expected in place of some other action in the plan, the question was categorised as type FQ3. If the contrast case was implicit or explicitly negating the fact, the question was categorised as either type FQ1 or FQ2, depending on the fact. If the user questioned an ordering between two actions the question was categorised as type FQ5. If the question was in these types FQ1 or FQ2 but referred to a specific time, it was categorised as type FQ4, FQ6 or FQ7.  \textit{What} and \textit{how} questions were categorised as type FQ8, as they are not regarded as contrastive questions. We also categorised any question about the video itself, i.e discrepancies in the animation as FQ8. Our proposed compilation techniques do not extend to these types of questions. 
The questions were categorised into Miller's taxonomy by the interrogative word used in the question. 

\subsection{Analysis}
The results in Table~\ref{table2} show that users want to understand why certain decisions were made by the planning system rather than how the planning system works, or what a specific component of the system's purpose is. There were 157 instances of contrastive \textit{why} questions, of which 151 were contrastive \textit{why} questions in reference to specific decisions made in the plan. There were only 11 \textit{what} and \textit{how} questions which asked for a deeper understanding of the planner behaviour. Of the questions posed by users, 89.9\% were contrastive \textit{why} questions.

Performing a chi-square test, $\chi^2(2, 168) = 273.25$, P-value $< 0.00001$, these results are therefore significant at $p < 0.001$. We can therefore reject our null hypothesis, $H_0$, and accept our alternate hypothesis, $H_a$.

Table~\ref{table2} supports this further when comparing the results of video 1 with videos 2 and 3.
This shows that when there are no constraints on the questions users can ask, or when they are explicitly asked to question the plan, in both cases they want to understand why certain decisions were made in the plan.

There were a small number (6) of \textit{why} questions that classified as out of the scope of this paper. These were questions that were not related to the planning system or the plan produced. For example a participant asked about one of the videos, ``Why did the pink square change to green?'' This is not a question about a decision but the inner workings of the animation software used. Another asked the question, ``Why am I still expecting the AI to make a human logical decision?'', which is clearly a complex question outside the scope of this paper. Notice that these questions are still contrastive in nature, just not questions relevant to planning systems and therefore not ones we are concerned with answering. 

Table~\ref{table1} shows that the most commonly asked questions (type FQ2) are about actions that were performed, rather than absent actions they expected to have been in the plan.
However, when users do question why an action they expected did not happen, they are more likely to ask it as an explicit contrastive question with respect to some other action that did happen (type FQ3). Users do not question the times in which actions are performed (types FQ4, FQ6, FQ7), or the ordering of actions (type FQ5) as much as why an action was performed or not. The results show that the majority of user questions are constrastive. The contrast case is more likely to be the negation of the fact (types FQ1, FQ2, FQ4 - 7). However, a significant proportion of the questions specify a specific action as the contrast case which the user expected to have been performed instead of the factual action (type FQ3). This shows that users likely have an idea (mental model) of a plan which they use to question the factual plan. They might question why an action was performed, when it was not part of their ideal plan. Or they might question why an action, that was present in their ideal plan, did not appear in the factual plan. 

We can provide compilations of 89.8\% of the 168 questions that users asked.
\bibliography{bib}
\bibliographystyle{theapa}
\end{document}